\documentclass[twoside,11pt]{article}

\usepackage{jmlr2e}

\usepackage{amsmath, amssymb}
\usepackage{tikz}
\usepackage{color}
\usepackage{minitoc}
\usetikzlibrary{matrix}
\usepackage{etoolbox}
\usepackage{comment}
\usepackage{float}
\usepackage{array}
\usepackage{colortbl}
\usepackage{array,multirow,makecell}
\usepackage{caption}
\usepackage{graphicx}
\usepackage{bbm}
\RequirePackage[shortlabels]{enumitem}

\usepackage{ulem}

\newcommand{\Pb}{\mathbb{P}}
\newcommand{\ep}{\epsilon}
\newcommand{\de}{\delta}

\newcommand{\om}{\omega}

\def\cC{{\mathcal{C}}}

\newcommand{\cov}{N_{\Omega}^{(c)}}
\newcommand{\pac}{N_{\Omega}^{(p)}}
\newcommand{\di}{dim\, \Omega}

\newcommand{\covh}{\widehat{N}_{\Omega}^{(c)}}
\newcommand{\pach}{\widehat{N}_{\Omega}^{(p)}}

\newcommand{\covs}{N_{\om_1,\ldots,\om_n}^{(c)}}

\newcommand{\covsap}{N_{\om_1,\ldots,\om_n}^{(ap.c)}}

\newcommand{\covhap}{\widehat{N}^{(ap.c)}_{\Omega}}

\newcommand{\pachne}{\widehat{N}^{(p.new)}_{\Omega}}

\newcommand{\covbi}{N_{\Omega'}^{(c)}}

\jmlrheading{1}{2019}{1-48}{1/10}{10/00}{meila00a}{Yann Issartel}

\ShortHeadings{Graphon dimension}{Issartel}
\firstpageno{1}

\begin{document}

\title{On the Estimation of Network Complexity:\\ Dimension of Graphons}

\author{\name Yann Issartel \email yann.issartel@math.u-psud.fr \\
       \addr Laboratoire de Math\'ematiques d'Orsay\\
       Universit\'e Paris-Saclay\\
       Orsay, 91405, France}

\editor{}

\maketitle

\begin{abstract}Network complexity has been studied for over half a century and has found a wide range of applications. Many methods have been developed to characterize and estimate the complexity of networks. However, there has been little research with statistical guarantees. In this paper, we develop a statistical theory of graph complexity in a general model of random graphs, the graphon model.

Given a graphon, we endow the latent space of the nodes with the neighborhood distance. Our complexity index is then based on the covering number and the Minkowksi dimension of this metric space. Although the latent space is not identifiable, these indices turn out to be identifiable. This notion of complexity has simple interpretations on popular examples: it matches the number of communities in stochastic block models; the dimension of the Euclidean space in random geometric graphs; the regularity of the link function in H\"older graphons.   

From a single observation of the graph, we construct an estimator of the neighborhood-distance and show universal non-asymptotic bounds for its risk, matching minimax lower bounds. Based on this estimated distance, we compute the corresponding covering number and Minkowski dimension and we provide optimal non-asymptotic error bounds for these two plug-in estimators.
\end{abstract}

\begin{keywords}random graph model, graphon, neighborhood distance, covering number, Minkowski dimension.
\end{keywords}

\section{Introduction}
Networks appear in many areas where data is a collection of objects interacting with each other. Examples include numerous phenomena in the fields of physics, biology, neuroscience and social sciences. A major issue is to extract information from these data repositories. This exciting challenge has led researchers to seek characterizations of networks, among which their complexity has received a lot of attention for more than half a century. See \citep{dehmer2011history,zenil2018review} for two recent reviews. Indeed, network complexity is a key feature used in various applications, for example, to quantify the complexity of chemical structures \citep{Biobonchev2005quantitative}, to describe business processes \citep{business}, to characterize software libraries \citep{softwareLibrary}, and to study general graphs \citep{generalgraph}.  

The definition and estimation of network complexity is an active line of research \citep{morzy2017measuring,zufiria2017entropy,claussen2007offdiagonal}. However, there appear to be little (or no) mathematical results on the statistical side of the problem. In this paper, we develop a statistical theory of graph complexity in a universal model of random graphs. To the best of our knowledge, it is the first contribution on complexity estimation with statistical guarantees.

\subsection{Modeling assumption}
Statistical inference on random graphs is a fast-growing area of research \citep{matias2014modeling,bubek2, abbe} and has found a wide range of applications \citep{goldenberg2010survey,sarkar}. Usually, it assumes there exists an unknown feature in the underlying model and the goal is to recover this feature from a single realization of the random graph. 

Here, we follow this direction with the \textit{W-random graph model} (also known as graphon model). This general model falls into the category of non-parametric descriptions of networks \citep{Bickel} and satisfies some forms of universality \citep{diaconis}. See section \ref{sect:literature:intro} for details. In this paper, we define a notion of complexity for this model and then consider the problem of inferring this complexity from a single graph observation. 

W-random graphs allow to model many real-world networks, such as social networks where nodes represent different people and edges people's friendships. In this example, one may expect that the friendship probability $p_{ij}$ between individual $i$ and $j$ depends on their personal attributes (like jobs, ages, leisure). To model such mechanism, one may assume the observed graph is generated according to the W-random graph model, i.e. (1) for each node $i$ of the network, an attribute $\om_i$ is drawn from a distribution $\mu$ on a space $\Omega$ (where $\Omega$ can be seen as the social space of all possible individual features: jobs, ages,\ldots); (2) two people are friends, independently of the others, with probability $p_{ij} = W(\om_i,\om_j)$, where $W:\Omega \times \Omega \rightarrow[0,1]$ is a symmetric function. Thus, a W-random graph is specified by the triplet of parameters $(\Omega, \mu, W)$, often called \textit{graphon} in the literature \citep{lovasz2}. 

Such modeling falls into the popular ``latent space approach" \citep{hoff2002latent}. Indeed, the personal attributes may not be observed in practice and accordingly, the W-random graph model assumes that the $\om_i$ and $\Omega$ are latent (unobserved). In fact, all parameters of the graphon $(\Omega, \mu, W)$ are unknown, and the only observation is the edges of the graph, i.e. the adjacency matrix $A$ where $A_{ij} = 1$ stands for the presence of an edge between the $i^{\textup{th}}$ and $j^{\textup{th}}$ nodes, and $A_{ij} = 0$ otherwise. See Section \ref{section::W_random_graph} for a formal presentation of this model.

\subsection{Contribution}

\subsubsection{Complexity index}

Our first objective is the definition of a complexity index in the W-random graph model. As a natural candidate, one might think of the dimension of the latent space, like $d$ if $\Omega = [0,1]^d$. However, this index is inadequate because of a major identifiability issue. Indeed, it is known that \citep[see][]{lovasz2} the attribute space $\Omega$ is not identifiable from the observed adjacency matrix $A$. Even worse, it has been shown that all W-random graph distributions can be represented on the specific space $\Omega = [0,1]$ \citep{lovasz2}. It is therefore pointless to think about the graph complexity purely in terms of the latent space. Likewise, the regularity of the link function (like $\alpha$ if $W$ is $\alpha$-H\"older) is not suited due to the non-identifiability of $W$. 




These issues motivate the introduction of a more abstract index. Given a graphon $(\Omega, \mu, W)$, we endow the latent space $\Omega$ with the so-called \textit{neighborhood distance} \begin{equation}\label{intro:l2neigh}
r_W(\om, \om') =  \left( \int_{\Omega} \left| W(\om,\om'') - W(\om',\om'') \right| ^2  \hspace{0.1cm}\mu(d\om'') \right)^{1/2}.
\end{equation}From the above description of a W-random graph, we can see that the quantity $r_W(\om_i,\om_j)$ measures the propensity of the nodes $i$ and $j$ to be connected with similar nodes. Our complexity index is then defined as the covering number and the Minkowski dimension of a purified version of the (pseudo-) metric space $(\Omega, r_W)$. The purification process is detailed in section \ref{correction_definition_pure_graphon}. Recall the definitions of these two standard measures for metric spaces: the $\ep$-covering number $\cov(\ep)$ is the minimal number of balls of radius $\ep$ required to
entirely cover the (pseudo-) metric space $(\Omega, r_W)$. And the Minkowski dimension is the following limit on the covering number
\begin{equation}\label{intro:def:dim}
\di   := \underset{\epsilon \rightarrow 0}{\textup{lim}}\ \, \frac{\log \, \cov(\epsilon)}{- \textup{log} \, \epsilon} 
\end{equation}when the limit exists. In particular, the Minkowski dimension does not have to be an integer.

Although none of the three parameters $\Omega, \mu$ and $W$ are identifiable in the W-ranom graph model, we prove that the covering number and the Minkowski dimension of a purified version of $(\Omega, r_W)$ are identifiable.

We also illustrate that this notion of complexity is sound on classic examples of random graphs. Specifically, we show that $\cov(\ep)$ is equal to the number of well-spaced communities in the stochastic block model; that $\di$ matches the dimension of the Euclidean space in some random geometric
graphs; and that $\di$ is equal to the regularity of the link function in some H\"older graphon models. See Section \ref{section::illustExample} for details.

In addition to all applications listed in the introduction, these complexity indices may also be useful to adjust analytical methods to particular networks, for example, when estimating the link function $W$ (see section \ref{sect:literature:intro} and \ref{section::illustExample} for related comments) or in learning representation where the goal is to find an informative metric space to place/represent the nodes of the network \citep{hoff2002latent, perozzi2014deepwalk, grover2016node2vec}.


\subsubsection{Statistical estimation} 

From the observed adjacency matrix $A$ of a W-random graph, we estimate the neighborhood distance \eqref{intro:l2neigh} on the sampled points $\om_1,\ldots,\om_n$. The corresponding distance estimator $\widehat{r}$ is defined in Section \ref{subsec:def:disEstim}. We show universal non-asymptotic bounds for its risk (Theorem \ref{intro:estim:dist}). Let $\om_{m(i)} \in \{\om_1,\ldots,\om_n\}\setminus \{\om_i\}$ denote a nearest neighbor of $\om_i$ with respect to the distance $r_W$.

\begin{theorem}\label{intro:estim:dist} Consider the distance estimator $\widehat{r}$, defined in Section \ref{subsec:def:disEstim}. Then, for any graphon $(\Omega, \mu, W)$, we have\\
$\forall i,j \in [n],$
\begin{equation*}
  \left| r_W^2(\om_i,\om_j) - \widehat{r}^2( i, j) \right| \lesssim r_W(\om_j,\om_{m(j)}\,) + r_W(\om_i,\om_{m(i)}\,) + \sqrt{\textup{log}(n)/n }\end{equation*}
 
\noindent with probability at least $1-2/n$.
\end{theorem}In the upper bound, there is a bias term $r_W(\om_j,\om_{m(j)}\,)$ which is the distance between the sampled point $\om_j$ and its nearest neighbor $\om_{m(i)}$ (w.r.t. the neighborhood distance). This bias depends on the form of the underlying graphon $(\Omega, \mu, W)$, for example, it is equal to zero w.h.p. in the stochastic block model (i.e., when the link function $W$ is piecewise constant on $\Omega = [0,1]$). We also derive a minimax lower bound that matches the upper bound of Theorem \ref{intro:estim:dist}. See Section \ref{subsec::consitencyDistestim} for details on the distance estimation.

Based on the estimated distances $\widehat{r}(i,j)$, we estimate the covering number $\cov(\epsilon)$ by plug-in and provide universal non-asymptotic error bounds for this estimator. See Section \ref{subsection::theoreticGuaranteeCovNUmb} for details. Our results on the distance and covering number are therefore valid for all graphons, unlike most results in the graphon literature.

Combining the above covering number estimator $\covh$ with formula \eqref{intro:def:dim}, we derive an estimator of the Minkowski dimension 
\begin{equation*}
    \widehat{dim}_D := \frac{\textup{log} \, \covh(\epsilon_{D})}{-\textup{log}\,\epsilon_{D}}
\end{equation*}which satisfies a high probability convergence rate (Theorem \ref{intro:thm:dimestim}). For this result, we assume the Mikowski dimension is upper bounded by some constant $D$ and use a particular radius $\ep_D$ defined in Section \ref{section:estimDim}. We also make some mild assumptions on the graphon geometry, which are inspired by the problem of estimation of manifold dimension (see section \ref{sect:literature:intro} for this related literature). Besides, we show that this set of assumptions is minimal, in the sense that, if any of these assumptions is removed, all dimension estimators make an estimation error of the order $1$. 

\begin{theorem}\label{intro:thm:dimestim}Under some mild assumptions, defined in Section \ref{section:estimDim}, the following holds. If $\di$ is any real in $[0,D]$, then
\begin{equation*} \left|\, \widehat{dim}_D - \di \,\right| \lesssim \ \,\frac{1}{\textup{log}\, n} 
\end{equation*}
with probability at least $1-C'/n$ for some constant $C'$ independent of $n$.
\end{theorem}Finally, we prove that the upper bound $\log^{-1}n$ is optimal, which means that no estimator can improve on this error. For detailed results, see Section \ref{section:estimDim}. 

As extensions, we show that the above results also cover the important setting of sparse networks, which has been considered several times in the literature \citep[see][]{bickel2011method,wolfe,klopp, Massoul}. In addition, we describe a polynomial-time algorithm to approximate the covering number estimator; we do so by using a classic greedy algorithm that is known to satisfy some theoretical guarantees. See Section \ref{section::further_consideration} for these two extensions. 

Finally, we test if the packing number (of a purified version of $(\Omega,r_W)$) is smaller than $K$, with a specific care for controlling the type I error probability uniformly over all graphons. We prove this error is smaller than $2/n$ for any graphon. For technical reasons detailed in Section \ref{subsec::test_complexity}, we use here the packing number instead of the covering number, which are essentially the same measures (see Appendix \ref{subsec::additionalInfo} for a reminder about these usual measures for metric spaces).

\subsection{Connection with the literature}\label{sect:literature:intro}

\subsubsection{W-random graph model}

The most simple random graph is the Erd\"os-R\'enyi model where each edge has the same probability $p$ of being present, independently of the other edges. The study of this generative model has been impressively fruitful in mathematics \citep{bollobas1998random} but does not replicate even the simplest properties of real-world networks. Hence, the assumption of a constant connection probability $p$ has been relaxed in the celebrated stochastic block model \citep{1983stochastic} where the connection probabilities may vary with the community membership of each node. Although this model has attracted a lot of attention \citep{abbe}, it fails to catch some subtle aspects of very large graphs. Such modeling issues have led to a non-parametric view of network analysis \citep{Bickel}, in particular the introduction of the W-random graph model \citep{diaconis}. 

The universality of the W-random graphs has two parts. On the one hand, the graphon $(\Omega, \mu, W)$ plays a key role in network analysis as a powerful representation of many graph properties. Indeed, it has been shown that many sequences of growing graphs can be represented by graphons. For details, see the theory of graph limits introduced by \citet{lovasz2006limits} or the comprehensive monograph by \citet{lovasz2}. On the other hand, the W-random graph model is connected with the theory of exchangeable random graphs. In fact, every distribution on random graphs that is invariant by permutation of nodes can be expressed with W-random graphs \citep{diaconis,aldous,kallenberg}. Thus, the W-random graphs encompass many random graph models, including stochastic block models, random geometric graphs \citep{penrose} and random dot product graphs \citep{tang2013universally,athreya2017statistical}.

\subsubsection{Graphon estimation}

There has been much interest in the recovery of the function $W$ (or the matrix of probabilities $[W(\om_i,\om_j)]_{i,j\leq n}$) on the specific space $\Omega = [0,1]$. Usually, authors assume the graphon has some regularity (e.g. $W$ is H\"older continuous on $[0,1]$) and then use an approximation by SBM, which can be seen as an approximation by constant piecewise functions of $W$ \citep{borgs2015private,wolfe,gao,klopp,latouche2016variational}. We also mention an alternative approach based on neighborhood-smoothing \citep{Levina,Massoul}. In comparison with this literature, our objective is less ambitious since we only estimate a feature of the graph (its complexity). In return, we carry out a general analysis and do not assume any smoothness condition on $\Omega = [0,1]$. Indeed, our results on the neighborhood distance and covering number estimations are valid for all graphons. For the dimension, we make mild assumptions which are similar to those in the ``intrinsic dimension estimation" literature (see subsection \ref{intro:lit:dimEstim} for a brief description of this related problem).

In the estimation problem of $W$, the latent space $[0,1]$ is sometimes considered instead of $\Omega$. This choice is not restrictive (if no assumption is made on the function $W$ on $[0,1]$) because both settings generate the same W-random graph distributions \citep{lovasz2}. However, the restricted setting $[0,1]$ is not always convenient to work with, whereas the general setting $\Omega$ leads to simpler and cleaner situations \citep{lovasz2}. Indeed, many random graph distributions are naturally represented on $\Omega$ so that their properties are easy to interpret. See Section \ref{section::illustExample} for illustrative examples.

The $l_2$-neighborhood distance \eqref{intro:l2neigh} is a variant of the $l_1$-neighborhood distance introduced by \citet{lovasz2}. This variant has been leveraged several times for the estimation of $[W(\om_i,\om_j)]_{i,j\leq n}$ \citep{Levina,Massoul} where the authors use it as a criterion to select neighborhoods of nodes. Here, our estimator of the $l_2$-neighborhood distance is inspired by the work of \citet{Levina}, as will be discussed later. 

In sparse settings as well, the matrix $[W(\om_i,\om_j)]_{i,j\leq n}$ can be estimated by averaging over "similar" observed entries, where ``similar'' is defined by some neighbor estimator. In \citep{li2019nearest} for example, the neighbor estimator is simply the $l_2$-distance between two rows in the observed matrix, and the random fraction of observed entries is larger than $n^{-1/2+ \de}$ for any $\de >0$. We also consider this sparse regime for the estimation of the $l_2$-neighborhood distance, but we do not use the same estimator as \cite{li2019nearest} since it suffers from a constant bias $O(1)$. For details, see the equations page 7 in \citep{li2019nearest}.

However, the distance estimators in \citep{li2019nearest,Levina} and the current paper are only based on immediate neighborhoods, which are not sufficiently informative in very sparse regimes, where the fraction of observed entries is $o(n^{-1/2})$. Although this is not the focus of our paper, we mention that that there exist successful methods in such very sparse regimes. For example, the similarity between two nodes can be estimated by comparing the two sets of paths starting from these nodes \citep{borgs2017iterative}. By considering paths of length (strictly) larger than $1$, this approach allows to gather information within larger neighborhoods than immediate neighbors. However, the theoretical guarantees proved for this method \citep{borgs2017iterative} are restricted to functions $W$ of finite spectrum and defined on $[0,1]$, whereas we study any function on any space $\Omega$.


\subsubsection{Intrinsic dimension estimation}\label{intro:lit:dimEstim}

There is a considerable body of literature on the estimation of intrinsic dimension of a manifold \citep{kim2016minimax,Kegl,koltchinskii,levina2005maximum}. In the simplest setting, points are sampled on a manifold of $\mathbb{R}^m$ whose dimension is an integer, and the objective is to recover this dimension from the sample. In contrast, here we do not assume the dimension is an integer, we do not observe the $n$ sampled points $\om_1,\ldots,\om_n$, and we are not in the Euclidean metric space $\mathbb{R}^m$. Indeed, the neighborhood distance $r_W$ is unknown, and our only observation is the connections of the graph. \bigskip

\textsc{Outline of the paper.} Section~\ref{section::W_random_graph} gives a formal presentation of the problem. Section~\ref{sect::complexity_index} presents the complexity index and some illustrations. In Section~\ref{sec:est:dist}, we focus on statistical estimation (distance, covering number, dimension). In Section \ref{subsec::test_complexity}, we test the graph complexity. In Section~\ref{section::further_consideration}, we provide two extensions (estimation on sparse graphs, and a polynomial-time algorithm). Proofs are deferred to the appendix. \bigskip

\textsc{Notation.} we write $a \lesssim b$, if there exists a constant $C$ such that $a \leq C b$; and note $a \asymp b$, if there exist two constants $c, c'$ such that $c a \leq b \leq c' a$. We denote by $a\vee b$ (respectively $a\wedge b$) the maximum (resp. minimum) between $a$ and $b$; by $[a]_+$ the maximum between $0$ and $a$; by $[n]$ the set $\{1,\ldots,n\}$; by $B(x,\ep)$ a ball of radius $\ep$ and center $x$. We note $1_{\mathcal{E}}$ the indicator function corresponding to any event $\mathcal{E}$. We write ``a.e." for ``almost everywhere"; and ``w.r.t." for ``with respect to"; and ``w.h.p." for "with high probability", which means that the probability converges to $1$ as the number of graph nodes tends to infinity.


\section{Model}\label{section::W_random_graph}

\subsection{Setting}\label{seubsec:setting}

For a set of vertices $V= \{1,\ldots,n\}$, a \textit{W-random graph} $G =(V,E)$ is generated as follows. Let $(\Omega, \mu, W)$ be an unknown triplet of parameters, which is composed of a measurable set $\Omega$, a probability measure $\mu$ on $\Omega$, and a symmetric (measurable) function $W : \Omega \times \Omega \rightarrow [0,1]$. Such a triplet is called a graphon and we write $\mathcal{W}$ the collection of all graphons. For each node $i\in V$, an unknown attribute $\om_i \in \Omega$ is drawn in an i.i.d. manner from the distribution $\mu$. Conditionally to the attributes $\boldsymbol{\om} = (\om_1,\ldots,\om_n)$, an edge connects two vertices $i$ and $j$, independently of the other edges, with probability $W(\om_i,\om_j)$.
\begin{equation}\label{model::data}
    \mathbb{P}\big{(}(i,j) \in E\,\big{|}\,\boldsymbol{\om} \big{)}= W(\om_i,\om_j)
\end{equation}



Our data are a single observation of the W-random graph. Formally, it is an adjacency matrix $A= [A_{ij}]_{i,j\leq n}$ defined by $A_{ij} = 1$ if $(i,j) \in E$, and $0$ otherwise. This symmetric binary matrix with zero-entries on the diagonal represents an undirected, unweighted graph with no self edges. The distribution of $A$ is called the data distribution and is denoted by  $\mathbb{P}^{(n)}_{(\Omega, \mu, W)}$. The goal will be to define an index characterizing the complexity of the \textit{limiting} distribution $\mathbb{P}^{(\infty)}_{(\Omega, \mu, W)}$. In particular, the index should be identifiable from all distributions $\mathbb{P}^{(n)}_{(\Omega, \mu, W)}$, $1 \leq n \leq \infty$, in order to be estimated from the data $A$. The dependence in $n$ is often dropped out in the rest of the paper. 

\subsection{Non-identifiability and equivalence class of graphons}
\label{prop:lov:equiv}
From the observation $A$, the function $W$ is not identifiable. Indeed, for any measure-preserving bijection $\phi: \Omega \rightarrow \Omega$, we can observe that the map $W^{\phi}(x,y) = W(\phi(x),\phi(y))$ leaves the data distribution unchanged, i.e.: \begin{equation*}
    \mathbb{P}_{(\Omega, \mu, W)} = \mathbb{P}_{(\Omega, \mu, W^{\phi})}.
\end{equation*}In fact, even the latent space $\Omega$ is not identifiable. The full picture is described by \citet[chap.10]{lovasz2}:
\begin{quote}\textit{Two graphons $(\Omega, \mu, W)$ and $(\Omega', \mu', W')$ parametrize the same data distributions for all $n$, if and only if, there exist some measure-preserving maps $\phi : [0,1] \rightarrow \Omega$ and $\psi : [0,1] \rightarrow \Omega'$ such that $W^{\phi}(x,y)= W'^{\psi}(x,y)$ a.e.}\end{quote}
where $[0,1]$ is the probability space endowed with the uniform measure. This characterization will be useful to prove the identifiability of our complexity index. For clarity of this future discussion, we consider the corresponding quotient space $\mathcal{W}/\sim$, which is the set of equivalence classes of graphons leading to the same data distributions.

\section{Complexity index}\label{sect::complexity_index}
 
Given a graphon $(\Omega, \mu, W)$, we endow the latent space $\Omega$ with the \textit{neighborhood distance} \begin{equation}\label{def:rW}
r_W(\om, \om') =  \left( \int_{\Omega} \left| W(\om,\om'') - W(\om',\om'') \right| ^2  \hspace{0.1cm}\mu(d\om'') \right)^{1/2}
\end{equation}which is the $l_2$-norm $||W(\om,.)-W(\om',.)||_{2, \mu}$ between the slices of the function $W$ in $\om$ and $\om'$. Then, we measure the complexity of the pseudo-metric space $(\Omega, r_W)$ in a classic way, using its covering number $\cov(\ep)$ and its Minkowski dimension:
\begin{equation}\label{def:dim}
\di   := \underset{\epsilon \rightarrow 0}{\textup{lim}}\ \, \frac{\log \, \cov(\epsilon)}{- \textup{log} \, \epsilon} 
\end{equation}when the limit exists. See appendix \ref{subsec::additionalInfo} for additional information about these two standard measures of metric spaces.

Unfortunately, the covering number and the Minkowski dimension of a graphon are not identifiable from the data distribution $\mathbb{P}_{(\Omega, \mu, W)}$. Indeed, they are not robust to changes of the graphon on null-sets, whereas such changes leave the data distribution unaltered (a null-set is a set of zero measure in the probability space $(\Omega, \mu)$). This fact is illustrated in the following example where two equivalent graphons (i.e. leading to the same data distributions) have two different Minkowski dimensions. As we can see, this problem is due to the presence of a ``big" null-set in $\Omega$.\smallskip

\noindent \textsc{Example.} Let $\Omega$ := \{2\} and $\Omega'$ : = \{2\} $\sqcup$ $[0, 1]$ be two latent spaces endowed with a common probability distribution $\mu$ such that $\mu[\{2\}] = 1$. Let $W'$ be a function defined on $\Omega'\times \Omega'$ such that $W'(x',y') = (x' + y' )/3$ for $x' , y' \in [0,1]$. Let $W$ be any measurable function on $\Omega \times \Omega$ such that $W(2,2) = W'(2,2)$. Then, the two graphons $(\Omega, \mu, W)$, $(\Omega', \mu, W')$ are equivalent, 
and yet they have two different Minkowski dimensions: $\di = 0$ since $r_{W} = 0$ on $\Omega$, while $\di ' = 1$ since $r_{W'}(x',z') =  |x' -z'|/3$ for $x' , z' \in [0,1]$. $\square$

\subsection{Purification process for identifiability}\label{correction_definition_pure_graphon}
To define an identifiable index of complexity, we need to take care of ``big" null-sets (seen in the above example). Usually, these pathological sets are not present in standard representations $(\Omega, \mu, W)$ and even useless in terms of modeling. Thus, we get rid of them; we do so by using a general remedy, called pure graphon.\medskip

\noindent \textbf{Definition \citep[chap.13]{lovasz2}} \label{thmdef:PURE}\textit{A graphon $(\Omega, \mu, W)$ is called pure if $(\Omega, r_W)$ is a complete separable metric space and the probability measure has full support (that is, every ball of non-zero radius has positive measure). Besides, there is a pure graphon in each equivalence class of graphons.} \medskip

For illustrative examples of pure graphons, see Section \ref{section::illustExample}. There is no ``big" null-set in pure graphons (since their measure $\mu$ has full support by definition) and the complexity index takes the same value on the pure graphons of a same equivalence class of $\mathcal{W}/\sim$ (Lemma \ref{inv:thm}). \begin{lemma}\label{inv:thm}If two pure graphons are equivalent, then their covering numbers are equal.\end{lemma}

The proof of Lemma \ref{inv:thm} is written in Appendix~\ref{appB}.
Lemma \ref{inv:thm} directly implies that the Minkowski dimension takes the same value for two equivalent pure graphons. Thus, for any limiting W-random graph distribution $\mathbb{P}^{(\infty)}_{(\Omega, \mu, W)}$, we define its complexity as the covering number and the Minkowski dimension of any pure graphon from the corresponding equivalence class. According to the above lemma, these indices are identifiable from all data distributions $\mathbb{P}^{(n)}_{(\Omega, \mu, W)}$, $1 \leq n \leq \infty$. From now on, we can work exclusively with pure graphons without the loss of generality, since there are pure graphons in each equivalence class of $\mathcal{W}/\sim$. In the remaining of the subsection, we describe two consequences of working with pure graphons. \medskip

The metric properties are preserved between equivalent pure graphons (Lemma \ref{lem:inv:dist}). 
\begin{lemma} \label{lem:inv:dist}Let $(\Omega, \mu, W)$ and $(\Omega', \mu', W')$ be two pure graphons, endowed with their respective neighborhood distances $r_W$ and $r_{W'}$. If the two graphons are in a same equivalence class of $\mathcal{W}/\sim$, then for some bijective measure-preserving map $\phi: \Omega' \rightarrow \Omega$, we have
\begin{equation*}r_{W'}\left(x, y\right) = r_W\left(\phi(x), \phi(y)\right) \hspace{0.3cm}almost\ \,  surely \ \, on \ \, \Omega' \times \Omega'.\end{equation*}
\end{lemma}

Lemma~\ref{lem:inv:dist} states that the metric spaces $(\Omega, r_W)$ and $(\Omega',r_{W'})$ are isometric up to a null-set, it is therefore not surprising that they share the same covering number (Lemma~\ref{inv:thm}). The proof of lemma \ref{lem:inv:dist} is written in Appendix \ref{proof:lem:inv:dist}. Note that Lemma~\ref{lem:inv:dist} ensures that the future distance estimation is a well-posed problem.

Another consequence of working with pure graphon is that the sample $\om_1,\ldots,\om_n$ is asymptotically dense in $\Omega$. Lemma \ref{sample::asymptoticConsistency} is proved in Appendix \ref{proof:sample:asymptoticDense}. 
\begin{lemma}\label{sample::asymptoticConsistency}
For a pure graphon $(\Omega, \mu, W)$ such that $\cov(\ep) < \infty$ for all $\ep > 0$, the sample $\om_1,\ldots,\om_n$ is asymptotically dense in the metric space $(\Omega, r_W)$. That is, for all radii $\ep >0$, the event
\begin{equation*}
    \mathcal{E}(\ep) = \{\textup{each ball of radius } \ep \textup{ in } (\Omega, r_W) \textup{ contains at least a sampled point } \om_i\}
\end{equation*}
holds with a probability tending to one as $n \rightarrow \infty$. 
\end{lemma}

\subsection{Illustrative examples}
\label{section::illustExample}

We exemplify the complexity index with instances of W-random graphs that are often considered in the literature: a stochastic block model \citep{1983stochastic, abbe}, a random H\"older graph \citep{gao,Levina} and a random geometric graph \citep{penrose,ariasVerzeGG,deCast,bubek}.\smallskip




\textsc{Stochastic Block Model.} It produces a structure of community dividing the node set into $K$ subsets of nodes which share a same pattern of connection. More precisely, the edges are independently sampled from each others, and the probability of an edge between two nodes only depends on their community membership. The SBM with $K$ communities can be written in the framework of the W-random graph model, by setting $\Omega = \{c_1, \ldots, c_K\}$, so that each node belongs to one of the $K$ communities $c_i$, and connects to each other with probability $W(c_i,c_j)$. A natural notion of complexity for SBM is the number $K$ of communities, which coincides with the $\ep$-covering number of $\{c_1, \ldots, c_K\}$ for small radii $\ep$. \smallskip



\textsc{Approximation by SBM.} In the estimation of $W$ based on the classic approximation by SBM \citep{gao,klopp}, the right number of communities can be selected using the covering number. Indeed, Proposition \ref{prop:approxSBM} states that, for any graphon $(\Omega, \mu, W)$, the function $W$ can be ``$O(\epsilon)$-approximated" in $l_2$-norm by an SBM with at most $\cov(\epsilon)$ communities. The proof is written in Appendix~\ref{appendix::approxSBM}. \begin{proposition}\label{prop:approxSBM}
Consider any graphon $(\Omega, \mu, W)$ and its $\ep$-covering number $\cov(\epsilon)$, defined in Section~\ref{sect::complexity_index}. There exists a graphon $(\Omega, \mu, \overline{W})$ equivalent to an SBM with $\cov(\epsilon)$ communities, such that, 
$$\int_{\Omega^2} (W(\om,\om') - \overline{W}(\om,\om'))^2\mu(d\om)\mu(d\om') \leq (4 \ep)^2.$$  
\end{proposition} 

\textsc{Random H\"older graph.} Let $\Omega =[0,1]^d$ be endowed with the uniform measure, and $W$ fulfill a double H\"older condition:  \begin{equation}\label{doubleHoldercond}  m \big{|}\big{|}\om'-\om\big{|}\big{|}_2^{\alpha} \leq \big{|}W(\om',\om'')-W(\om,\om'')\big{|} \leq M \big{|}\big{|}\om'-\om\big{|}\big{|}_2^{\alpha}\end{equation} for some H\"older exponent $\alpha > 0$ (and some constants $m, M > 0$). This means that each node has its specific attribute of $d$ variables, and connects to another node with a probability that smoothly depends on the node attributes. A natural notion of complexity for this graph distribution should increase with the number $d$ of variables, and decrease with the level $\alpha$ of smoothness. This intuitive notion is matched by the Minkowski dimension, which is equal to $d/\alpha$. See Appendix \ref{subsec::additionalInfo} for details.\smallskip


\textsc{Random geometric graph.} It generates simple spatial networks placing nodes in a Euclidean metric space and connecting two nodes if their Euclidean distance is small. Let $\Omega = [0,1]^d$ be endowed with the uniform measure and the indicator function $W(\om,\om') = \mathbb{I}_{||\om-\om'||_2 \leq \delta}$ for some constant $\delta >0$. Appendix \ref{subsec::additionalInfo} shows that $\di = 2 d$. Thus, the Minkowski dimension matches the Euclidean dimension of the latent space, up to a factor $2$.  

\section{Estimation of the complexity index}
\label{sec:est:dist}

Given a pure graphon $(\Omega, \mu, W)$, assume a W-random graph is generated from the probability distribution $\mathbb{P}_{(\Omega, \mu, W)}$ defined in Section \ref{seubsec:setting}. From a single observation of the adjacency matrix $A$ of this graph, we want to estimate the complexity index (introduced in Section \ref{correction_definition_pure_graphon}). In particular, the underlying graphon $(\Omega, \mu, W)$ is unknown, and the sampled points $\om_1,\ldots,\om_n$ are not observed. 

This section is organized in the following manner. We first estimate the neighborhood distance \eqref{def:rW} on the sampled points $\om_1,\ldots,\om_n$. Based on these estimated distances, we then estimate the $\ep$-covering number of $(\{\om_1,\ldots,\om_n\},r_W)$ by plug-in. Denote by $\covh(\ep)$ this estimator. We finally estimate the Minkowski dimension using $- \log \, \covh(\ep)\big{/} \log \,\ep $ at a well chosen radius $\ep$. 



\subsection{Distance-estimator}\label{subsec:def:disEstim}
Let us explain the construction of the distance estimator. The $l_2$-neighborhood distance is naturally associated with a structure of inner product. Given some square-integrable functions $f$ and $g$ on $\Omega$, we write their inner product  $\langle f, g\rangle := \int_{\Omega}f(z)g(z) \mu (dz)$. Let $W(\omega_{i},.)$ denote the function $x \mapsto W(\omega_{i},x)$, then the neighborhood distance admits the following decomposition
\begin{eqnarray}
\label{def:dist:quad}r_W^{2} (\om_i, \om_j) &=& \langle W(\omega_{i},.),W(\omega_{i},.)\rangle + \langle W(\omega_{j},.),W(\omega_{j},.)\rangle - 2\langle W(\omega_{i},.),W(\omega_{j},.)\rangle.
\end{eqnarray}
We estimate separately the crossed term and the two quadratic terms of (\ref{def:dist:quad}).

Note $A_{i}$ the $i^{\textup{th}}$ row vector of the adjacency matrix $A$, and $\langle A_{i}, A_{j} \rangle_n $ = $\sum_{i=1}^n A_{ik}A_{jk}/n$ the inner product between two such rows. Given $\om_i, \om_j$, we observe that $\langle A_{i}, A_{j} \rangle_n $ is (almost) a sum of i.i.d. random variables (up to a duplicated entry because of the symmetry of the adjacency matrix $A$). Indeed, the $n-2$ random variables $\{A_{ik}A_{jk}:\, k \in [n] \textup{ and } k \neq i,j\}$ are independent with the same mean conditionally to $\om_i, \om_j$ : \begin{equation*}\mathbb{E}\left[A_{ik}A_{jk}|\om_i,\om_j\right] = \langle W(\omega_{i},.),W(\omega_{j},.)\rangle\end{equation*} where the mean $\mathbb{E}$ is taken over the data distribution $\mathbb{P}_{(\Omega, \mu, W)}$. It is therefore possible to use Hoeffding's inequality to prove that $\left|  \langle A_{i}, A_{j} \rangle_n - \langle W(\omega_{i},.),W(\omega_{j},.)\rangle \,\right|  \lesssim \sqrt{\textup{log}\hspace{0.1cm}n / n}$ w.h.p. (see Proposition \ref{prop:dem:crossTerm} in Appendix \ref{proof:thm:dist:uppBound}). Thus, the inner product between two \textit{different} rows is a consistent estimator of the crossed term $\langle W(\omega_{i},.),W(\omega_{j},.)\rangle$ in (\ref{def:dist:quad}). 

To estimate the remaining quadratic term $\langle W(\omega_{i},.),W(\omega_{i},.)\rangle$ in (\ref{def:dist:quad}), we cannot proceed in the same way since $\frac{1}{n} \langle A_{i},A_{i} \rangle$ is an inconsistent estimator of $\langle W(\omega_{i},.),W(\omega_{i},.)\rangle$; indeed, we have \begin{equation*}
    \mathbb{E} \big{[}A_{ik}A_{ik}| \om_i\big{]}= \mathbb{E} \big{[}A_{ik}| \om_i\big{]} =\langle W(\omega_{i},.), 1\rangle \neq \langle W(\omega_{i},.),W(\omega_{i},.)\rangle.
\end{equation*} To work around this issue, we simply approximate the quadratic term by a crossed term to be back to the previous case. Specifically, the approximation consists in replacing a sampled point by its nearest neighbor as follows: let $\om_{m(i)}\in \{\om_1,\ldots,\om_n\}$ denote a nearest neighbor of $\om_i$ according to the distance $r_W$, that is $m(i) \in \mbox{argmin}_{t :\, t \neq i} \ \, r_W(\om_i, \om_t)$, then we have the following approximation: 
\begin{align}\label{neighbError}
|\big{\langle} W(\om_i,.),W(\om_i,.) \big{\rangle} - \big{\langle} W(\om_i,.),W(\om_{m(i)},.) \big{\rangle}| &=  |\big{\langle} W(\om_i,.), W(\om_i,.)-W(\om_{m(i)},.)\big{\rangle}| \nonumber \\
&\leq  r_W(\om_i,\om_{m(i)})  
\end{align}
using Cauchy-Schwarz inequality. Thus, the nearest neighbor approximation (\ref{neighbError}) entails a bias in our estimation procedure, which is equal to the distance between $\om_i$ and its nearest neighbor $\om_{m(i)}$. 

Since the index $m(i)$ is unknown, we define an index estimator $\widehat{m}(i)$ such that $\om_{\widehat{m}(i)}$ is hopefully close to $\om_i$ according to $r_W$, and then we use $\langle A_{i} , A_{\widehat{m}(i)} \rangle_n$ to estimate the quadratic term. Formally, $\widehat{m}(i)$ is a minimizer of the distance function $j \mapsto \widehat{f}(i,j)$ defined by  \begin{equation}\label{minim}\widehat{f}(i,j) = \underset{k: \, k \neq i, j}{\mbox{max}}\left| \langle A_{k},A_{i}-A_{j}\rangle_n \right|
\end{equation}where $\widehat{f}(i,j)$ represents a proxy for the distance between the $i^{\textup{th}}$ and $j^{\textup{th}}$ rows of the adjacency matrix, which is enough to define the index estimator \begin{equation}\label{def:indexEstim}
    \widehat{m}(i) = \underset{j: \, j\neq i}{\textup{argmin}} \widehat{f}(i,j).
\end{equation}Note that $\widehat{f}(i,j)$ is small in expectation if $\om_i$ and $\om_j$ are close according to the neighborhood distance; indeed, $\mathbb{E} \big{[}\,\widehat{f}(i, j) | \om_i, \om_j, \om_k \big{]} = \textup{max}_{k \neq i,j} \,|\big{\langle} W(\om_i,.)-W(\om_j,.),W(\om_k,.) \big{\rangle}| \leq r_W(\om_i,\om_j)$ using Cauchy-Schwarz inequality.

Putting together the estimators of the crossed term and the two quadratic terms, we get the following estimator of the square distance  $r_W^2(\omega_i,\omega_j)$: 
\begin{equation}\label{def:dist:estimator}
 \widehat{r}^{2}(i , j) = \langle A_{i} , A_{\widehat{m}(i)} \rangle_n + \langle A_{j} , A_{\widehat{m}(j)} \rangle_n - 2\, \langle A_{i} , A_{j} \rangle_n 
\end{equation}for all $i,j \in [n]$, where $\widehat{m}(i)$ is given by \eqref{def:indexEstim}. \bigskip

\textsc{Remark: }The distance-estimator (\ref{def:dist:estimator}) is inspired by the work of \citet{Levina}, in which the authors want to recover the expectation of the adjacency matrix $A$, based on neighborhood smoothing. They rely on the proxy \eqref{minim} to select neighborhood of points with respect to the neighborhood distance. Restricting themselves on graphons of the form $([0,1], \lambda, W)$ with $\lambda$ the uniform measure and $W$ a piecewise Lipschitz function, they derive risk bounds for the estimation of $W.$ In contrast, here we do not make any assumption on the graphon, and our objective is to provide an estimator of the neighborhood distance per se.  

\subsection{Consistency of the distance-estimator}\label{subsec::consitencyDistestim}
The statistical recovery of the set of distances $\{r_W(\omega_i,\omega_j):  i,j\in [n]\}$ is a well-posed problem, since the neighborhood distance is invariant on each equivalence class of graphons (Lemma~\ref{lem:inv:dist}). Theorem \ref{thm:dist:upperBound} gives non-asymptotic error bounds for the distance-estimator \eqref{def:dist:estimator}. The proof is written in Appendix \ref{proof:thm:dist:uppBound}.
\begin{theorem}
\label{thm:dist:upperBound}
Given any (pure) graphon $(\Omega, \mu, W)$, consider the data distribution $\mathbb{P}_{(\Omega, \mu, W)}$ defined in model (\ref{model::data}).
For all $1 \leq i \leq n,$ let $\om_{m(i)} \in \{\om_1,\ldots,\om_n\}\setminus \{\om_i\}$ denote a nearest neighbor of $\om_i$ according to the distance $r_W$. Then, for the distance-estimator (\ref{def:dist:estimator}), the event
\begin{equation*}
\mathcal{E}_{dist}=\left\{\forall i, j \in[n]: \left| r_W^2(\om_i,\om_j) - \widehat{r}^2( i, j) \right| \leq  3 r_W(\om_j,\om_{m(j)}\,) + 3 r_W(\om_i,\om_{m(i)}\,) + 36 \sqrt{\textup{log}(n)/n }\right\} \end{equation*}holds with probability $\mathbb{P}_{(\Omega, \mu, W)}[\mathcal{E}_{dist}] \geq 1 - \frac{2}{n}.$
\end{theorem}

Theorem \ref{thm:dist:upperBound} implies that the distance-estimator (\ref{def:dist:estimator}) is a consistent estimator of the neighborhood distance \eqref{def:rW}, provided that the $\ep$-covering number is finite for all radii $\ep >0$. Indeed, for a finite covering number, Lemma \ref{sample::asymptoticConsistency} ensures that the sample $\om_1,\ldots,\om_n$ is asymptotically dense in $(\Omega, r_W)$, which implies that the bias $r_W(\om_i,\om_{m(i)}\,)$ is convergent in probability to zero as $n$ grows to infinity.

Let us describe the upper bound of Theorem \ref{thm:dist:upperBound}. On the one hand, there is a fluctuation term $\sqrt{\mbox{log}(n)/n}$ that corresponds to the convergence property of the inner products between rows of $A$, i.e.: $\left|  \langle A_{i}, A_{j} \rangle_n - \langle W(\omega_{i},.),W(\omega_{j},.)\rangle \,\right|$ $\lesssim \sqrt{\textup{log}\hspace{0.1cm}n / n}$ w.h.p. for $i \neq j$. On the other hand, there is a bias term $r_W(\om_i,\om_{m(i)}\,)$ that results from the nearest neighbor approximation (\ref{neighbError}). Its value depends on the graphon regularity. For instance, in the SBM example of Section \ref{section::illustExample}, the bias term $r_W(\om_i,\om_{m(i)}\,)$ is equal to zero w.h.p. (indeed, $\om_i$ and its nearest neighbor $\om_{m(i)}$ are in the same community w.h.p., and thus separated by a distance zero w.r.t. $r_W$). In the random H\"older graph example, the bias term is of the order of $(\textup{log}( n) / n)^{\alpha/d}$ w.h.p.. \medskip

The next result gives a lower bound matching the upper bound of Theorem \ref{thm:dist:upperBound}, up to a numerical constant. Specifically, there exists a graphon $(\Omega, \mu, W_n)$ for each sample size $n \geq 1$, such that the lower bound holds for (at least) some of the $O(n^2)$ distances. 

\begin{theorem}\label{new:thm:clari}
There exists a sequence of graphons $(\Omega, \mu, W_n)_{n \in \mathbb{N}}$ and some numerical constants $p,c,c' > 0$, such that the following holds for any estimator $\widehat{d}$ and any permutation $\sigma$ of the $n$ indices. With a probability larger than $p$, the lower bound 
\begin{equation*}
\left| r_{W_n}^2(\om_i,\om_j) - \widehat{d}^2( \sigma(i),\sigma(j)) \right|  \geq c' \left( r_{W_n}(\om_j,\om_{m(j)}\,) + r_{W_n}(\om_i,\om_{m(i)}\,) + \sqrt{\frac{\textup{log}\, n }{n} }\right)\end{equation*}\smallskip
 
\noindent is satisfied for (at least) $c\,n$ different pairs $(i,j)$.

\end{theorem}

The lower bound in Theorem \ref{new:thm:clari} holds regardless of the nodes labels $i\in \{1,\ldots,n\}$ since it is satisfied for any permutation $\sigma$ of the $n$ indices $\{1,\ldots,n\}$. This is relevant in the model \eqref{model::data} where the data distribution is invariant by relabeling of the nodes.  
\medskip

In particular, Theorem \ref{new:thm:clari} can be written for a graphon $(\Omega, \mu, W_n)_{n}$ whose bias $r_{W_n}(\om_i,\om_{m(i)}\,)$ is equal to a numerical constant, thus giving the next result.

\begin{theorem}\label{new:coro:clari}
There exists a sequence of graphons $(\Omega, \mu, W_n)_{n \in \mathbb{N}}$ and some numerical constants $p,c > 0$, such that the following holds for any estimator $\widehat{d}$ and any permutation $\sigma$ of the $n$ indices. With a probability larger than $p$, the lower bound 
\begin{equation*}
\left| r_{W_n}^2(\om_i,\om_j) - \widehat{d}^2( \sigma(i),\sigma(j)) \right|  \geq \frac{1}{800}\end{equation*}\smallskip
 
\noindent is satisfied for (at least) $c\,n$ different pairs $(i,j)$.

\end{theorem}

The constant error in this lower bound does not violate the vanishing error bound in Theorem~\ref{thm:dist:upperBound}. Indeed, the graphon changes with $n$ in the above lower bounds whereas it remains fixed in the upper bound of Theorem~\ref{thm:dist:upperBound}.

\medskip

Although Theorem \ref{new:thm:clari} is a standard way of stating lower bounds in the literature, it is not sufficient to show that the right dependence on the bias $r_W(\om_i,\om_{m(i)}\,)$ is linear. For example, if $r_W(\om_i,\om_{m(i)}\,)$ is equal to a numerical constant, then one can replace $r_W(\om_i,\om_{m(i)}\,)$ with $r_W^2(\om_i,\om_{m(i)}\,)$ in Theorem \ref{new:thm:clari} (by adapting the numerical constant $c'$ which was not tight anyway). One can observe that such a change from $r_W(\om_i,\om_{m(i)}\,)$ to $r_W^2(\om_i,\om_{m(i)}\,)$ is also possible in the case where $r_W(\om_i,\om_{m(i)}\,)$ is smaller than the fluctuation term $\sqrt{\log(n)/n}$. Therefore, neither the lower bound \eqref{eq:LB:dist} nor the lower bound \eqref{eq:LB:dist:cst} is enough informative to decipher whether the optimal rate is of the order of $r^2_W(\om_i,\om_{m(i)}\,) + \sqrt{\log(n)/n}$ or $r_W(\om_i,\om_{m(i)}\,)+\sqrt{\log(n)/n}$.

A quadratic dependency $r_W^2(\om_i,\om_{m(i)}\,)$ would improve on the linear dependency since the neighborhood distance $r_W$ is always smaller than $1$ by definition. Hence, one may wonder whether such an improvement is possible, especially that Theorem~\ref{thm:dist:upperBound} is an error bound on \textit{square} distances, $\left| r_{W}^2(\om_i,\om_j) - \widehat{d}^2(i,j) \right|$. It turns out that even replacing the bias $r_W(\om_i,\om_{m(i)}\,)$ with $r^{1+\gamma}_W(\om_i,\om_{m(i)}$) for some $\gamma>0$ is impossible. In other words, no estimator $\widehat{d}$ simultaneously satisfies the following inequalities, with high probability, over all graphons $(\Omega, \mu, W)$:  \\
$\forall i , j \in [n]:$
\begin{equation}\label{ineq::imp::illus}
  \left| r_{W}^2(\om_i,\om_j) - \widehat{d}^2(i,j) \right| \leq C \left( r^{1+\gamma}_{W}(\om_j,\om_{m(j)}\,) + r^{1+\gamma}_{W}(\om_i,\om_{m(i)}\,) + \sqrt{\textup{log}(n)/n }\right)  
\end{equation}where $C$ is a numerical constant. Indeed, Theorem \ref{thm:simple:minimax} ensures that the uniform bound \eqref{ineq::imp::illus} cannot be achieved by any estimator $\widehat{d}$ as soon as $\gamma >0$, thus proving that the bias-dependence is linear $(\gamma =0$) at best. 

\begin{theorem}\label{thm:simple:minimax}There exists a sequence of graphons $(\Omega, \mu, W_n)_{n \in \mathbb{N}}$ and some numerical constants $p > 0$ , $\gamma \geq 0$ and $c > 0$, such that the following holds for any estimator $\widehat{d}$ and any permutation $\sigma$ of the $n$ indices. With a probability larger than $p$, the following lower bound is satisfied for (at least) $c\,n$ different pairs $(i,j)$,
\begin{equation*}
\left| r_{W_n}^2(\om_i,\om_j) - \widehat{d}^2( \sigma(i),\sigma(j)) \right| \gtrsim \kappa_n(\gamma) \left( r^{1+\gamma}_{W_n}(\om_j,\om_{m(j)}\,) + r^{1+\gamma}_{W_n}(\om_i,\om_{m(i)}\,) + \sqrt{\frac{\textup{log}\, n}{n} } \right) \end{equation*}\smallskip
 
\noindent  where $\kappa_n(\gamma) \underset{n}{\longrightarrow} \infty $ as soon as $\gamma >0$, and $\kappa_n(0) = O(1)$.
\end{theorem}

Theorem \ref{new:thm:clari} is a direct consequence of Theorem \ref{thm:simple:minimax} for $\gamma =0.$ The proof of Theorem \ref{thm:simple:minimax} can be found in Appendix \ref{proof:lowerboundDist}. 

In conclusion, the upper bounds in Theorem \ref{thm:dist:upperBound} have the right dependence on the bias.

\subsubsection{Discussion about the distance estimation and open questions }

The lower bound in Theorem \ref{new:thm:clari} only holds for $O(n)$ distances among the $O(n^2)$ ones. As to whether this could be extended to $O(n^2)$ distances, the question is left open. This generalization would confirm that the dependence on the bias should be linear and thus that the upper bounds of Theorem \ref{thm:dist:upperBound} have the optimal bias-dependence.

This gap may come from technical shortcomings of our proof, rather than an inherent feature of the estimation problem. Indeed, the instance of W-random graph constructed for the lower bounds, has only one outlier node whose $O(n)$ distances are incorrectly estimated. A possible difficulty for generalizing the proof to $O(n^2)$ distances is the randomness of the sample $\om_1,\ldots,\om_n \overset{iid}{\sim} \mu$ and the symmetry of the matrix $A$, as they bring regularization effects to the problem.

Another element suggesting that the lower bounds in Theorem \ref{new:thm:clari} may not be specific to one outlier, is that the upper bounds in Theorem \ref{thm:dist:upperBound} are not sensitive to $O(1)$ nodes. Indeed, by reading the proof of Theorem \ref{thm:dist:upperBound}, one can observe that the upper bounds will not improve after removing $O(1)$ nodes. 

Note that our lower bound has probably no direct consequences on graphon estimation since the bound only hinges on $O(1)$ outlier nodes, which would be negligible in the estimation of the probabilities matrix $[W(\om_i,\om_j)]_{1 \leq i , j \leq n}$ with respect to the Frobenius norm.

The fluctuation term $f_n:=\sqrt{\log(n)/n}$ in the upper bounds of Theorem \ref{thm:dist:upperBound} is natural, as each inner product is computed with $\Theta(n)$ entries, leading to an error $1/\sqrt{n}$, and the $\sqrt{\log(n)}$ allows us to show high probability bounds for each of the $\theta(n^2)$ distances. However, this explanation is specific to our estimator, and the optimality of $f_n$ with respect to all estimators remains to be proved. The fact that $f_n$ appears in the lower bound of Theorem \ref{new:thm:clari} does not prove that $f_n$ is inevitable since $f_n$ is actually negligible compared to the bias term $r_{W_n}(\om_i,\om_{m(i)})$ in this lower bound. Indeed, Theorem \ref{new:thm:clari} is based on an instance of graphon $W_n$ satisfying $r_{W_n}(\om_i,\om_{m(i)}) \gtrsim f_n$.


\subsection{Consistency of the covering number estimator}\label{subsection::theoreticGuaranteeCovNUmb} 


We have defined the $\ep$-covering number estimator $\covh(\ep)$ as the covering number of the set $\{1,\ldots,n\}$ w.r.t. the distance-estimator $\widehat{r}$. Consider $e_{sup}$ the supremum of the errors of $\widehat{r}$ : \begin{equation*} 
e_{sup} := \underset{i, j \in [n]}{\textup{sup}} \, \left| r_W(\om_i,\om_j) - \widehat{r}( i, j) \right|.\end{equation*}
Then, the covering number estimator is linked with the true covering number of $\{\om_1,\ldots,\om_n\}$ by the following inequalities
\begin{equation*}
\label{cov}
\forall \ep > e_{sup}, \hspace{1cm} \covs \left(\epsilon + e_{sup}\right) \leq \covh\left(\epsilon\right) \leq  \covs \left(\ep -e_{sup}\right).
\end{equation*}

To compare the covering numbers of $\{\om_1,\ldots,\om_n\}$ and $\Omega$, we need to measure the difference between the sample $\om_1,\ldots,\om_n$ and the space $\Omega$. We do so by introducing the sampling error $s_{\om}$ defined as
\begin{equation}\label{def:error:samp}
s_{\om} = \underset{{\om\in \Omega}}{\textup{sup}} \ \,\underset{i \in \{1,\ldots,n\}}{\textup{inf}}\ \, r_W(\om,\omega_i)
\end{equation}which is the greatest distance that separates a point of $\Omega$ from the set $\{\om_1,\ldots,\om_n\}$. Thus, the covering numbers (w.r.t. the true distance $r_W$) of $\om_1,\ldots,\om_n$ and $\Omega$ are linked by the following inequalities \begin{equation*}
\forall \ep > s_{\omega}, \hspace{1cm} \cov\left(\epsilon + s_{\om}\right) \leq \covs\left(\epsilon\right)  \leq \cov\left(\epsilon - s_{\om}\right).    
\end{equation*}

Finally, for \begin{equation} \label{dist:sup} b_{sup}^2 := 6\, \underset{i\in[n]}{\textup{sup}} \, r_W(\om_i,\om_{m(i)}\,) + 36 \, \sqrt{\textup{log}(n)/n }, \end{equation}Theorem \ref{thm:dist:upperBound} ensures that $e_{sup} \leq b_{sup}$ with probability at least $1-2/n$. From the above displays, we obtain the following proposition. 
\begin{proposition}
\label{thm:cov:true}Given any (pure) graphon $(\Omega, \mu, W)$, consider the data distribution $\mathbb{P}_{(\Omega, \mu, W)}$ defined in model (\ref{model::data}). Let $b_{sup}$ and $s_{\om}$ be the distance error bound \eqref{dist:sup} and the sampling error (\ref{def:error:samp}). Then, the estimator $\covh$ satisfies the following non-asymptotic bounds  
\begin{equation}\label{explic:inequ:cov:prop}
\forall \ep > b_{sup} + s_{\om}, \hspace{1cm} \cov\left(\epsilon + b_{sup} + s_{\om}\right) \leq\, \covh\left(\epsilon\right) \, \leq \cov\left(\epsilon - b_{sup} - s_{\om}\right)    
\end{equation}with probability at least $1-\frac{2}{n}$ according to the distribution $\mathbb{P}_{(\Omega, \mu, W)}$.
\end{proposition}

As a result, we have a consistent estimation of the $\ep$-covering number for almost every $\ep$, provided that the covering number is finite for all radii. Indeed, if $\cov(\ep) < \infty$ for all $\ep >0$, then the sample $\om_1,\ldots\om_n$ is asymptotically dense in $(\Omega, r_W)$ by Lemma \ref{sample::asymptoticConsistency}, which implies that $b_{sup}$ and $s_{\om}$ converge in probability to zero; Then, taking the limit $n \rightarrow \infty$ in \eqref{explic:inequ:cov:prop}, one has the convergence in probability of $\covh$ towards $\cov(\ep)$,   for all $\ep$ where the step function $\ep \mapsto \cov(\epsilon)$ is continuous (i.e., for almost every $\ep$).


\subsection{Consistency of the dimension estimator}
\label{section:estimDim} 

We estimate the Minkowski dimension of $(\Omega, r_W)$ using the data-function $- \log \, \covh(\ep)\big{/} \log \,\ep $ at a well chosen radius $\ep$. The following observation makes it clear that each graphon requires a specific choice of radius, and thus no (universal) radius is suited for all graphons. 

\begin{quote}
\textsc{Observation.} (1) at very small scale (i.e. very small $\ep$), the covering number may just count the points of the sample $\om_1,\ldots,\om_n$ and the data look zero-dimensional; (2) if the scale is comparable to the noise due to the distance estimation, the covering number estimator $\covh(\ep)$ is not reliable; (3) for an intermediate scale, it is possible to have a good estimation of the dimension, as we shall see in Theorem \ref{coro:asympDim}; (4) at very big scale, the apparent geometry may not reflect the Minkowski dimension (which is, by definition, a measure of the complexity at infinitesimal scale). 
\end{quote}

Hence, we consider a subset of graphons for which there exists a radius that is well-suited for dimension estimation. We sometimes denote $\di$ by $d$ for brevity, and write $B(\om,\ep)$ the ball of center $\om\in \Omega$ with radius $\ep$ (w.r.t. the neighborhood distance). Given constants $D, v, \alpha> 0$ and $M \geq 1 \geq m > 0$ , we define the set $\mathcal{W}(D,\alpha,m,M,v)$ of all (pure) graphons $(\Omega, \mu, W)$ satisfying
\begin{enumerate}
    \item $\di \leq D$. 
    \item For $\di := d$ and all $\ep \in (0,v]$,
    \begin{equation}\label{ass1}
        \tag{$H_1^{\alpha,v}$}\alpha \,\ep^d \ \, \leq \ \,\mu\, \Big{[}B(\om,\ep)\Big{]}
    \end{equation}
 \begin{equation}\label{ass2}
       \tag{$H_2^{m,M,v}$}
m \ep^{-d} \ \, \leq \ \, \cov(\ep)\ \, \leq \ \, M \ep^{-d}.
    \end{equation}
\end{enumerate}
The assumption \ref{ass2} links the covering number with the Minkowski dimension of the graphon. The condition \ref{ass1} enforces a minimal measure for each ball of $(\Omega, r_W)$; in particular, it strengthens the non-zero measure of balls of pure graphons, seen in Section \ref{thmdef:PURE}. Mention can be made of the problem of recovery of the dimension of a manifold, where similar hypotheses are often considered \citep[see][for example]{koltchinskii}. Besides, \ref{ass1} may be seen as a small-ball condition used in learning problems \citep{mendelson2014learning,lecue2018regularization}. 

\smallskip

With the radius \begin{equation}\label{def:good:radius}
\ep_{D} \asymp  \left(\frac{\textup{log}\, n}{n}\right)^{1/(4 \vee 2 D)}
\end{equation}we consistently estimate the Minkowski dimension (Theorem~\ref{coro:asympDim}) using the following estimator
\begin{equation}\label{estimateur:dim}
    \widehat{dim}_D := \frac{\textup{log} \, \covh(\epsilon_{D})}{-\textup{log}\,\epsilon_{D}}.
\end{equation}

\smallskip

\begin{theorem}\label{coro:asympDim} For all graphons $(\Omega, \mu, W)$ in $\mathcal{W}(D,\alpha,m,M,v)$ and all large enough $n$, we have

\begin{equation*} \left|\, \widehat{dim}_D  - \di \,\right| \leq \ \,\frac{C(D,\alpha,m,M)}{\textup{log}\, n} \end{equation*}\smallskip

\noindent with probability at least $1-C'(\alpha,M)/n$ w.r.t. the distribution $\mathbb{P}_{(\Omega, \mu, W)}$, and for some constants $C'(\alpha,M)$ and $C(D,\alpha,m,M)$ that are independent of $n$.
\end{theorem}

Theorem \ref{coro:asympDim} is a corollary of Theorem \ref{thm:dimMink} in Appendix \ref{appen:proof:thm:dimMink}, which gives a non-asymptotic high probability bound for  $-\textup{log} \, \covh(\epsilon)\big{/} \textup{log}\,\epsilon$  at any radius $\ep$. 

\smallskip  

One can observe that the convergence rate $\textup{log}^{-1}\, n$ of Theorem~\ref{coro:asympDim} is optimal, in the sense that faster convergence rates cannot be achieved by any estimator of the form $\textup{log} \, \covh(\hat{\epsilon})/-\textup{log}\,\hat{\epsilon}$ \footnote{where $\covh$ is any consistent estimator of the covering number, and $\hat{\ep}$ is any estimator of a ``well chosen radius"}. To see it, take a graphon of dimension $d > 1$ with covering number $\cov(\epsilon) = m \ep^{-d}$ for some constant $m>1$. Even if there exists a covering number estimator that gives a perfect estimation, i.e. $\covh = \cov$, this still entails an error for the dimension estimation. Indeed, in such a case we have:  \begin{equation*}\left|\, \frac{\textup{log} \, \covh(\epsilon)}{-\textup{log}\,\epsilon} - d \,\right| = \frac{ \textup{log} \, m }{-\textup{log}\,\epsilon}
\end{equation*}which is (at least) of the order $\textup{log}^{-1}\, n$ since the radius $\ep$ cannot be taken smaller than $n^{-1}$ in general (otherwise, the estimator of the covering number may just count the $n$ sampled points). Thus, the convergence rate $\textup{log}^{-1}\, n$ is optimal for the classical method of estimation of the Minkowski dimension, which is based on the the plug-in of a covering number estimate into formula \eqref{def:dim}. \bigskip

Next we show that no estimator \footnote{defined as a function of the adjacency matrix $A \in \{0,1\}^{n \times n}$.} can improve on the error bound $\log^{-1}n$, over the following sequence of sets. Given $n> 0$, let $\mathcal{W}_{n}(D,\alpha,m,M,v)$ be the class of all (pure) graphons fulfilling \ref{ass1} and \ref{ass2} for all $\ep  \in (1/n, v]$ instead of $\ep \in (0, v],$ so $\mathcal{W}_{n}(D,\alpha,m,M,v)$ is a subset of $\mathcal{W}(D,\alpha,m,M,v)$. On this sequence of sets, one can readily extend Theorem \ref{coro:asympDim} and retrieve the same error bound, using the same estimator \eqref{estimateur:dim}. This means that there exist some constants $C(D,\alpha,m,M)$ and $C'(\alpha,M)$ that are independent of $n$, such that for all graphons in $\mathcal{W}_n(D,\alpha,m,M,v)$ and all large enough $n$, the following error bound holds
\begin{equation}\label{ineq:minimax:dim} \left|\, \widehat{dim}_D  - \di \,\right| \leq \ \,\frac{C(D,\alpha,m,M)}{\textup{log}\, n} \end{equation}
with probability at least $1-C'(\alpha,M)/n$. Then, Theorem \ref{thm:dim:lowerbound} shows that no estimator can improve on the (order of the) bound \eqref{ineq:minimax:dim}. The proof is written in Appendix \ref{append:dim:lowerbound}.

\begin{theorem}\label{thm:dim:lowerbound}For any $D >2$, some numerical constants $\alpha,m,M,v > 0$ and all large enough $n$, we have

\begin{equation*}
\underset{\widehat{d}}{\textup{inf}} \quad    \underset{ \mathcal{W}_{n}(D,\alpha,m,M,v)}{\textup{sup}} \quad  \Pb_{(\Omega, \mu, W)} \Bigg{[}| \hat{d} - dim\, \Omega | \geq  \frac{1}{2 \log (n)}\Bigg{]} \geq \frac{1}{4}
\end{equation*}\smallskip

\noindent where $\underset{\widehat{d}}{\textup{inf}}$ is the infimum over all estimators.\end{theorem}

Let us discuss the minimal aspect of the conditions defining $\mathcal{W}_{n}(D,\alpha,m,$ $M,v)$. First, the assumption that the dimension is upper bounded seems natural, as our available data $A \in \{0,1\}^{n \times n}$ is a finite set. Indeed, for metric spaces $(\Omega_n, r_{W_n})$ with arbitrary large dimensions (like $\di_n/n \rightarrow \infty$ for instance), a finite sample $\om_1,\ldots,\om_n$ may look like a set of distant and isolated points, which does not reflect the true geometry of $(\Omega_n, r_{W_n})$. Since this situation is not conducive to accurate estimates of the complexity of $\Omega_n$, we avoid it by assuming the dimension is upper bounded. Second, we show that the assumptions \ref{ass1} and \ref{ass2} are minimal, in the sense that, removing any one of them entails a large loss for any estimator. Specifically, let $\mathcal{W}_{n}^{min(j)}(D,\alpha,m,M,v)$ be the collection of all (pure) graphons satisfying all conditions of the set $\mathcal{W}_{n}(D,\alpha,m,M,v)$ except the condition $H_j$ (where $H_j$ denotes \ref{ass1} or \ref{ass2} according to the value of $j \in \{1,2\}$). Then, Theorem \ref{large:loss:dim} shows that any estimator suffers from an error of the order $D$, over the class $\mathcal{W}_{n}^{min(j)}(D,\alpha,m,M,v)$. The proof is written in Appendix \ref{append:dim:lowerbound}.

\begin{theorem}\label{large:loss:dim}
For any $D >2$, some numerical constants $\alpha,m,M,v > 0$, all $j \in \{1,2\}$ and all large enough $n$, we have \begin{equation*}\underset{\widehat{d}}{\textup{inf}} \quad  \underset{ \mathcal{W}_{n}^{min(j)}(D,\alpha,m,M,v)}{\textup{sup}} \,  \Pb_{(\Omega, \mu, W)} \Bigg{[}| \hat{d} - dim\, \Omega | \geq  \frac{D}{2}\Bigg{]} \geq \frac{1}{4}
\end{equation*}\smallskip

\noindent where $\underset{\widehat{d}}{\textup{inf}}$ is the infimum over all estimators.
\end{theorem}

\bigskip

\textsc{Remark:} our optimal rate of estimation may seem at odds with the faster rates of convergence in the literature about intrinsic dimension estimation, see \citep{kim2016minimax} for instance. This is due to the important differences in the modeling assumptions. In the work of \citet{kim2016minimax}, for example, the observed data are $n$ i.i.d. sampled points from a well-behaved manifold in $\mathbb{R}^m$ whose dimension is an integer. In contrast, here we do not assume the dimension is an integer, we do not observe the $n$ sampled points $\om_1,\ldots,\om_n$, and do not know the metric $r_W$.\bigskip

\textsc{Comments on \ref{ass1}, \ref{ass2} :} we only make the assumptions \ref{ass1}, \ref{ass2} at a small scale, that is for $\ep \in (0,v]$. Besides, the right hand side of \ref{ass2} is almost free since it is already implied by \ref{ass1} for $M = 2^d/\alpha$. Let us briefly explain how these assumptions imply the error bound of Theorem \ref{coro:asympDim}. The assumption \ref{ass1} ensures that the difference between the sampled points $\om_1,\ldots,\om_n$ and the latent space $\Omega$ is not too large. By definition, this implies that the sampling error \eqref{def:error:samp} and the distance error \eqref{dist:sup} are small. Accordingly, we can choose a radius $\ep_D$ that is larger than these two errors, and reliably estimate the $\ep_D$-covering number $\cov(\ep_D)$ by Proposition~\ref{thm:cov:true}. Then, we use a plug-in to estimate the quantity $- \log \, \cov(\ep_D)\big{/} \log \,\ep_D$, which is a good approximation of the dimension by assumption~\ref{ass2}. To sum up, the radius $\ep_D$ must be larger than the sampling and distance errors, but still small enough to well approximate the Minkowski dimension with $- \log \, \cov(\ep_D)\big{/} \log \,\ep_D $.

\section{Testing the complexity}
\label{subsec::test_complexity}

Given the adjacency matrix of a W-random graph, we want to known if the graph is simple or complex. In other words, we would like to test the null-hypothesis $\cov(\ep) \leq K$ for a given $K >0$, with a specific care for minimizing the assumptions on the graphon. However, instead of using the covering number we use the packing number $\pac(\ep)$ for some reasons to be specified in Section \ref{subsec:underestim::useless}. For now, note that it is essentially the same measure as the covering number, and all previous results of the paper can be adapted to the packing number (without any significant difference). See Appendix \ref{subsec::additionalInfo} for a reminder of this usual measure for metric spaces. 

In hypothesis testing, it is common to be conservative and focus on the minimization of the type I error, which is the probability of rejecting the null-hypothesis incorrectly. Accordingly, our objective is to control the type I error without any assumption on the graphon, while keeping a control of the type II error under reasonable assumptions. (the type II error is the probability of accepting the null-hypothesis incorrectly)

\subsection{Testing the null-hypothesis without assumption on the graphon, via under-estimation of the packing number}\label{subsec:underestim::useless}
To test the null-hypothesis without assumption on the graphon, we want to define a complexity estimator that does not overestimate the true complexity w.h.p.. Unfortunately, the inequality on the covering number estimator from Proposition~\ref{thm:cov:true}
\begin{align*} \covh\left(\epsilon + b_{sup} + s_{\om}\right) \leq \cov\left(\epsilon\right)\end{align*}is difficult to leverage for an under-estimation since the errors $b_{sup}$ and $s_{\om}$ are unknown and take specific values for each graphon. However, we show below that the sampling error $s_{\om}$ can be removed, by working with the packing number instead of the covering number. Then, we show that the distance error bound $b_{sup}$ can be handled with a slight modification of the distance-estimator $\widehat{r}$, defined earlier by \eqref{def:dist:estimator}. 

Based on the distance estimator $\widehat{r}$, we can define a plug-in estimator $\pach(\ep)$ of the packing number, as we did for the covering number estimator. This estimator satisfies almost the same non-asymptotic bounds as the covering number estimator, see the following proposition, which is a slight variant of Proposition~\ref{thm:cov:true}. The proof is omitted.

\begin{proposition}\label{prop:ineq:packnumb}Given any graphon $(\Omega, \mu, W)$, consider the data distribution $\mathbb{P}_{(\Omega, \mu, W)}$ defined in model (\ref{model::data}). Let $b_{sup}$ and $s_{\om}$ be the distance error bound (\ref{dist:sup}) and the sampling error (\ref{def:error:samp}). Then, the packing number estimator $\pach$ satisfies the following inequalities    
\begin{equation*}
 \forall \ep > b_{sup}, \hspace{1cm} \pac\left(\epsilon + b_{sup} + 2s_{\om}\right) \leq \pach \left(\epsilon\right)  \leq \pac\left(\epsilon - b_{sup} \right)\end{equation*}with probability at least $1-\frac{2}{n}$ with respect to the distribution $\mathbb{P}_{(\Omega, \mu, W)}$.\end{proposition}
 
Hence, we have 
\begin{equation*}
    \pach\left(\epsilon + b_{sup}\right) \leq \pac\left(\epsilon\right)
\end{equation*}without the sampling error $s_{\om}$ anymore. 

The next step is to control the remaining error term $b_{sup}$ coming from the estimator $\widehat{r}$.  Obviously, when $b_{sup}$ is positive and relatively large, the estimator $\widehat{r}$ over-estimates some distances, and thus the plug-in estimator $\pach$ based on $\widehat{r}$ may over-estimate the packing number. In order to avoid such an over-estimation, we modify the previous estimator $\widehat{r}$, by taking a maximum in front of the negative term involved in (the expression of) $\widehat{r}^2$ :
\begin{equation}\label{def:new:dist:estim} \small
 \widehat{r}_{new}^2(i,j) := \left[\langle A_i,A_{\widehat{m}(i)}\rangle_n+ \langle A_j,A_{\widehat{m}(j)}\rangle_n- 2\max_{k\in \{i,\widehat{m}(i)\}, l\in \{j,\widehat{m}(j)\}} \langle A_k,A_l\rangle_n\right]_+ 
\end{equation}We show in Appendix \ref{proof:lem:undere} that the new distance estimator $\widehat{r}_{new}$ satisfies the same upper bound as $\widehat{r}$ in Theorem~\ref{thm:dist:upperBound} (up to a numerical constant $5/3$) but also successfully under-estimates the neighborhood distance $r_W$ in some sense (see Lemma \ref{lem:undere} for details). Then, defining a new packing number estimator based on $\widehat{r}_{new}$, which is denoted by $\pachne$ in the following,  we finally get the wanted under-estimation of the packing number, see Theorem \ref{pack:under:last} below. The proof is written in Appendix \ref{proof:lem:undere}. 

\begin{theorem}\label{pack:under:last}Given any graphon $(\Omega, \mu, W)$, consider the data distribution $\mathbb{P}_{(\Omega, \mu, W)}$ defined in model (\ref{model::data}). Then, for the radius $\widehat{\ep} = \sqrt{\ep^2 + t_n}$ with $t_n = 12 \,\sqrt{\frac{\textup{log}\, n}{n}}$, the estimator $\pachne$ satisfies the following inequalities    
\begin{equation}\label{lem:articl:underestim}
    \forall \ep > 0, \hspace{1cm} \pac\left(\widehat{\epsilon} + \frac{5}{3}b_{sup} + 2s_{\om}\right) \leq \pachne(\widehat{\ep}\,) \leq \pac(\ep)
\end{equation} with probability at least $1-\frac{2}{n}$ with respect to the distribution $\mathbb{P}_{(\Omega, \mu, W)}$.
\end{theorem} Thus, without any assumption on the graphon, the estimator $\pachne(\widehat{\ep}\,)$ does not overestimate the $\ep$-packing number with high probability. Besides, the left hand side of \eqref{lem:articl:underestim} shows that it does not under-estimate (significantly) more than the previous estimator $\pach$ of the packing number (seen in Proposition \ref{prop:ineq:packnumb}). 


\subsection{Results on the packing number test}


We accept the null hypothesis if and only if $\pachne(\widehat{\ep}\,) \leq K$. The upper bound \eqref{lem:articl:underestim} ensures that the type I error is controlled for all graphons, which gives the following result.

\begin{corollary}\label{coro:typeIerror}
For any graphon, the type I error is lower than  $\frac{2}{n}$ 
with respect to the distribution $\mathbb{P}_{(\Omega, \mu, W)}$.
\end{corollary}



By definition of the packing number, the type II error is small as soon as $K+1$ sampled points are separated by at least a distance $\widehat{\ep} + err$, where $err$ upper bounds all errors of distance estimation between the $K+1$ points. This condition on the sampled points is satisfied w.h.p. by each of the following graphons. 

Given two parameters $\eta>0$ and $\beta > 1/ n$, let $\mathcal{W}(\eta, \beta)$ denote a collection of graphons for which there exist $K+1$ balls $B(x_1,\eta_1),\ldots,B(x_{K+1},\eta_{K+1})$ in $(\Omega,r_W)$ such that\begin{enumerate}
     \item the $K+1$ balls are weighted enough: $\mu\left[ B(x_i, \eta_i) \right] \geq \beta$ for all $i\in[K+1],$
     \item the radii are small enough: $\eta_i \leq \eta/2$ for all $i\in[K+1]$,
     \item the centers are spaced enough: $r_W(x_i, x_j) \geq \sqrt{\ep^2 + 10\eta + 6 t_n} +  \eta.$
\end{enumerate}The small-ball condition 1. is similar to the assumption \ref{ass1} for the dimension estimation; it ensures that some of the sampled points $\om_1,\ldots,\om_n$ belong to the $K+1$ balls w.h.p.. The third condition 3. ensures that these balls are enough distant from each other, so that the sampled points in these balls are separated enough, in order to have $\pachne(\widehat{\ep}) \geq K+1$ and confirm the alternative hypothesis correctly. 

\begin{theorem}\label{thm:type2}
Assume the graphon $(\Omega, \mu, W)$ belongs to $\mathcal{W}(\eta, \beta)$ for some $\beta > 1/n$. Then, the type II error is smaller than \begin{equation*}
\frac{2}{n} + 2 \, \beta n (K+1)\, \textup{exp}[-\beta (n-1)]    
\end{equation*} with respect to the distribution $\mathbb{P}_{(\Omega, \mu, W)}$.
\end{theorem}

The proof of Theorem \ref{thm:type2} is written in Appendix \ref{appendiTypetwoerror}. This result implies that, for any graphon in $\mathcal{W}(\eta, \beta)$, the type II error is convergent to zero as soon as the measure of each ball $B(x_i,\eta_i)$ is large enough to satisfy $\beta \gtrsim n^{-1}.$ For example, if each of the $K+1$ balls has a measure that is larger than $\textup{log} [K n]/n$, then the type II error is smaller than $\textup{log} ( n) /n$ up to some numerical constant. In Appendix \ref{improvement:test:2}, Theorem \ref{thm:type2} is improved by using the graphon regularity at a finer level (see Theorem \ref{coro:type2}).
 

\section{Further considerations}\label{section::further_consideration}

\subsection{Estimation of the complexity with sparse observations}\label{sect:sparse:dim:dist}
In the W-random graph model (\ref{model::data}), each node has an average degree that is linear with $n$ the total number of nodes. However, real-world networks are often sparse with node degrees varying from zero to $n$. This motivates to consider a model of sparse graph where the node degree can be an order of magnitude smaller than $n$. 

Given a sequence $\rho_n$ such that $\rho_n \rightarrow 0$, the definition of model (\ref{model::data}) can be modified to have average node degrees of the order of $\rho_n n$. Consider the adjacency matrix $A$, defined by model (\ref{model::data}), whose edges are independently retained with probability $\rho_n$ and erased with probability $1-\rho_n$. We refer to this set-up as ``the sparse setting" and denote by $\mathbb{P}_{(\Omega, \mu, W), \rho_n}$ the corresponding data distribution. This model has been considered several times in the literature \citep[see][]{bickel2011method,wolfe,klopp, Massoul}.  \smallskip

We now extend the results of Section \ref{sec:est:dist} to this sparse setting. Corollary \ref{thm:dist:upperBound:sparse} gives non-asymtotic error bounds for the distance estimation. It is a slight variant of Theorem \ref{thm:dist:upperBound}. For completeness, the proof is written in Appendix \ref{proof:thm:dist:uppBound:sparse}. 

\begin{corollary}
\label{thm:dist:upperBound:sparse}
Assume the scaling parameter $\rho_n$ is lower bounded by
\begin{equation}\label{condition:sparsite}
    \rho_n \geq 2\sqrt{\textup{log}(n )/ (n-2)}.
\end{equation}
Then, the following event \begin{align*}
\mathcal{E}_{dist}^{sp}=\bigg{\{}\forall i, j \in[n]: &\left| \rho_n^2r_{W}^2(\om_i,\om_j) - \widehat{r}^2( i, j) \right| \\ & \leq   3\rho_n  \Big{(}\rho_n  r_{W}(\om_j,\om_{m(j)}) + \rho_n r_{W}(\om_i,\om_{m(i)})  + 20 \sqrt{\textup{log}(n)/n }\Big{)}\bigg{\}} \end{align*}
holds with probability $\mathbb{P}_{(\Omega, \mu, W), \rho_n}\left(\mathcal{E}_{dist}^{sp}\right) \geq 1 - \frac{2}{n}$.
\end{corollary}


We estimate the Minkowski dimension using the following radius  \begin{equation}\label{def:good:radius:sparse}
\ep_{D, \rho_n} \asymp  \left(\frac{\textup{log}\, n}{n}\right)^{1/( 2 D)} \vee \rho_n^{-1/2}\left(\frac{\textup{log}\, n}{n}\right)^{1/4}
\end{equation} Corollary \ref{coro:asympDim:sparse} is an adaptation of Theorem \ref{coro:asympDim} for the sparse setting. The proof is written in Appendix \ref{append:E:dim:sparse}. \begin{corollary}\label{coro:asympDim:sparse}
For all graphons $(\Omega, \mu, W)$ in $\mathcal{W}(D,\alpha,m,M,v)$, all scaling parameters $\rho_n$ fulfilling \eqref{condition:sparsite}, and all radii satisfying \eqref{def:good:radius:sparse}, the following rate of estimation of the dimension holds with probability tending to $1$ as $n \rightarrow \infty$ (w.r.t. the distribution $\mathbb{P}_{(\Omega, \mu, W), \rho_n}$).  \begin{equation*} \left|\, \frac{\textup{log} \, \covh(\ep_{D, \rho_n} )}{-\textup{log}\,\ep_{D, \rho_n} } - d \,\right| \leq \ \,C(D,\alpha,m,M,t) \left\{
    \begin{array}{ll}
       1 & \textup{ if }  \rho_n \asymp \sqrt{\textup{log}(n )/n}, \\
       \\
       (\textup{log}\, n)^{-1}  & \textup{ if }  \rho_n \asymp \left(\textup{log}(n )/n\right)^{(1/2)-t},
    \end{array}
\right. \end{equation*}
where $t\in (0, 1/2)$ and $C(D,\alpha,m,M,t)$ is some constant independent of $n$. 
\end{corollary}


\subsubsection{Discussion about the distance estimation in sparse regimes and open question }

Corollary \ref{thm:dist:upperBound:sparse} extends the upper bounds to the sparse setting where each entry of the adjacency matrix $[A_{ij}]_{1\leq i,j \leq n}$ is observed with probability $\rho_n$ independently. The scaling parameter is assumed to satisfy $ \rho_n \gtrsim \sqrt{\log(n)/n }$ where the $\sqrt{\log(n)}$ in the numerator allows us to show high probability concentration, and the $1/\sqrt{n}$ ensures that there are enough data for the distance estimator to work. Indeed, the distance estimator is a linear combination of scalar products, and most scalar products are equal to zero in the very sparse regime $\rho_n = o(\sqrt{1/n})$ where pairs of nodes tend to have no common neighbors (i.e. common non-zero entries in the adjacency matrix).

Therefore, one needs to find new strategies to estimate the distances in very sparse regimes. A direction of research could be, for example, the nearest neighbor method in \citep{borgs2017iterative} which is successfully applied to matrix estimation when $\rho_n \gtrsim n^{-1}\log(n)^{2}$. Overall, it would be interesting to complete the picture of the distance estimation problem according to different regimes of sparsity. 

\subsection{Polynomial-time algorithm (with theoretical guarantee)}
\label{section:illustration}

In contrast with the previous sections, here we take into account the computational aspect of the problem. Computing the covering number of a finite set is NP-hard, hence we approximate it with a greedy algorithm \citep{greedy}.

For completeness, the polynomial-time procedure for estimating $\cov(\ep)$ is described below. 
The algorithm proceeds in two steps: Step 1 computes all distances $\widehat{r}(i,j)$ using the distance-estimator \eqref{def:dist:estimator}; in particular, this step requires the computation of all index estimators $\widehat{m}(j)$ defined by \eqref{def:indexEstim}. Step 2 approximates the $\ep$-covering number of $\{1,\ldots,n\}$ w.r.t. the distance estimator $\widehat{r}$, by sequentially selecting balls (of radius $\ep$) according to one rule: at each stage, select the ball that contains the largest number of uncovered elements. At the end of the process, the number of selected balls is returned. This output is denoted by $\covhap(\ep)$. \newpage

\noindent { \def\arraystretch{1.3}
\begin{tabular}{|l|}
\hline
 \textsc{Covering Number Algorithm }\label{alg:pseudoCode} \\
\hline
\begin{minipage}{0.95\textwidth} \centering
\begin{minipage}{0.9\textwidth}

\medskip
{\bf Input:} $A = [A_{ij}]$ adjacency matrix of size $n \times n,$ a radius $\ep.$

\medskip
\noindent \textbf{Step 1 : constructing the distance-estimator $\widehat{r}$}
\begin{itemize}
\item[1.] Compute the nearest neighbor's index of each sampled point $\om_i$:\\ $\forall i \in \{1,\ldots,n\}$, \ \, \ \, $\widehat{m}(i)\,  = \, \underset{j : \ \,  j\neq i}{\mbox{argmin}}\,  \ \,  \underset{k: \ \, k \neq i, j}{\mbox{max}} \ \, \big{|} \langle A_{k},A_{i}-A_{j}\rangle_n \big{|}$.
\item[2.] Compute all the distances: \\ $\forall i, j \in \{1,\ldots,n\},$ \ \, $\widehat{r}(i,j) = \langle A_{i} , A_{\widehat{m}(i)} \rangle_n + \langle A_{j} , A_{\widehat{m}(j)} \rangle_n - 2\, \langle A_{i} , A_{j} \rangle_n$.
\end{itemize}

\noindent \textbf{Step 2 : computing an approximation of the $\ep$-covering number}
\begin{itemize}
\item[3.] In the space $\mathcal{S}_0 = \{1,\ldots,n\}$ endowed with the distance function $\widehat{r}$, consider $\mathcal{B}_0 = \{B_j\}_{j\leq n}$ the set of all the balls of radius $\ep$.
\item [4.] Obtain a cover of $\{1,\ldots,n\}$ as follows: \\
Set $i=0$. While $\mathcal{S}_i \neq \emptyset,$ do:
\begin{itemize}
    \item [(a)] Select a ball $B$ in $\mathcal{B}_i$ that contains the largest number of elements of $\mathcal{S}_i$.
     \item [(b)] Set $\mathcal{S}_{i+1} = \mathcal{S}_i \setminus B$ to remove the elements covered by $B$,
     \item [(c)] Set $\mathcal{B}_{i+1} = \mathcal{B}_{i}\setminus \{B\}$ to update the set of available balls, 
     \item[(d)] Set $i = i+1$ to continue the algorithm.
\end{itemize}
\end{itemize}

\smallskip

\noindent \textbf{Output:} the number $i$ of selected balls, denoted by $\covhap(\ep)$.
\end{minipage}%
\end{minipage} \\
\hline
\end{tabular} }\\

\begin{figure}
\centering
\includegraphics[width=0.55\textwidth]{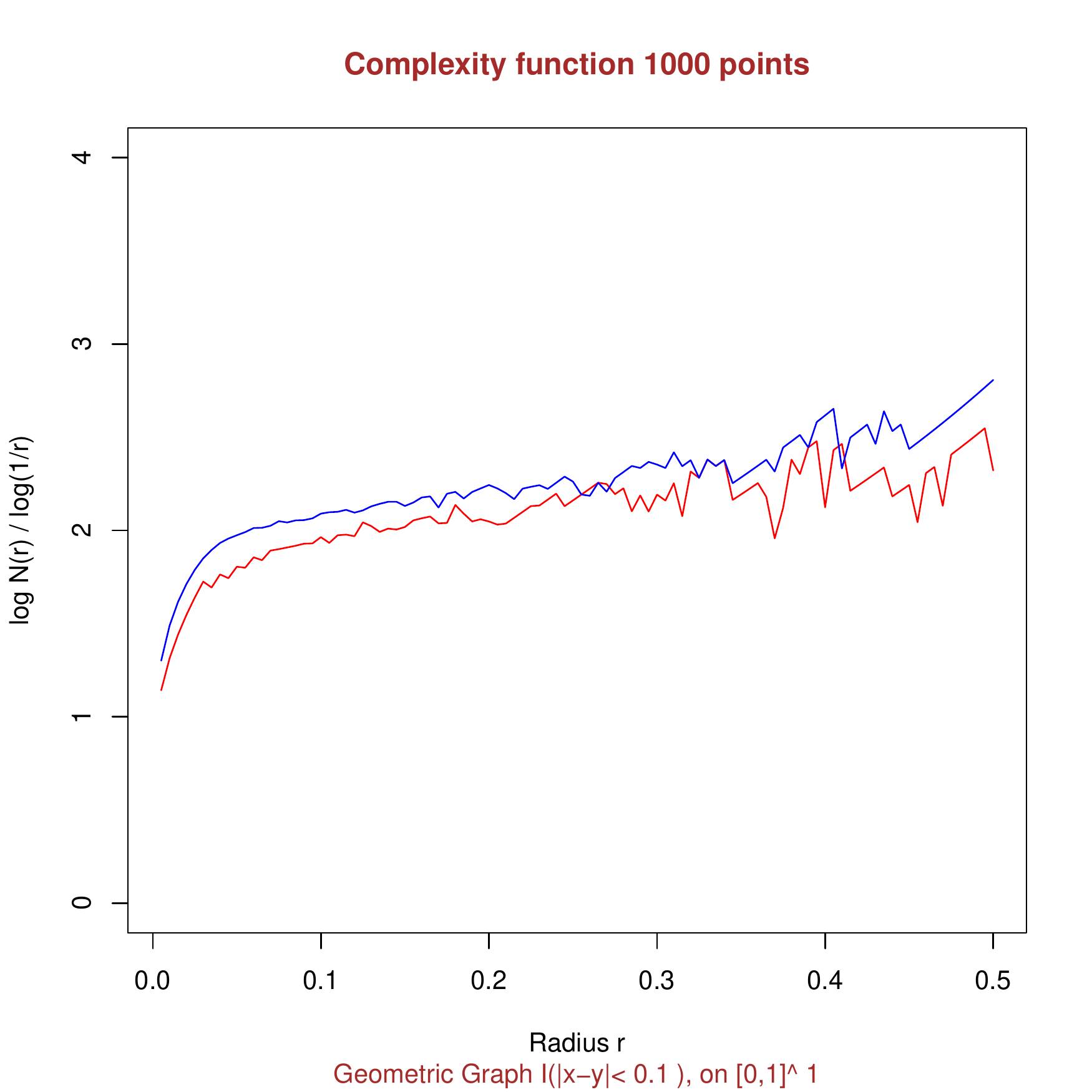}
\caption{W-random graph with Minkowski dimension $2$}
\label{fig::numeric}
\end{figure}

We also suggest an heuristic for tuning $\ep$ in the estimation of the Minkowski dimension. First, run several times \textsc{Covering Number Algorithm} for a range of different radii $\ep_1,\ldots,\ep_t$, and then plot $\textup{log}\,\covhap(\ep_j) \big{/} \textup{log}\, \ep_j$ for $j = 1,\ldots,t$. As in Figure \ref{fig::numeric}, we look for a graph function that (roughly) admits the three following parts: (1) for big radii, the shape of the curve is irregular and seems sawtooth; (2) for medium radii, there is almost a plateau whose value is the dimension estimate; (3) for small radii, there is an abrupt drop towards zero.

According to the theoretical guarantee of the greedy algorithm \citep{greedy}, one has \begin{equation*}\covh(\ep) \leq \covhap(\ep) \leq 2 \, \textup{log}(n)  \covh(\ep)\end{equation*} where $\covh(\ep)$ is the consistent estimator introduced in Section~\ref{sec:est:dist}. 
Then, for graphons fulfilling the assumptions of Theorem \ref{coro:asympDim}, there exist some radii $\ep$ such that $-\textup{log}\, \covhap(\ep)\big{/}\textup{log}\,\ep$ is close to the Minkowski dimension up to a small error term $-\textup{log}\, \big{(}2\,\textup{log}(n)\big{)}\big{/}\textup{log}\,\ep$.\medskip



We shortly illustrate the empirical performance of our algorithm on the random geometric graph, introduced in Section \ref{section::illustExample}. Consider the latent space $[0,1]$, endowed with the uniform measure and the function $W(x,y) = \mathbb{I}_{||x-y||_2 \leq 0.1}$, which has a Minkowski dimension $2$ and satisfies the assumptions of Theorem \ref{coro:asympDim}. We sample $n = 1000$ points uniformly on $[0,1]$ and plot the outputs $-\textup{log}\, \covhap(\ep)\big{/}\textup{log}\,\ep$ over the range of radii $\ep \in \Big{\{}0.005+k*0.005\, ; \,k \in \{0,\ldots,100\}\Big{\}}$. This is represented by the red curve in Figure \ref{fig::numeric}. As we can see, it is close to the true dimension at some intermediate radii, which coincides with our theoretical results. Specifically, we observe the three typical parts in the graph function:  (1) on the right of the figure, the sawtooth-shaped curve means that the radius is too big for approaching the Minkowski dimension (which is by definition a limit in $\ep \rightarrow 0$);  (2) on the middle, there is a plateau whose value is close to the dimension;  (3) on the left, there is an abrupt drop because the covering number estimator eventually just counts the sampled points $\om_1,\ldots,\om_n$. As a reference, we also plot $-\textup{log}\, \covsap(\ep)\big{/}\textup{log}\,\ep$ in blue, where $ \covsap(\ep)$ is the \textit{approximated} covering number of the sample $\{\om_1,\ldots,\om_n\}$ w.r.t. to the true distance $r_W$.  




\section{Discussion}

\subsection{On the definition of our complexity index}

We have introduced an index of complexity for the limiting distributions $\mathbb{P}^{(\infty)}_{(\Omega, \mu, W)}$ of $W$-random graphs. It has some geometric flavor as the index is based on the Minkowski dimension of the metric space $(\Omega,r_W)$, where $\Omega$ is the latent space of the $W$-random graph and $r_W$ is the neighborhood distance defined on $\Omega$. Accordingly, the index inherits from the general features of the Minkowski dimension which hold for any metric space. One of these features if the following maximum property: for any metric space $(X,d)$ which admits a partition $X = X_1 \sqcup \ldots \sqcup X_k$, it is known that the Minkowski dimension of $X$ is equal to the maximum of the Minkowski dimensions of the $X_i$, $i\in [k].$ Thus, our index captures a maximum complexity of $(\Omega, r_W)$, which is an interesting information about the distributions $\mathbb{P}^{(\infty)}_{(\Omega, \mu, W)}$ of $W$-random graphs. In order to complete our index, it would be worthwhile to investigate other notions of complexity as well. For example, one could think of some sort of average complexity instead of the maximum complexity measured by our index, maybe replacing the Minkowski dimension with a dimension that satisfies the following kind of property: for any metric space $X = X_1 \sqcup \ldots \sqcup X_k$ endowed with a probability distribution $\mu$, the dimension of $X$ would be equal to an average of the dimensions of the $X_i$ weighted according to $\mu$. However, for such an index based on a dimension of $(\Omega, r_W)$, we remind that a major difficulty is to verify the identifiability of the index from the distribution $\mathbb{P}^{(\infty)}_{(\Omega, \mu, W)}$. 

We have approximated the Minkowski dimension using a greedy algorithm in section \ref{section:illustration}, because the computational cost of this dimension is prohibitive. As an alternative, the correlation dimension is widely-used in manifold learning for its computational simplicity. Given sampled points $x_1,\ldots,x_n \overset{iid}{\sim} \mu$ in a metric space $(X,d)$ endowed with a probability distribution $\mu$, the correlation integral is usually defined as $$C(\ep) = \underset{n \rightarrow \infty}{\textup{lim}} \quad \frac{2}{n(n-1)}\sum_{i=1}^n \sum_{\substack{j=1 \\ j\neq i}}^n \mathbb{I}_{\{d(x_i,x_j) < \ep\}}$$
where $\mathbb{I}_{\mathcal{A}}$ denotes the indicator function of any event $\mathcal{A}.$ If the limit exists, the correlation dimension of $(X,d)$ is  $$D_{corr}= \underset{\ep \rightarrow 0}{\textup{lim}} \, \frac{\log \, C(\ep) }{\log\, \ep}.$$ It is known that $D_{corr}$ approximates well the Minkowski dimension when the distribution $\mu$ is nearly uniform on $X$, whereas it may be smaller for non-uniform distributions \citep{Kegl}. In fact, one can observe that $D_{corr}$ satisfies a minimum property, that is, the correlation dimension of $X = X_1 \sqcup \ldots \sqcup X_k$ is equal to the minimum of the correlation dimensions of the $X_i$, $i\in [k].$ A direction of research could be to extend the correlation dimension to $W$-random graph distributions, so that one get a complexity index that is simple to compute. To this end, one could consider the following form of the correlation integral $$C(\ep) = \mathbb{E}_{\, \substack{ \om_1 \sim \mu\\ \om_2 \sim \mu}} \left[ \mathbb{I}_{\{r_W(\om_1,\om_2) < \ep\}}\right]$$with respect to the metric space $(\Omega, r_W)$ associated with a graphon $(\Omega, \mu, W)$.



\subsection{On the rates of estimation}

The focus of the current paper is on the whole class of graphons, which is a very general setting. In particular, our error bounds for the neighborhood distance and the covering number hold for any graphon. Therefore, a natural question is whether faster rates could be derived on sub-classes of graphons.

To estimate the Minkowski dimension of $(\Omega, r_W)$, we consider traditional assumptions in manifold learning, namely \eqref{ass1} and \eqref{ass2}, which essentially say that the metric space $(\Omega,r_W)$ behaves like a Euclidean space of dimension $d$ endowed with a uniform distribution $\mu$. Under these Euclidean-type assumptions, we prove that the optimal rate of estimation is $\textup{log}^{-1}\, n$. This rate matches known results for the Minkowski dimension in manifold learning, see for instance \citep{koltchinskii}. A future research direction for $W$-random graphs could be the definition of new indices that enjoy faster rates of estimation.

\acks{We would like to thank Nicolas Verzelen for sharing his ideas, and Christophe Giraud for valuable advice.}
\vskip 0.2in
\bibliography{biblio}
\appendix

\section{Additional information}

\label{subsec::additionalInfo}
\subsection{Basic information on the covering and packing numbers and the Minkowski dimension}

Given any set $S$, its covering number $N^{(c)}(\ep)$ is the minimal number of balls of radius $\ep$ required to entirely cover $S$, with the constraint that the ball centers are in $S$. This measure is widely used for general metric spaces. Likewise, the packing number $N^{(p)}(\ep)$ is the maximum number of points in a given space (strictly) separated by at least a given distance $\ep$. Both measures are similar and linked by the following inequalities 
$N^{(c)}(\ep) \leq N^{(p)}(\ep) \leq N^{(c)}(\ep/2)$. In all the paper (except the last subsection~\ref{subsec::test_complexity}), our results are mostly stated with the covering number, but each of them can be adapted to the packing number. 

The covering number requires to choose the scale $\ep$ at which we look at the data. To get rid of this parameter, it is common to consider the Minkowski dimension which is defined by $\textup{lim}_{\epsilon \rightarrow 0} \ \, - \textup{log} \, N^{(c)}(\epsilon) \big{/} \textup{log} \, \epsilon.$ Note that the same formula holds with the packing number instead. The Minkowski dimension is used for infinite (separable) spaces, when the covering number diverges to infinity as $\epsilon$ goes to zero. This dimension is therefore complementary to the covering number. It is known to match with some other classical notions of dimension in simple cases, for example the Minkowski dimension of the hypercube $[0,1]^d$ is equal to its Euclidean dimension $d$. The Minkowski dimension has the advantage to be applicable on a wide range of spaces (whose dimension is not necessarily an integer) and to be easy to compute (in comparison with the Hausdorff dimension for example). 

\subsection{Details on the illustrative examples}
\label{subsec:detailsIllustratExamp}

\textsc{Random H\"older graph.} Recall that the graphon $(\Omega, \mu, W)$ is $([0,1]^d, \lambda, W)$ where $\lambda$ is the uniform measure on $[0,1]^d$ and $W$ satisfies the following condition:
there exist three constants $m, M, \alpha >0$ such that for all $\om, \om', \om'' \in [0,1]^d$, \begin{equation*}
m \big{|}\big{|}\om'-\om\big{|}\big{|}_2^{\alpha} \leq \big{|}W(\om',\om'')-W(\om,\om'')\big{|} \leq M \big{|}\big{|}\om'-\om\big{|}\big{|}_2^{\alpha}    
\end{equation*} where $\alpha$ is the level of regularity of the function $W$ and $||\om'-\om||_2$ is the Euclidean distance between $\om'$ and $\om$ in $[0,1]^d$. From the above display, we directly deduce some bounds on the neighborhood distance (\ref{def:rW}) :\\ $\forall \om,\om' \in [0,1]^d,$ \label{exmp:Holder:dist} \begin{equation*}m||\om'-\om||_2^{\alpha} \leq r_W(\om',\om) \leq M ||\om'-\om||_2^{\alpha}.\end{equation*} Thus, the distance $r_W$ behaves (up to some constants) like the Euclidean distance on $[0,1]^d$ raised to the power of $\alpha$. As the covering number of the Euclidean hypercube $([0,1]^d, ||.||_2)$ is approximately equal to $\ep^{-d}$ for small radii, we have \begin{equation*}\left(\epsilon/m\right)^{-d/\alpha} \lesssim \cov(\epsilon) \lesssim  \left(\epsilon/M\right)^{-d/\alpha}.\end{equation*} Hence $\di = d/\alpha$, which means that the Minkowski dimension of $(\Omega, r_W)$ is equal to the ratio between the Euclidean dimension of the latent space $[0,1]^d$ and the regularity of the function $W$.\smallskip

\noindent \textsc{Random geometric graph example.} Recall that the graphon is $([0,1]^d, \lambda, W)$ where $\lambda$ is the uniform measure, and $W$ is defined as
$W(\om,\om') = \mathbb{I}_{||\om-\om'||_2 \leq \delta}$ for some parameter $\delta \in ]0, 1[$, and $||\om-\om'||_2$ is the Euclidean distance between $\om, \om'\in [0,1]^d$. Here, the bounds on the neighborhood distance are rather involved and deferred to the Appendix \ref{appendix::randomGeom}. The main message is that \begin{equation*}
r_W(\om,\om') \asymp \sqrt{||\om-\om'||_2}
\end{equation*}if $||\om-\om'||_2$ is small enough, which means that the distance $r_W$ behaves like the squared root of the Euclidean norm in $[0,1]^d$. Following the line of the Random H\"older graph example, we can see that $\cov(\epsilon)$ behaves like $\epsilon^{-2d}$ for $\ep$ small enough. By definition of the Minkowski dimension, it follows that $\di = 2 d$.

\subsection{Test: improvement of the type II error}\label{improvement:test:2}

The control of the type II error can be refined using the graphon regularity at a finer level. Instead of considering the set $\mathcal{W}(\eta, \beta)$ of graphons with $K+1$ well separated balls (Theorem \ref{thm:type2}), here we consider the new set $\mathcal{W}(\eta, \beta, M, K')$ of graphons with $M$ disjoint collections of $K+1+K'$ separated balls. That is, for a collection of $K+1+K'$ balls, we assume the same conditions of separation, size and measure as in a collection of $K+1$ balls defined by $\mathcal{W}(\eta, \beta)$ (in Theorem \ref{thm:type2}). In addition, we assume that the $M$ formations of $K+1+K'$ balls do not intersect each other (i.e. no ball from a collection overlaps a ball from another collection). Thus, the new set $\mathcal{W}(\eta, \beta, M, K')$ of graphons is linked with the previous one by the following equality $\mathcal{W}(\eta, \beta, 1, 0) = \mathcal{W(\eta, \beta)}$. 

\begin{theorem}
\label{coro:type2}
If the underlying graphon belongs to $\mathcal{W}(\eta, \beta, M, K')$ with $\beta \geq 1/n$, then the type II error is smaller than $\frac{2}{n} + \widetilde{p}_{\, n}^{\, M}$, where $\widetilde{p}_n$ admits the following upper bound  $$ \binom{K+K'+1}{K'+1} \Big(2\beta n~\textup{exp}[-\beta (n-1)]\Big)^{(K'+1)}.$$
\end{theorem}

The proof of Theorem \ref{thm:type2} is written in Appendix \ref{appendiTypetwoerror}.\medskip

\section{Proofs for illustrative examples}

\subsection{Proof of Proposition \ref{prop:approxSBM}: approximation by SBM}
\label{appendix::approxSBM}Given a graphon $(\Omega, \mu, W)$ and a radius $\ep>0$, we consider a cover of $(\Omega, r_W)$ whose the cardinality is $\cov(\ep)$ (written $N$ for brevity), and the ball centers are $x_1,\ldots,x_N$. The Voronoi cell $V_j$ of $x_j$ is the set of all elements in $\Omega$ that are closer to $x_j$ than to any other $x_k$, $k \neq j$, according to the metric $r_W$. In the case of equality, where a point $\omega$ is at equal distance of several ball centers $x_i$, it belongs to the Veronoi cell of smallest index $i$.
\begin{align*}
    V_j := \Big{\{}\omega \in \Omega: \, r_W(\omega,x_j)< r_W(\omega,x_k) \textup{ if }  k < j, \textup{ and }\, r_W(\omega,x_j)\leq  r_W(\omega,x_k) \textup{ otherwise}\Big{\}} 
\end{align*} Define the SBM approximation of $W$ as follows: 
\[\overline{W}(x,y)= \sum_{i,j=1}^N 1_{x\in V_i}1_{y\in V_j}\frac{1}{\mu(V_i)\mu(V_j)}\int_{V_i}\int_{V_j}W(z_1,z_2)\mu(dz_1)\mu(dz_2)\]
By triangular inequality and Jensen inequality:
\[
 \int_{\Omega^2} (W(x,y) - \overline{W}(x,y))^2\mu(dx)\mu(dy) \leq    2\int_{\Omega^2} \sum_{j=1}^N 1_{y\in V_j}\Big{[}\frac{1}{\mu(V_j)}\int_{V_j} [W(x,y) -  W(x,z_2)]^2\mu(dz_2)\Big{]}\mu(dx)\mu(dy)  +\]
\[ 2\int_{\Omega^2}\Big{[} \sum_{i,j=1}^N 1_{x\in V_i}1_{y\in V_j}\frac{1}{\mu(V_i)\mu(V_j)}\int_{V_i}\int_{V_j}[W(x,z_2) -  W(z_1,z_2)]^2\mu(dz_1)\mu(dz_2)\Big{]}\mu(dx)\mu(dy)
\]
Note that the first term is smaller than $8\epsilon^2$ by integrating with respect to $x$ and using the fact that $y$ and $z_2$ belong to the same Voronoi cell. The second term simplifies 
$$2\int_{\Omega^2}\Big{[} \sum_{i,=1}^N 1_{x\in V_i}\frac{1}{\mu(V_i)}\int_{V_i}[W(x,y) -  W(z_1,y)]^2\mu(dz_1)\Big{]}\mu(dx)\mu(dy)
$$
which is again smaller than $8\ep^2$. The approximation error of $W$ by $\overline{W}$ is therefore lower than $4 \ep$ in $l_2$-norm. The proposition is proved.\hfill $\blacksquare$ 

\subsection{The neighborhood distance for the random geometric graph example}
Lemma \ref{claim:dist:RGG} gives bounds on the neighborhood distance for the random geometric graph of Section~\ref{section::illustExample}. For simplicity, we neglect the side effects associated with a point too close of the side of $\Omega = [0,1]^d$. That is, we assume the parameter $\de$ is small compared to $1$ (where $1$ is the length of a side of $[0,1]^d$). Write $V_d$ the volume of the unit ball in $[0,1]^d$ endowed with the Euclidean norm $||.||_2$, and write $I_x(., .)$ the (regularized) incomplete beta function \citep[see][Eq.8.17.2 for a definition]{NIST}. 
\begin{lemma}\label{claim:dist:RGG}If $ ||x-y||_2>2\delta,$ then $r_W^2(x,y)=2V_d \delta^d$; otherwise $r_W^2(x,y) = 2V_d \delta^d I_{x}(\frac{1}{2},\frac{d+1}{2})$ for $x=\left(\frac{||x-y||_2}{2 \delta}\right)^2$. As a consequence, $\sqrt{||x-y||_2} \lesssim r_W(x,y) \lesssim \sqrt{||x-y||_2}$ as soon as $||x-y||_2$ is small enough (compared to $\de$). 
\end{lemma}\label{appendix::randomGeom}
According to the above lemma, the neighborhood distance $r_W$ behaves like the squared root of the Euclidean norm of $[0,1]^d$ if $||x-y||_2$ is small enough.
For lower dimensions, for instance $d=3$, we can also use the paper of \citet{li} to get the simpler formula:\\ if $||x-y||_2 < 2\delta,$ then
\begin{equation*}r_W^2(x,y)=2\pi\left(\delta^2-\frac{||x-y||_2^2}{12}\right) ||x-y||_2.\end{equation*} 

\begin{proof}\textbf{of Lemma \ref{claim:dist:RGG}.} For the random geometric graph, observe that the computation of the neighborhood distance is equivalent to the computation of the volumes of hypersherical caps. Using the formula (3) in the paper of \citet{li} (and neglecting the side effects due to the boundary of the latent space), we have:\\  if $ ||x-y||_2 < 2\delta,$ then
\begin{equation*}r_W^2(x,y) = 2V_d \delta^d \left[1-I_{x}(\frac{d+1}{2},\frac{1}{2})\right]\end{equation*} where $x=1-\left(\frac{||x-y||_2}{2 \delta}\right)^2$. Basic properties of the (regularized) incomplete beta function \citep[see][Eq.8.17.4]{NIST}  allows to rewrite the last formula:\\ if $ ||x-y||_2<2\delta,$ then\begin{equation}\label{eq:dem:rgg:cap}
r_W^2(x,y) = 2V_d \delta^d I_{x}(\frac{1}{2},\frac{d+1}{2})\end{equation} \label{proof:claim:dist:RGG}where $x=\left(\frac{||x-y||_2}{2 \delta}\right)^2$. Let $B(a,b)$ denote the beta function \citep[][Eq.5.12.1]{NIST}, then the above formula (\ref{eq:dem:rgg:cap}) can be developed using the recurrence formula $I_x(a,b+1) = I_x(a,b) + \frac{x^a(1-x)^b}{bB(a,b)}$ \citep[][Eq.8.17.21]{NIST}.  It follows that $r_W$ satisfies the following bounds: $\sqrt{||x-y||_2} \lesssim r_W(x,y) \lesssim \sqrt{||x-y||_2}$ as soon as $||x-y||_2$ is small enough.\end{proof}

\section{Identifiability and pure graphons}

\subsection{Proof of Lemma \ref{lem:inv:dist} : invariance of the neighborhood distance}
 
Given two equivalent pure graphons $(\Omega, \mu, W)$ and $(\Omega', \mu', W')$, let us show that their respective neighborhood distances $r_W$ and $r_{W'}$ are linked by the following $\mu'\otimes \mu'$-almost surely equality 
\begin{equation*}
    r_W\left(\phi(x), \phi(y)\right) = r_{W'}\left(x, y\right)
\end{equation*}for some measure-preserving bijection $\phi : \Omega' \rightarrow \Omega$.

It follows from Lemma \ref{thm:puregraphon:lov}, which links any two equivalent pure graphons. Denote by $W^{\phi}$ the function $(x,y)\mapsto W(\phi(x),\phi(y))$. \begin{lemma}[\citeauthor{lovasz2}, \citeyear{lovasz2}, Section 13.3] \label{thm:puregraphon:lov}If two pure graphons $(\Omega, \mu, W)$ and $(\Omega', \mu', W')$ are equivalent, then there exists a bijective measure-preserving map $\phi : \Omega' \rightarrow \Omega$ such that $W^{\phi}(x,y)= W'(x,y)$  $\mu' \otimes \mu'$-almost surely.
\end{lemma}\label{proof:lem:inv:dist}

Indeed, by definition of the neighborhood distance, 
\begin{equation*}
r_{W'}\left(x, y\right) = \left( \int_{\Omega'} \left| W'(x, z') - W(y, z')\right| ^2  \mu'(dz') \right)^{1/2}\end{equation*}
which gives the following $\mu' \otimes \mu'$-almost surely equality by Lemma~\ref{thm:puregraphon:lov}, 
\begin{equation*}\label{dist:interm:util}
   r_{W'}\left(x, y\right) = \left( \int_{\Omega'} \left| W(\phi(x),\phi(z')) - W(\phi(y),\phi(z')) \right| ^2  \mu'(dz') \right)^{1/2}
\end{equation*}for some measure-preserving bijection $\phi : \Omega' \rightarrow \Omega$. Then, using a pushforward measure (or image measure),
\begin{equation*}
   r_{W'}\left(x, y\right) = \left( \int_{\Omega} \left| W(\phi(x),z) - W(\phi(y),z) \right| ^2 \mu(dz) \right)^{1/2}
\end{equation*} $\mu' \otimes \mu'$-almost surely, so that, by definition of the neighborhood distance,
\begin{equation*}
   r_{W'}\left(x, y\right) = r_W\left(\phi(x), \phi(y)\right)
\end{equation*}$\mu' \otimes \mu'$-almost surely. Lemma \ref{lem:inv:dist} is proved.\hfill $\blacksquare$

\subsection{Proof of Lemma \ref{inv:thm} : identifiability of the covering number}

Given two equivalent pure graphons $(\Omega, \mu, W)$ and $(\Omega', \mu', W'),$ let us prove that their respective covering numbers are equal: $\cov(\epsilon ) = \covbi(\epsilon)$ for all $\ep > 0$. 

According to Lemma~\ref{lem:inv:dist}, there exists a measure-preserving bijection $\phi$, such that the two metric spaces $(\Omega, r_W)$ and $(\Omega',r_{W'})$ are linked by the equality $r_{W'}\left(x, y\right) = r_W\left(\phi(x), \phi(y)\right)$ on a subset of measure $1$, say $\Sigma \subseteq \Omega'$ with $\mu'(\Sigma)=1$. This means that both subpaces $(\phi(\Sigma), r_W$) and $(\Sigma, r_{W'})$ are linked by a bijection that preserves the distances, which directly implies equality between their covering numbers: $N^{(c)}_{\phi(\Sigma)}(\ep) = N^{(c)}_{\Sigma}(\ep)$ for all $\ep > 0.$

Then, for proving Lemma \ref{inv:thm}, it is enough to show the two following inequalities 
\begin{align}
    &\cov(\epsilon) \geq N^{(c)}_{\phi(\Sigma)}(\epsilon + \delta)\label{proof:ineq:identif:1}\\ 
    &N^{(c)}_{\Sigma}(\epsilon + \delta)\geq \covbi(\epsilon + \delta)\label{proof:ineq:identif:2}
\end{align}
for any $\delta > 0$. Indeed, combining these two inequalities with the covering number equality from the above paragraph, one has $\cov(\epsilon) \geq \covbi(\epsilon + \delta)$. Taking the limit $\de \rightarrow 0$ and using the right-continuity of the covering number (Lemma \ref{lem2:proof}), this gives $\cov(\epsilon) \geq \covbi(\epsilon)$. As the reverse inequality holds by symmetry of the proof, one obtain the equality $\cov(\epsilon ) = \covbi(\epsilon)$ of Lemma \ref{inv:thm}.

\begin{lemma}
\label{lem2:proof}Given a pure graphon $(\Omega, \mu, W)$, the function $\epsilon \mapsto \cov(\epsilon)$ is piecewise constant and right-continuous (note that we use closed balls in the definition).\\
Likewise, $\epsilon \mapsto \pac(\epsilon)$ is a right continuous piecewise function.
\end{lemma}\label{appB}


Assume $\Sigma$ is dense in $(\Omega', r_{W'})$. Each cover of $\Sigma$ is closed as a finite union of closed balls. Hence it is also a cover of $\Omega'$ by density of $\Sigma$ in $\Omega'$. This proves \eqref{proof:ineq:identif:2}. Likewise, assume $\phi(\Sigma)$ is dense in $(\Omega, r_W)$. An $\ep$-cover of $\Omega$ can be transformed into an $(\ep+\de)$-cover of $\phi(\Sigma)$ by moving the ball centers from $\Omega$ to $\Sigma$ and increasing the ball radius of $\de$ (for arbitrary small $\de$). This proves \eqref{proof:ineq:identif:1} for any $\de>0$.

Let us show the density of $\phi(\Sigma)$ in $(\Omega, r_W)$. One has $\mu(\phi(\Sigma))= \mu'(\Sigma) = 1$ by definition of a (bijective) measure-preserving map, which implies that $\phi(\Sigma)$ intersects each ball of non-zero measure in $(\Omega, r_W)$. As the measure of a pure graphon has full-support by definition, then each ball of non-zero radius has a non-zero measure. Thus, $\phi(\Sigma)$ intersects each ball of non-zero radius in $(\Omega, r_W)$, which means that $\phi(\Sigma)$ is dense in $(\Omega, r_W)$. Similarly, we can show the density of $\Sigma$ in $(\Omega',r_{W'})$.



Lemma \ref{inv:thm} is proved for the covering number. The proof for the packing number is similar and omitted.\hfill $\blacksquare$ \bigskip

\begin{proof}\textbf{of Lemma \ref{lem2:proof}.}\label{proof:rightContin} The function $\epsilon \mapsto \cov(\epsilon)$ is non-increasing from $[0, \infty[$ to the set of all non-negative integers, it is therefore a piecewise constant function. Thus, for any radius $\ep_0>0$, there exists a (strictly) larger radius $\ep_1$ such that the covering number $\cov(\epsilon)$ is equal to a constant, say $N$, over the interval $(\ep_0, \ep_1)$. To prove the right continuity in $\ep_0$, let us show the inequality $\cov(\epsilon_0) \leq N$ (since we already know the reverse inequality by monotonicity of the covering number function), or equivalently that there exists a cover of $\Omega$ that is composed of $N$ balls of radius $\ep_0$. 

Given a radius $\ep$ and $K$ points $c=(c_1,\ldots,c_K)\in \Omega^K$, denote by $C_{\Omega}(c, \ep)$ the union of $K$ balls of centers $c_1,\ldots,c_K$. In the following, we prove: (1) the existence of some $c_0 \in \Omega^N$ such that $C_{\Omega}(c_0, \ep)$ covers $\Omega$ for all $\ep \in (\ep_0, (\ep_1+\ep_0)/2]$;  (2) for such a $c_0$, $C_{\Omega}(c_0, \ep_0)$ covers $\Omega$. Thus, Lemma \ref{lem2:proof} will be proved. \smallskip

(1) Define the set $E_{\Omega}(\epsilon) := \{ c \in \Omega^N : \Omega \subseteq C_{\Omega}(c,\epsilon)\}$ for any given radius $\ep >0$. Then, consider the following sequence of nested sets $\widetilde{E}_k := E_{\Omega}(\epsilon_{0} + (\ep_1-\ep_0)/k)$ where $k \geq 2$ is an integer. The Cantor's intersection theorem (recalled in Lemma~\ref{lem:appen:cantor} below) ensures that $\cap_{k \geq 2} \widetilde{E}_k \neq \emptyset$, provided that the assumptions of the theorem hold. For clarity, this verification is deferred to the end of the proof. As the set $\cap_{\ep_0 < \ep < \ep_1} E_{\Omega}(\epsilon)$ is equal to $\cap_{k \geq 2} \widetilde{E}_k $, one has $\cap_{\ep_0 < \ep < \ep_1} E_{\Omega}(\epsilon) \neq \emptyset$, which means that there exists some $c_0 \in \Omega^N$ such that $C_{\Omega}(c_0, \ep)$ covers $\Omega$ for all $\ep \in (\ep_0, (\ep_1+\ep_0)/2]$. 


(2) By contradiction, let us prove that $C_{\Omega}(c_0, \ep_0)$ covers $\Omega$. If $C_{\Omega}(c_0,\epsilon_0)$ does not cover $\Omega$, then there exists some $y$ in the open set $\Omega \setminus C_{\Omega}(c_0,\epsilon_0)$, which implies that there exists an open ball $B(y, \eta)$ in $\Omega \setminus C_{\Omega}(c_0,\epsilon_0)$ for some radius $\eta >0$. Hence, $r_W(y, c_{0,j}) \geq \eta + \ep_0$ for all $j \in  \{1,\ldots,N\}$, which means that $C_{\Omega}(c_0,\epsilon)$ does not cover $\Omega$ for the radius $\ep = \epsilon_0 + \eta/2$ for instance. This is a contradiction with point (1) above. 
\begin{lemma}[Cantor's intersection theorem]\label{lem:appen:cantor}
Suppose that $(X, d)$ is a complete metric space, and $C_n$ is a sequence of non-empty closed nested subsets of $X$ whose diameters tend to zero. 
Then the intersection of the $C_n$ contains exactly one point, that is $\cap_{k=1}^{\infty} C_k = \{x\}$ for some $x$ in $X$.
\end{lemma}

\textit{Verification of the assumptions of Lemma \ref{lem:appen:cantor}.} Since $(\Omega, r_W)$ is a complete metric space by definition of a pure graphon, the product space $(\Omega^N,r_W^{sup})$ is also complete for the sup-distance $r_W^{sup} (x,y) := \textup{sup}_{1 \leq j \leq N}\,r_W(x_j,y_j)$ with $x =(x_1,\ldots,x_N), y=(y_1,\ldots,y_N) \in \Omega^N$. By definition of $\widetilde{E}_k$, the sequence $(\widetilde{E}_k)_k$ is composed of nested sets, which are also non-empty since $\cov(\ep) = N$ over $(\ep_0,\ep_1)$. To prove that each $\widetilde{E}_k$ is a closed subset of $\Omega^N$, it is enough to show that $E_{\Omega}(\epsilon)$ is closed for any $\ep\in(\ep_0, \ep_1)$. Let $(x^k)_{k\geq 0}$ be a sequence in $E_{\Omega}(\epsilon)$ such that $x^k \rightarrow x \in \Omega^N$ as $k\rightarrow \infty$. Then, for any $\eta>0$, there exists some $k_0$ such that the sup-distance between $x^{k_0}=(x^{k_0}_1,\ldots,x^{k_0}_N)$ and $x=(x_1,\ldots,x_N)$ is at most $\eta$. As $x^{k_0} \in E_{\Omega}(\epsilon)$, one know that, for any $y \in \Omega$, there exists some $j_0$ such that $r_W(y,x^{k_0}_{j_0})\leq \ep$. Thus, using the triangle inequality, one has for any $\eta >0$,
\begin{equation*}r_W(y,x_{j_0})\leq r_W(y,x^{k_0}_{j_0}) + r_W(x^{k_0}_{j_0}, x_{j_0}) \leq \ep + \eta \end{equation*}
which implies that $r_W(y,x_{j_0})\leq \ep$. Hence, $y\in C_{\Omega}(x,\epsilon)$ for any $y\in \Omega$, which means that $x \in E_{\Omega}(\epsilon)$. $E_{\Omega}(\epsilon)$ is therefore a closed subset of $\Omega^N$. All the conditions of Lemma~\ref{lem:appen:cantor} are checked. 

The part of Lemma \ref{lem2:proof} on the covering number is proved. For the packing number, the proof is similar and omitted.
\end{proof}
\subsection{Proof of Lemma \ref{sample::asymptoticConsistency}: asymptotic density of the sample}
Given $\ep >0$, consider a cover of $(\Omega,r_W)$ whose cardinality is the integer $\cov(\ep/4)$ (written $N$ for brevity) and whose balls are written $B_1, \ldots, B_N$. Let us upper bound the probability that (at least) one of these balls contains zero sampled point $\om_i$. Using the union bound, this probability is smaller than
\begin{equation*}\sum_{j=1}^N\mathbb{P}_{(\Omega, \mu, W)}\left\{B_j\text{ contains zero sampled point among } \om_1,\ldots,\om_n\right\}\end{equation*} 
which is upper bounded by $N (1-\mu(B_j))^n \leq N (1-\beta)^n$ where $\beta := \textup{min}_{j \in [N]} \, \mu(B_j).$ One has $\beta > 0$ since each ball of a pure graphon has non-zero measure. And as $N$ is not equal to infinity by assumption, this probability tends to zero with $n$. Thus, with high probability, all balls $B_j$ from the cover contains at least a sampled point. Finally, the asymptotic density of the sample follows from the fact that each ball of radius $\ep$ of $(\Omega, r_W)$ contains a ball $B_j$ from the cover. Lemma \ref{sample::asymptoticConsistency} is proved. \hfill $\blacksquare$
\label{proof:sample:asymptoticDense}\bigskip



\section{Estimation of the neighborhood distance}

\subsection{Proof of Theorem \ref{thm:dist:upperBound} : the upper bound}
Theorem \ref{thm:dist:upperBound} is a direct consequence of the two following propositions. 
Proposition~\ref{prop:dem:crossTerm} shows the consistency of the inner products between the rows of the adjacency matrix $A$. That is, $\langle A_{i}, A_{j} \rangle_n$ is convergent in probability towards $\langle W(\omega_{i},.),W(\omega_{j},.)\rangle$ if $i \neq j$. Actually, Proposition~\ref{prop:dem:crossTerm} gives a uniform convergence over all $i, j \in [n],$ $i\neq j$.

\begin{proposition}\label{prop:dem:crossTerm}
The following event on inner products
$$\mathcal{E}_{in} := \left\{\forall i,j \in[n]: \ \, \left|  \langle A_{i}, A_{j} \rangle_n - \langle W(\omega_{i},.),W(\omega_{j},.)\rangle \,\right|  \leq  3\sqrt{\frac{ \textup{log}\hspace{0.1cm}n}{n}}\right\}$$
holds with probability  $\mathbb{P}_{(\Omega, \mu, W)}(\mathcal{E}_{in}) \geq 1 - \frac{2}{n}$  as soon as $n \geq 6$.
\end{proposition}\label{proof:thm:dist:uppBound}

We have seen that the neighborhood distance $r_W$ can be decomposed into one crossed term and two quadratic terms as follows \begin{equation}\label{proof:decompo:dist}
    r_W^{2} (\om_i, \om_j) = \langle W(\omega_{i},.),W(\omega_{i},.)\rangle + \langle W(\omega_{j},.),W(\omega_{j},.)\rangle - 2\langle W(\omega_{i},.),W(\omega_{j},.)\rangle.
\end{equation} Proposition~\ref{prop:dem:crossTerm} ensures that the crossed term is consistently estimated. Proposition~\ref{prop:quad:dist} deals with the quadratic terms $\langle W(\omega_{i},.),W(\omega_{i},.)\rangle$.\smallskip

\begin{proposition}\label{prop:quad:dist}
Conditionally to the event $\mathcal{E}_{in}$ (defined above), the following inequalities
$$ \forall i \in [n]: \ \,\left| \langle A_{i} , A_{\widehat{m}(i)} \rangle_n - \langle W(\omega_{i},.),W(\omega_{i},.)\rangle \right|
             \leq 3 \,r_{W}(\om_i,\om_{m(i)}\,) \, + \,15   \sqrt{\textup{log}(n)/n }$$
hold simultaneously as soon as $n \geq 6.$
\end{proposition}

The estimation error of \eqref{proof:decompo:dist} by our distance estimator
\begin{equation*}
 \widehat{r}^{2}(i , j) = \langle A_{i} , A_{\widehat{m}(i)} \rangle_n + \langle A_{j} , A_{\widehat{m}(j)} \rangle_n - 2\, \langle A_{i} , A_{j} \rangle_n 
\end{equation*}follows directly from Propositions~\ref{prop:dem:crossTerm} and \ref{prop:quad:dist}. Theorem \ref{thm:dist:upperBound} is proved. \hfill $\blacksquare$ \bigskip

\begin{proof}\textbf{of Proposition \ref{prop:dem:crossTerm}.} By triangle inequality, the expression
\begin{equation*}
     \Big{|}\sum_{k}\frac{A_{ik}A_{kj}}{n} - \int_{\Omega}W(\om_i,z)W(\om_{j},z)\mu(dz)\Big{|}
\end{equation*}is smaller than 
\begin{align*}
    &\leq \frac{1}{n}\Big{|}\sum_{k\neq i,j}A_{ik}A_{kj} - 				(n-2) \int_			{\Omega}W(\om_{i},z)W(\om_{j},z)\mu(dz)			\Big{|}\\
        &+\frac{1}{n} \Big{[}(A_{ii}+A_{jj})A_{ij}+2 \int_{\Omega}W(\om_i,z)W(\om_{j},z)			\mu(dz)\Big{]}
\end{align*}which is upper bounded by

\begin{equation*}
    \leq \frac{1}{n-2}\Big{|}\sum_{k\neq i,j}A_{ik}A_{kj} - (n-2) \int_{\Omega}W(\om_{i},z)W(\om_{j},z)\mu(dz)\Big{|} + \frac{4}{n}.
\end{equation*} Conditionally to $\om_i, \om_j$ (with $i\neq j$), the $n-2$ random variables $\{A_{ik}A_{kj}: k \in[n], k \neq i,j\}$ are independent with a mean $\mathbb{E}\left[A_{ik}A_{kj}|\om_i,\om_j\right] = \int_			{\Omega}W(\om_{i},z)W(\om_{j},z)\mu(dz) $ for all $ k \neq i,j$ (where $\mathbb{E}$ is the expectation with respect to the distribution $\mathbb{P}_{(\Omega, \mu, W)}$). It follows from Hoeffding's inequality that 

\begin{align*} \mathbb{P}_{(\Omega, \mu, W)}\left(\frac{1}{n-2}\Big{|}\sum_{k\neq i,j}A_{ik}A_{kj} - 				(n-2) \int_			{\Omega}W(\om_{i},z)W(\om_{j},z)\mu(dz)			\Big{|} \geq  \epsilon \ \, \Bigg{|} \om_i, \om_j\right) \end{align*} is lower than \begin{equation*}
     \leq 2 \textup{exp}\left(-2(n-2)\ep^2\right) \leq  2 \textup{exp}\left(-n\epsilon^{2}\right)
\end{equation*}
for $\epsilon > 0$ and $n \geq 4$. Since the above inequality is satisfied for almost every $\om_i, \om_j \in \Omega$, one has the same upper bound with probability $1$ without conditioning. Hence, taking a union bound over all $i \neq j$ one obtain
\begin{equation*} \mathbb{P}_{(\Omega, \mu, W)}\left(\bigcup_{i,j: i\neq j} \left\{\frac{1}{n-2}\Big{|}\sum_{k\neq i,j}A_{ik}A_{kj} - 				(n-2) \int_			{\Omega}W(\om_{i},z)W(\om_{j},z)\mu(dz)			\Big{|} \geq \epsilon \right\}\right) \leq 2 n^{2} \textup{exp}\left(-n\epsilon^{2}\right). \end{equation*}
Then, setting $\epsilon = \sqrt{\frac{3 \,\textup{log}\hspace{0.1cm}n}{n}}$ one derive
\begin{equation*}
 \mathbb{P}_{(\Omega, \mu, W)}\left(\bigcup_{i,j: i\neq j} \left\{\frac{1}{n-2}\Big{|}\sum_{k\neq i,j}A_{ik}A_{kj} - 				(n-2) \int_			{\Omega}W(\om_{i},z)W(\om_{j},z)\mu(dz)		\Big{|} \geq \sqrt{\frac{3 \,\textup{log}\hspace{0.1cm}n}{n}} \right\}\right) \leq \frac{2}{n}. 
\end{equation*}
Combining the above expressions, we get the following inequality 
\begin{equation*}\underset{i,j: i\neq j} {\mbox{max}}  \Big{|}\sum_{k}\frac{A_{ik}.A_{kj}}{n} - \int_{\Omega}W(\om_i,z)W(\om_{j},z)\mu(dz)\Big{|}  \leq \sqrt{\frac{3 \,\textup{log}\hspace{0.1cm}n}{n}} + \frac{4}{n} \leq 3\sqrt{\frac{ \textup{log}\hspace{0.1cm}n}{n}}\end{equation*}
with probability at least $1 - \frac{2}{n}$ as soon as $n \geq 6$.\end{proof}

\begin{proof}\textbf{of Proposition \ref{prop:quad:dist}.}
\begin{align}\label{eq:proof:prop:quad}
 \left| \langle A_{i} , A_{\widehat{m}(i)} \rangle_n - \langle W(\omega_{i},.),W(\omega_{i},.)\rangle \right|
             &\leq \left| \langle A_{i} , A_{\widehat{m}(i)} - A_{m(i)}              \rangle_n \right| \nonumber \\ &+       \left| \langle A_{i} , A_                    {m(i)} \rangle_n - \langle W(\omega_{i},.),W(\omega_{i},.)\rangle \right|           
\end{align}
 For the second term of the upper bound (\ref{eq:proof:prop:quad}),
\begin{align*}
     \left| \langle A_{i} , A_{m(i)} \rangle_n - \langle W(\omega_{i},.),W(\omega_{i},.)\rangle \right|
        &\leq \left| \langle A_{i} , A_{m(i)} \rangle_n  - \langle W(\om_i,.),            W(\om_{m(i)},.) \rangle \right|\\ & + \left| \langle W(\om_i,.), W(\om_{m(i)},.) - W                 (\om_i,.)             \rangle \right|\\
       &\leq 3 \sqrt{\textup{log}(n)/n } + r_W(\om_i,\om_{m(i)})
  \end{align*}
by Proposition~\ref{prop:dem:crossTerm} and Cauchy-Schwarz inequality. For the first term of the upper bound (\ref{eq:proof:prop:quad}), if $\widehat{m}(i) \neq m(i)$, 
\begin{align*}
    \left| \langle A_{i} , A_{\widehat{m}(i)} - A_{m(i)} \rangle_n \right| 
         &\leq \left| \langle A_{i} - A_{m(i)} , A_{\widehat{m}(i)}\rangle_n \right| + \left| \langle  A_{i} - A_{\widehat{m}(i)} , A_{m(i)}\rangle_n \right| \\
         & \leq \widehat{f}(i, m(i)) + \widehat{f}(i, \widehat{m}(i))\\
         & \leq 2\widehat{f}(i, m(i))
  \end{align*}
by definition of $\widehat{m}(i)$ and $\widehat{f}$ in (\ref{minim}). We upper bound $\widehat{f}(i, m(i))$ as follows.
\begin{align*}
     \widehat{f}(i, m(i)) := \underset{k \neq i, m(i)}{\mbox{max}}\left|             \langle A_{k},A_{i}-A_{m(i)}\rangle_n \right|
     &\leq \underset{k \neq i, m(i)}{\mbox{max}} \left|\langle W(\om_k,.), W(\om_i,.)- W(\om_{m(i)},.)\rangle \right| \\
     & + 2\underset{l, t :\,  l \neq t}{\mbox{max}} \left| \langle A_{l} ,        A_{t} \rangle_n  - \langle W(\om_l,.),W(\om_t,.) \rangle \right|\\
     & \leq r_W(\om_i,\om_{m(i)}) + 6 \sqrt{\textup{log}(n)/n }
  \end{align*}
by Proposition \ref{prop:dem:crossTerm} and Cauchy-Schwarz. Combining the upper bounds on (\ref{eq:proof:prop:quad}), Proposition \ref{prop:quad:dist} is proved. \end{proof}

\subsection{Proof of Theorem \ref{thm:simple:minimax} : the lower bound}
Theorem \ref{thm:simple:minimax} is a corollary of Theorem \ref{prop:imp2} (written below). 
Let $\textbf{r}_{\omega}$ denote the $n\times n$ symmetric matrix with entries $r_W(\om_i,\om_j)$, $1\leq i \leq j \leq n$. Given a real $\de > 0$, a graphon $(\Omega, \mu, W)$, a permutation $\sigma$ of $\{1,\ldots, n\}$ and an estimator $\widehat{d}$, we define  
\begin{align*}
 \mathcal{S}_{(\Omega, \mu, W)}(\,\widehat{d} \,, \sigma, \textbf{r}_{\omega}) = \bigg{\{}(i,j) \, : \,32 &  \left| \widehat{d}^2(\sigma(i), \sigma(j)) - r_{W}^2(\om_{i}, \om_{j}) \right| \, \geq \, 2\,\de \\ & \textup{ and } \quad 2\,\de \geq \,  r_{W}(\om_{i}, \om_{m(i)})+r_W(\om_j,\om_{m(j)})\bigg{\}}
\end{align*}and \begin{align}\label{mesureDiffBI}
\Phi_{(\Omega, \mu, W)}(\,\widehat{d} \,, \textbf{r}_{\omega}) =  \underset{\sigma}{\textup{inf}} \ \, \text{Card} \quad \mathcal{S}_{(\Omega, \mu, W)}(\,\widehat{d} \,,  \sigma, \textbf{r}_{\omega}) 
\end{align}\label{proof:lowerboundDist}where $\Phi_{(\Omega, \mu, W)}(\,\widehat{d} \,, \textbf{r}_{\omega})$ is the number of pairs $(i,j)$ where the estimator $\widehat{d}$ is no better than our estimator $\widehat{r}$, roughly speaking. That is, $\Phi_{(\Omega, \mu, W)}(\,\widehat{d} \,, \textbf{r}_{\omega})$ counts the pairs $(i,j)$ for which the error of $\widehat{d}$ is larger than the bias of our distance estimator $\widehat{r}$, which is $r_W(\om_i,\om_{m(i)}\,)+r_W(\om_j,\om_{m(j)}\,)$ up to some numerical constants. We put an infimum over all permutations $\sigma$ of the $n$ indices because we consider the problem of recovery of the set of distances $r_W(\om_i,\om_j)$, $1\leq i \leq j \leq n$, regardless of their labeling. According to Theorem \ref{prop:imp2}, there exists a sequence of graphons $(\Omega, \mu, W_n)$ such that for any estimator $\widehat{d}$, the quantity $\Phi_{(\Omega, \mu, W_n)}(\,\widehat{d} \,, \textbf{r}_{\omega})$ grows linearly with $n$ (on an event of positive probability). 
\begin{theorem}\label{prop:imp2} There exists a sequence $(\Omega, \mu, W_n)_{n \geq 0}$ of SBM such that for all $n \geq 10$, all $\de \in (\sqrt{\frac{8}{n-2}}, 1/40)$ and some numerical constants $c>0$ and $p >0$, the following lower bound holds

\begin{equation}
  \underset{\widehat{d}}{\textup{inf}}\ \,  \mathbb{P}_{(\Omega, \mu, W_n)} \Big{[}\Phi_{(\Omega, \mu, W_n)}(\,\widehat{d}, \textbf{r}_{\omega}) > c n\Big{]}  \geq p
\end{equation}
where $\underset{\widehat{d}}{\textup{inf}}$ is the infimum over all estimators.
\end{theorem}   
Theorem \ref{thm:simple:minimax} follows from Theorem \ref{prop:imp2}, choosing $\de = \left(\sqrt{\frac{\textup{log}\,n}{n}}\right)^{1/(1+\gamma)}$. \hfill $\blacksquare$ 

\bigskip

Note that Theorem \ref{new:coro:clari} follows from Theorem \ref{prop:imp2} for $\de = 1/50$.


\bigskip

\begin{proof}\textbf{of Theorem~\ref{prop:imp2}.} The proof follows the general scheme of reduction for testing two hypotheses \citep[see][]{yu1997assouad,tsybabook}. We start with the definition of some SBM with five communities where the latent space $\Omega$ is $\{\cC_1,\ldots,\cC_5\}$. We then show that for these SBM, any distance estimator suffers from a large loss. 

Let $n \geq 10$ and $\de \in (\sqrt{8/n-2}, 1/40)$. Consider the symmetric functions $W_n : \{\cC_1,\ldots,\cC_5\}^2 \rightarrow \{\cC_1,\ldots,\cC_5\}$ as described in Table \ref{Table1} below. That is, for the two diagonal blocks $\{\cC_1, \cC_2\}^2 $ and $\{\cC_3, \cC_4, \cC_5\}^2$, it is a constant function:
$$W_n(x,y) = \left\{
    \begin{array}{ll}
       1/2 \ \, & \mbox{ if } (x, y) \in \{\cC_1, \cC_2\}^2, \\
       1/2 \ \, & \mbox{ if } (x, y)\ \in \{\cC_3, \cC_4, \cC_5\}^2, 
    \end{array}
\right.$$
and for the upper right corner block $\{\cC_1, \cC_2\} \times \{\cC_3, \cC_4, \cC_5\}$: $$
W_n(x,y) = \left\{
    \begin{array}{ll}
       1/2+u_x\sqrt{\de/2} & \mbox{ if }  y \in \cC_3, \\
       1/2+u_x\de & \mbox{ if }  y \in \cC_4,\\
       1/2+u_x/2 & \mbox{ if }  y \in \cC_5,
    \end{array}
\right. \hspace{1cm}
u_x = \left\{
    \begin{array}{ll}
       +1 \ \, & \mbox{ if } x \in \cC_1, \\
       -1 & \mbox{ if }  x \in \cC_2.
    \end{array}
\right.
$$ The latent space $\{\cC_1,\ldots,\cC_5\}$ is endowed with the probability measure $\mu$ defined as follows:\begin{equation*}\mu(\cC_1) = \mu(\cC_2) = \frac{1-2\eta}{2}\end{equation*}
\begin{equation*}\frac{1}{2} \mu(\cC_3) = \mu(\cC_4) = \mu(\cC_5) = \frac{\eta}{2}\end{equation*}where $\eta = 2/(n-2)$.
\begin{figure}[H]

  \begin{minipage}[b]{0.45\linewidth}
   \centering
 {\renewcommand{\arraystretch}{0.2} 
{\setlength{\tabcolsep}{0.2cm}
\begin{tabular}{|c|c|c|c|c|c|}
   \hline
   \rowcolor[gray]{0.9}  & $\cC_1$ \rule[-7pt]{0pt}{20pt} & $\cC_2$ & $\cC_3$ & $\cC_4$ & $\cC_5$ \\
   \hline
   \cellcolor[gray]{0.9} $\cC_1$ & \multicolumn{2}{c|}{\multirow{2}{*}{1/2}} & $1/2+\sqrt{\de/2}$ \rule[-7pt]{0pt}{20pt}& $1/2+\de$ & 1 \\ \cline{1-1}\cline{4-6}
   
    \cellcolor[gray]{0.9} $\cC_2$ & \multicolumn{2}{c|}{} & $1/2-\sqrt{\de/2}$ \rule[-7pt]{0pt}{20pt} & $1/2-\de$ & 0 \\
   \hline
   \cellcolor[gray]{0.9} $\cC_3$ &  \rule[-7pt]{0pt}{20pt} &   & \multicolumn{3}{c|}{\multirow{3}{*}{1/2}}\\\cline{1-3}
 
    \cellcolor[gray]{0.9} $\cC_4$ &  \rule[-7pt]{0pt}{20pt} & & \multicolumn{3}{c|}{}\\\cline{1-3}
    
    \cellcolor[gray]{0.9} $\cC_5$ & \rule[-7pt]{0pt}{20pt} &  & \multicolumn{3}{c|}{}\\ \hline
\end{tabular}}}
\captionof{table}{values of $W_n(\cC_i,\cC_j)$\label{Table1}}
  \end{minipage}
\hfill
  \begin{minipage}[b]{0.45\linewidth}
   \centering

{\renewcommand{\arraystretch}{0.1} 
{\setlength{\tabcolsep}{0.1cm}
\begin{tabular}{|c|c|c|c|c|c|}
   \hline
   \rowcolor[gray]{0.9}  & $\cC_1$ \rule[-7pt]{0pt}{20pt} & $\cC_2$ & $\cC_3$ & $\cC_4$ & $\cC_5$ \\
   \hline
   \cellcolor[gray]{0.9} $\cC_1$ & \multicolumn{2}{c|}{\multirow{2}{*}{$\leq 2 \eta$}} & \multirow{2}{*}{$ \geq \de/4$} \rule[-7pt]{0pt}{20pt}& \multirow{2}{*}{$ \leq 5\, \de^2$} & \multirow{2}{*}{$\geq 1/4$} \\ \cline{1-1}
   
    \cellcolor[gray]{0.9} $\cC_2$  \rule[-7pt]{0pt}{20pt} & \multicolumn{2}{c|}{} &  &  &  \\
   \hline
\end{tabular}}}
\captionof{table}{bounds on $r^2_W(\cC_i,\cC_j)$\label{Table 2}}
 
  \end{minipage}
 
  \label{fig:ma_fig}

\end{figure}

We compute some bounds on the neighborhood distance associated with the above SBM, see Table \ref{Table 2} for a summary. These bounds follow easily from the definition (\ref{def:rW}) of the distance. For example, \begin{align*}r^2_W(\cC_1, \cC_3) \geq \int_{\{\cC_1, \cC_2\}} \left| W(\cC_1,z) - W(\cC_3,z) \right| ^2  \hspace{0.1cm}\mu(dz) \geq \big{(}\mu(\cC_1)+\mu(\cC_2)\big{)}\de/2 \geq (1-2\eta) \de/2\end{align*}which is larger than $\de/4$ since $\eta = 2/(n-2)$ and $n \geq 10$.\medskip

We now introduce two events $\mathcal{R}_1$ and $\mathcal{R}_2$ on the sampled points $\om_1,\ldots,\om_n$, which lead to different sets of distances (for $r_W$), and yet are difficult to decipher for any estimator based on the adjacency matrix $A$. In addition, we want these two events to happen with a positive probability $p$ that is independent of $n$. Observe that the union of the two communities $\cC_1, \cC_2$ have a total weight $1- 2 \eta = 1-4/(n-2)$ and thus concentrate most of the probability measure, whereas each of the remaining communities $\cC_3, \cC_4,$ $\cC_5$ has a weight of the order of $n^{-1}$. It follows that most of the sampled points $\om_1,\ldots,\om_n$ belong to the communities $\cC_1, \cC_2$ with large probability. In particular, the two following events 
\begin{eqnarray*}\mathcal{R}_1 &=& \Big{\{}\cC_1 \cup \cC_2,\, \cC_4,\, \cC_5 \text{ respectively contain n-2, 1, 1 sampled points}\Big{\}}\\
 \mathcal{R}_2 &=& \Big{\{}\cC_1 \cup \cC_2,\, \cC_3 \text{  respectively contain n-2, 2 sampled points}\Big{\}}\end{eqnarray*}
happen with a positive probability that is independent of $n$ (Lemma~\ref{lem:lowProba}).

\begin{lemma}\label{lem:lowProba}The probability of each event $\mathcal{R}_1$ and $\mathcal{R}_2$ is lower bounded by some numerical constant $p > 0$ :
\begin{equation*}\mathbb{P}(\mathcal{R}_2) \geq \mathbb{P}(\mathcal{R}_1) \geq p
\end{equation*}\smallskip

\noindent where $\mathbb{P}(\mathcal{R}_k) :=\mathbb{P}_{(\Omega, \mu, W_n)}\left(\mathcal{R}_k\right) = \int_{(\om_1,\ldots,\om_n)\in \{\cC_1,\ldots,\cC_5\}^n} 1_{\mathcal{R}_k}(\om_1,\ldots,\om_n) \,  \mathrm{d}\mu(\om_1)\ldots\mu(\om_n)$.
\end{lemma}\smallskip

One of the interests of the two events $\mathcal{R}_1, \mathcal{R}_2$ is to lead to different sets of distances. Specifically, if $\mathcal{R}_1$ (resp. $\mathcal{R}_2$) holds, the random matrix $\textbf{r}_{\omega} = [r_W(\om_i,\om_j)]_{i,j\in[n]}$ of distances is denoted by $\textbf{r}_1 =[r_1(i,j)]_{i,j\in[n]}$ (resp. $\textbf{r}_2 =[r_2(i,j)]_{i,j\in[n]}$). We measure the difference between both matrices $\textbf{r}_1$, $\textbf{r}_2$ of distances as follows:

\begin{equation}\label{proof:def:measure:dist}
\widetilde{\Phi}(\textbf{r}_1, \textbf{r}_2 ) = \underset{\sigma}{\textup{inf}} \ \, \text{Card} \left\{(i,j) \, :  
    \begin{array}{ll}
       16 \Big{|}r_{2}^2(i,j) - r_{1}^2(\sigma(i)\,\sigma(j)) \Big{|} \geq 2\, \de \\
       
      r_{2}(i,m(i))+r_{2}(j,m(j)) \leq 2\,\de \\
      
      r_{1}(\sigma(i),m(\sigma(i)) + r_{1}(\sigma(j),m(\sigma(j)) \leq 2\,\de
    \end{array}
\right\}  
\end{equation}\medskip

\noindent where $\widetilde{\Phi}$ is the number of pairs $(i,j)$ on which $\textbf{r}_1$ and $\textbf{r}_2$ are separated by at least the bias of our distance estimator $\widehat{r}$ (up to some numerical constants). Note that this measure is independent of the labeling $i \in \{1\ldots,n\}$ since an infimum is taken over all permutations $\sigma$ of the $n$ indices. Lemma~\ref{lem:sep:tilde} ensures that $\textbf{r}_1$ and $\textbf{r}_2$ are different enough for a number of pairs $(i,j)$ that is linear with $n$, regardless of their labeling. 

\begin{lemma}\label{lem:sep:tilde}
There exists a numerical constant $c$ such that $\widetilde{\Phi}(\, \textbf{r}_1, \textbf{r}_2 ) \geq 2 c\, n.$
\end{lemma}

So far, we have two events of positive probability which lead to two different sets of distances. It remains to see that they are difficult to decipher from the observed adjacency matrix $A$ (Lemma \ref{lem:BI:minPro}). For simplicity, write $\Pb$ for $\mathbb{P}_{(\Omega, \mu, W)}$ in the following, and $\boldsymbol{\omega}$ the $n$-tuple $(\om_1,\ldots,\om_n)$, and $\{0,1\}^{n\times n}_{sym}$ the set of binary symmetric matrices of size $n \times n$.

\begin{lemma}\label{lem:BI:minPro}For any $M \in \{0,1\}^{n\times n}_{sym}$, one has 
\begin{equation*} \Pb [ A = M | \boldsymbol{\omega} \in \mathcal{R}_1] = \Pb [ A = M | \boldsymbol{\omega} \in \mathcal{R}_2].
\end{equation*}
\end{lemma}
\medskip

We now have all the ingredients to lower bound $\Pb \Big{[}\Phi_{(\Omega, \mu, W_n)}(\,\widehat{d}, \textbf{r}_{\omega}) > c n\Big{]}$ and prove Theorem~\ref{prop:imp2}. For clarity, $\Phi_{(\Omega, \mu, W_n)}(\,\widehat{d} \,,\textbf{r}_{\omega})$ is denoted by $\Phi(\,\widehat{d},\textbf{r}_{\omega})$ in the following. Then, one has
\begin{align*}
\Pb \Big{[}\Phi(\,\widehat{d}, \textbf{r}_{\omega}) > c n\Big{]}  \geq \Pb \Big{[}\Phi(\,\widehat{d}, \textbf{r}_{\omega}) &> c n | \mathcal{R}_1 \Big{]} \mathbb{P}(\mathcal{R}_1)\\
&+ \Pb \Big{[}\Phi(\,\widehat{d}, \textbf{r}_{\omega}) > c n | \mathcal{R}_2 \Big{]} \mathbb{P}(\mathcal{R}_2)
\end{align*}

\noindent By definition of the SBM, the matrix $\textbf{r}_1$ remains the same for any $\boldsymbol{\om}\in\mathcal{R}_1$, up to a permutation of the labeling. Combining with the fact that $\Phi$ is independent of the labeling, one obtain that $\Phi(\,\widehat{d}, \textbf{r}_1)$ takes a same value for all $\boldsymbol{\om}\in\mathcal{R}_1$. Similarly, $\Phi(\,\widehat{d}, \textbf{r}_2)$ takes the same value for all $\boldsymbol{\om}\in\mathcal{R}_2$. Hence, the above display says that $\Pb \Big{[}\Phi(\,\widehat{d}, \textbf{r}_{\omega}) > c n\Big{]}$ is larger than 

\begin{equation*}
 \Big{(}\Pb \big{[}\Phi(\,\widehat{d}, \textbf{r}_1) > c n | \mathcal{R}_1 \big{]} + \Pb \big{[}\Phi(\,\widehat{d}, \textbf{r}_2) > c n | \mathcal{R}_2 \big{]}  \Big{)} \Big{(} \mathbb{P}(\mathcal{R}_1) \wedge \mathbb{P}(\mathcal{R}_2) \Big{)}
\end{equation*}

\noindent and since $\mathbb{P}(\mathcal{R}_1) \wedge \mathbb{P}(\mathcal{R}_2) \geq p$ by Lemma \ref{lem:lowProba}, one has 

\begin{equation*}
\Pb \Big{[}\Phi(\,\widehat{d}, \textbf{r}_{\omega}) > c n\Big{]}  \geq \Big{(}\Pb \big{[}\Phi(\,\widehat{d}, \textbf{r}_1) > c n | \mathcal{R}_1 \big{]} + \Pb \big{[}\Phi(\,\widehat{d}, \textbf{r}_2) > c n | \mathcal{R}_2 \big{]}  \Big{)} p
\end{equation*}

\noindent Now assume that
\begin{equation}\label{eq:help:truc}
     \Pb \big{[}\Phi(\,\widehat{d}, \textbf{r}_2) > c n | \mathcal{R}_2 \big{]} \geq \Pb \big{[}c n > \Phi(\,\widehat{d}, \textbf{r}_1) | \mathcal{R}_1 \big{]}.
\end{equation}Then, combining the two last inequalities gives
\begin{equation*}
\Pb \Big{[}\Phi(\,\widehat{d}, \textbf{r}_{\omega}) > c n\Big{]}  \geq p
\end{equation*}which gives the lower bound of Theorem~\ref{prop:imp2}.\medskip

Let us show that \eqref{eq:help:truc} holds. Lemma \ref{lem:BI:minPro} gives 
\begin{equation*}
    \Pb \big{[}\Phi(\,\widehat{d}, \textbf{r}_2) > c n | \mathcal{R}_2 \big{]}  = \Pb \big{[}\Phi(\,\widehat{d}, \textbf{r}_2) > c n | \mathcal{R}_1 \big{]}.  
\end{equation*}Then, we use the generalized triangle inequality of Lemma~\ref{lem:fin:sep} with $B= \widehat{d}$. 

\begin{lemma}\label{lem:fin:sep}For any $B \in \{0,1\}^{n\times n}_{sym}$, we have 
$ \Phi(B, \textbf{r}_1) +  \Phi(B,\textbf{r}_2) \geq \widetilde{\Phi}(\textbf{r}_1, \textbf{r}_2 )$.
\end{lemma} 

\noindent That is, 

\begin{equation*}
   \Phi(\,\widehat{d}, \textbf{r}_2) \geq \widetilde{\Phi}(\, \textbf{r}_1, \textbf{r}_2 ) -  \Phi(\,\widehat{d}, \textbf{r}_1)
\end{equation*}which is larger than
\begin{equation*}
    2 c n -  \Phi(\,\widehat{d}, \textbf{r}_1)
\end{equation*}by Lemma \ref{lem:sep:tilde}. Combing the above displays, one has
\begin{equation*}
  \Pb \big{[}\Phi(\,\widehat{d}, \textbf{r}_2) > c n | \mathcal{R}_2 \big{]} \geq \Pb \big{[}c n > \Phi(\,\widehat{d}, \textbf{r}_1) | \mathcal{R}_1 \big{]}. 
\end{equation*}The line \eqref{eq:help:truc} is therefore proved and Theorem~\ref{prop:imp2} follows.
\end{proof}

\medskip

We now show the technical lemmas, used in the proof of Theorem~\ref{prop:imp2}.

\bigskip

\begin{proof}\textbf{of Lemma~\ref{lem:lowProba}.} Let $n \geq 10$. We show that each of the two events $\mathcal{R}_1, \mathcal{R}_2$ occurs with a positive probability that is independent of $n$. By definition of the events, one has
\begin{equation*}
\mathbb{P}(\mathcal{R}_2) \geq \mathbb{P}(\mathcal{R}_1)=\frac{n(n-1)}{2} \left(\frac{\eta}{2}\right)^2 \left(1-2\eta\right)^{n-2} \end{equation*} which is equal to the following expression for $\eta = 2/(n-2),$ \begin{equation*} \frac{n(n-2)}{2} \left(\frac{1}{n-2}\right)^2 \textup{exp}\left[(n-2)  \textup{log}\,(1-\frac{4}{n-2}) \right]\end{equation*}
Using $\log(1-x)\geq -x/(1-x)$ for all $x$ in $]0, 1[$,
\begin{equation*}
\mathbb{P}(\mathcal{R}_1) \geq \textup{exp}\left[- \frac{4}{1-\frac{4}{n-2}} \right]
\end{equation*} which is larger than some positive numerical constant.  Lemma~\ref{lem:lowProba} is proved.
\end{proof}

\begin{proof}\textbf{of Lemma~\ref{lem:sep:tilde}.}
The proof consists in finding a lower bound of $\widetilde{\Phi}(\textbf{r}_1,\textbf{r}_2)$ that is linear with $n$. As $\widetilde{\Phi}$ is independent of the labeling of the set of distances $\textbf{r}_1$ and $\textbf{r}_2$, one can assume the two following labelings without the loss of generality. For the matrix $\textbf{r}_1$ (defined on the event $\mathcal{R}_1$), assume the $(n-1)^{\textup{th}}$ and $n^{\textup{th}}$ columns correspond to the two sampled points in $\{\cC_4, \cC_5\}$. For $\textbf{r}_2$ (defined on $\mathcal{R}_2$), assume the $(n-1)^{\textup{th}}$ and $n^{\textup{th}}$ columns correspond to the two sampled points in $\cC_3$. Accordingly, the $n-2$ first columns of $\textbf{r}_1$ and $\textbf{r}_2$ are associated with the sampled points in $\{\cC_1, \cC_2\}$.

We focus on the $(n-1)^{\textup{th}}$ and $n^{\textup{th}}$ columns of $\textbf{r}_2$ corresponding to the points in $\cC_3$. For the measure $\widetilde{\Phi}$, at least one these two columns will be necessarily compared to one of the $n-1$ first columns of $\textbf{r}_1$. In other words, the distances associated with a point in $\cC_3$ will be compared to the distances associated with a point in $\cC_1, \cC_2$ or $\cC_4.$ As we can see in Table \ref{Table1} and \ref{Table 2}, such comparisons will lead to the lower bound $\widetilde{\Phi}(\textbf{r}_1,\textbf{r}_2) \geq n-3$. The corresponding computation are done below, focusing on the two vectors of distances $[r_2(k,n-1)]_{ k \leq n-2}$ and $[r_2(k,n)]_{k \leq n-2}$.

By definition, $\widetilde{\Phi}(\textbf{r}_1,\textbf{r}_2)$ is based on the infimum over all permutations. Let $\sigma$ be any permutation of $\{1,\ldots,n\}$ and prove the lower bound for $\sigma$, distinguishing three cases. \medskip

\noindent \textbf{Case 1}: if $\sigma(n) =n$, then $\sigma(j) \in \{1,\ldots,n-1\}$ for all $j \leq n-1$. For convenience, note $\cC_{i,j}$ for a point in $\cC_{i}\cup \cC_j.$ For all $j \leq n-2$, one has\begin{align*}
\Big{|}r_2^2(j, n-1) - r_1^2(\sigma(j), \sigma(n-1))  \Big{|} = \Big{|}r_W^2(\cC_{1,2}, \cC_3) - r_W^2(\cC_{1,2,4}, \cC_{1,2,4})  \Big{|} \end{align*}according to the chosen labelings (described above). It follows from Table \ref{Table 2} that:

\begin{equation*}\Big{|}r_W^2(\cC_{1,2}, \cC_3) - r_W^2(\cC_{1,2,4}, \cC_{1,2,4})  \Big{|} \geq \,  \de/4 - \textup{max}(2\eta, 5\de^2)\end{equation*}which is equal to $\de(1/4-5\de)$ since $\eta = 2/(n-2)$ and $\de^2 > 8/(n-2)$ by assumption. Hence, using the condition $\de \leq 1/40$, it is larger than $\de/8$, so that, \begin{equation*}
16\,\Big{|}r_2^2(j, n-1) - r_1^2(\sigma(j), \sigma(n-1))  \Big{|} \geq 2\de    
\end{equation*} for all $j\leq n-2$.

It remains to upper bound the bias terms by $2 \de$. The ones related to $\textbf{r}_2$ are easily obtained: for all $j \leq n$,
\begin{equation*}
r_{2}(j,m(j)) \leq r_{W}(\cC_1,\cC_2) \leq 2\eta \leq \de
\end{equation*}
since on the event $\mathcal{R}_2$, a point $\om_j$ is either in $\cC_1 \cup \cC_2$ and hence $r_{2}(j,m(j)) \leq r_{W}(\cC_1,\cC_2)$, or in $\cC_3$ and thus $r_{2}(j,m(j)) = 0$ (because its nearest neighbor is in $\cC_3$ too). This gives the bounds on the bias terms
\begin{equation*}r_{2}(i,m(i))+r_{2}(j,m(j)) \leq 2\,\de
\end{equation*}for all $i,j$. The corresponding bounds for $\textbf{r}_1$ are similarly obtained from Table \ref{Table1}, but with more calculations. It is therefore encapsulated in the following lemma.

\begin{lemma}\label{proof:proof:lemma:biasTerm}If $\sigma(n) =n$, we have \,  $r_{1}(\sigma(i),m(\sigma(i)) + r_{1}(\sigma(j),m(\sigma(j)) \leq 2 \de$ \,  
for all $j, i \leq n-1$ such that $i\neq j$.
\end{lemma} Combining the above displays, we obtain the lower bound 

\begin{equation}\label{proof:lem:eqp:card:hp}\text{Card} \left\{(i,j) \, :  
    \begin{array}{ll}
       16 \Big{|}r_{2}^2(i,j) - r_{1}^2(\sigma(i)\,\sigma(j)) \Big{|} \geq 2\, \de \\
       
      r_{2}(i,m(i))+r_{2}(j,m(j)) \leq 2\,\de \\
      
      r_{1}(\sigma(i),m(\sigma(i)) + r_{1}(\sigma(j),m(\sigma(j)) \leq 2\,\de
    \end{array}
\right\} \geq n-3\end{equation}

\noindent for all permutations fulfilling $\sigma(n) =n$.\bigskip

\noindent \textbf{Case 2}: if $\sigma(n-1) =n$, then $\sigma(n),\sigma(j) \in \{1,\ldots,n-1\}$ for all $j \leq n-2$. Following the same proof as above, we can show that $\Big{|}r_2^2(n, j) - r_1^2(\sigma(n), \sigma(j))  \Big{|} \geq 2 \de$ for all $j \leq n-2$. Likewise, the bounds on the bias terms are obtained as before. The inequality \eqref{proof:lem:eqp:card:hp} is therefore proved for all permutations fulfilling $\sigma(n-1) =n$. \medskip

\noindent \textbf{Case 3}: if $\sigma(n) \neq n$ and $\sigma(n-1) \neq n$. Following the same proof as above, we can show that $\Big{|}r_2^2(n, j) - r_1^2(\sigma(n), \sigma(j))\Big{|} \geq 2 \de$ for all $j \leq n-2$ such that $j \neq \sigma^{-1}(n)$. The inequality \eqref{proof:lem:eqp:card:hp} is therefore proved for all permutations $\sigma(n) \neq n$ and $\sigma(n-1) \neq n$. \smallskip 

Finally, the lower bound \eqref{proof:lem:eqp:card:hp} is true for all permutations $\sigma,$ in particular for the infimum over all of them. Lemma \ref{lem:sep:tilde} is proved.\end{proof}

\begin{proof}\textbf{of Lemma~\ref{proof:proof:lemma:biasTerm}.} Let us upper bound the bias terms for $\textbf{r}_1$, in the case of an arbitrary permutation $\sigma$ fulfilling $\sigma(n) = n.$  On the event $\mathcal{R}_1$, one has 
\begin{equation*}
r_{1}(\sigma(i),m(\sigma(i)) + r_{1}(\sigma(j),m(\sigma(j)) \leq r_{W}(\cC_1,\cC_4) + r_{W}(\cC_1,\cC_2). \end{equation*}
for all $j, i \leq n-1$ such that $j \neq i$. In Table \ref{Table1} and Table \ref{Table 2}, one observes that
\begin{align*}
   &r_{W}(\cC_1,\cC_2)\leq \sqrt{2\eta}\\
   &r_{W}(\cC_1,\cC_4) \leq \sqrt{\de^2(1-2\eta)+(\de/2)\eta+\de^2(\eta/2)+(1/4)(\eta/2)}. 
\end{align*}The second bound is smaller than $\sqrt{\de^2(1-(3\eta/2))+\eta/4}$ since $(\de/2)\eta \leq (1/4)(\eta/2)$ (using the assumption $\de \leq 1/40$). Hence, \begin{equation*}
r_{1}(\sigma(i),m(\sigma(i)) + r_{1}(\sigma(j),m(\sigma(j)) \leq \sqrt{\de^2+\eta/4} + \sqrt{2\eta} \end{equation*}which is lower than $\de + 2\sqrt{\eta}$, and again, lower than $2\de$
(since $\sqrt{\eta} = \sqrt{2/(n-2)}$ is smaller than $\de/2$ by assumption). Lemma~\ref{proof:proof:lemma:biasTerm} is proved. \end{proof}

\begin{proof}\textbf{of Lemma~\ref{lem:fin:sep}.}
Given any matrix $B \in \{0,1\}^{n\times n}_{sym}$, let us show the following inequality $\Phi(B, \textbf{r}_1) +  \Phi(B,\textbf{r}_2) \geq \widetilde{\Phi}(\textbf{r}_1, \textbf{r}_2 )$ where $\widetilde{\Phi}$ and $\Phi$ are respectively defined by (\ref{proof:def:measure:dist}) and (\ref{mesureDiffBI}). 

For all permutations $\sigma$ of $\{1,\ldots,n\}$, the triangle inequality gives \begin{equation*}2 \Big{|}B_{ij}- r_{2}^2(i,j)\Big{|} \vee 2 \Big{|}B_{ij} - r_{1}^2(\sigma(i)\,\sigma(j)) \Big{|} \geq \Big{|}B_{ij}-r_{2}^2(i,j)\Big{|} + \Big{|} B_{ij} -r_{1}^2(\sigma(i)\,\sigma(j))\Big{|} \geq \Big{|}r_{2}^2(i,j)- r_{1}^2(\sigma(i)\,\sigma(j))\Big{|}\end{equation*}so that $\text{Card} \left\{(i,j) \, :  
    \begin{array}{ll}
       16 \Big{|}r_{2}^2(i,j) - r_{1}^2(\sigma(i)\,\sigma(j)) \Big{|} \geq 2\, \de \\
       
      r_{2}(i,m(i))+r_{2}(j,m(j)) \leq 2\,\de \\
      
      r_{1}(\sigma(i),m(\sigma(i))) + r_{1}(\sigma(j),m(\sigma(j))) \leq 2\,\de
    \end{array}
\right\}$ 
lower bounds the sum \medskip

\noindent of the two cardinal numbers

\begin{eqnarray*}\text{Card} \, \bigg{\{}(i,j) \, &:& \,32  \left| B_{ij} - r_{2}^2(i,j) \right| \, \geq \, 2~\de \geq \,   r_{2}(i,m(i))+r_{2}(j,m(j))\bigg{\}} \ \, \ \, \textup{ and }\\
  \text{Card} \, \bigg{\{}(i,j) \, &:& \,32  \left| B_{ij} - r_{1}^2(\sigma(i)\,\sigma(j)) \right| \, \geq \, 2~\de \geq \,   r_{1}(\sigma(i),m(\sigma(i))) + r_{1}(\sigma(j),m(\sigma(j)))\bigg{\}}.\end{eqnarray*} 
Taking a permutation that minimizes the latter cardinal, one has \begin{equation*}
\text{Card} \, \bigg{\{}(i,j) \, : \,32  \left| B_{ij} - r_{2}^2(i,j) \right| \, \geq \, 2~\de \geq \,   r_{2}(i,m(i))+r_{2}(j,m(j))\bigg{\}} + \Phi(B,\textbf{r}_1) \geq \widetilde{\Phi}(\textbf{r}_1, \textbf{r}_2 )    
\end{equation*}by definition of $\Phi$ and $\widetilde{\Phi}$. The above inequality holds for any matrix in $\{0,1\}^{n\times n}_{sym}$, in particular for $B^{\sigma}$ defined by $B^{\sigma}_{ij}= B_{\sigma(i),\sigma(j)}$ (where $B\in \{0,1\}^{n\times n}_{sym}$ and any permutation $\sigma$). Using $\Phi(B^{\sigma},\textbf{r}_1) = \Phi(B,\textbf{r}_1)$, the above display becomes \begin{equation*}
\text{Card} \, \bigg{\{}(i,j) \, : \,32  \left| B^{\sigma}_{ij} - r_{2}^2(i,j) \right| \, \geq \, 2~\de \geq \,   r_{2}(i,m(i))+r_{2}(j,m(j))\bigg{\}} + \Phi(B,\textbf{r}_1) \geq \widetilde{\Phi}(\textbf{r}_1, \textbf{r}_2 )    
\end{equation*}and thus, choosing the permutation that minimize the left term, 
\begin{equation*}\Phi(B, \textbf{r}_1) +  \Phi(B,\textbf{r}_2) \geq \widetilde{\Phi}(\textbf{r}_1, \textbf{r}_2 ).
\end{equation*} This generalized triangle inequality holds for all $B \in \{0,1\}^{n\times n}_{sym}$. Lemma~\ref{lem:fin:sep} is proved.\end{proof}

\begin{proof}\textbf{of Lemma~\ref{lem:BI:minPro}.} In the following, we write $\mathbb{P}$ for $\mathbb{P}_{(\Omega, \mu, W)}$, and $\mu^{\otimes n}$ for the product measure, and $\boldsymbol{\omega}$ for the $n$-tuple $(\om_1,\ldots,\om_n)$. Lemma~\ref{lem:BI:minPro} states that for all $M \in \{0,1\}^{n \times n}_{sym}$,
\begin{equation*}
    \Pb [ A = M | \boldsymbol{\omega} \in \mathcal{R}_1] = \Pb [ A = M | \boldsymbol{\omega} \in \mathcal{R}_2]
\end{equation*}which is equivalent to 
\begin{equation}\label{eq:lo:bound:truc}
  p_{\mathcal{R}_1}(M) / \Pb(\mathcal{R}_1) = p_{\mathcal{R}_2}(M) / \Pb(\mathcal{R}_2) 
\end{equation}where $p_{\mathcal{R}_1}(M)$ denotes
\begin{equation*}p_{\mathcal{R}_1}(M) := \mathbb{P}\left(\{A=M\} \cap \mathcal{R}_1\right) =  \int_{\boldsymbol{\omega}\in \mathcal{R}_k} \, \mathbb{P}(A = M | \boldsymbol{\omega})\, \mathrm{d}\mu^{\otimes n}(\boldsymbol{\omega}).
\end{equation*} Hence, we want to prove that
\begin{equation*}
     2 p_{\mathcal{R}_1}(M) = p_{\mathcal{R}_2}(M) 
\end{equation*}since $2 \Pb(\mathcal{R}_1) = \Pb(\mathcal{R}_2)$ by definition of the events $\mathcal{R}_1$ and $\mathcal{R}_2.$\medskip

Let $\mathcal{R}_1(k,l)$ be the the event defined by $\mathcal{R}_1 \cap \{(\om_k,\om_l)\in \cC_4 \times \cC_5\}$. Thus, the event $\mathcal{R}_1$ is the union $\cup_{1\leq k \neq l \leq n}\mathcal{R}_1(k,l)$. For any matrix $M = [M_{ij}]_{i,j \leq n}$ in $\{0,1\}^{n\times n}_{sym}$,
\begin{equation*}
p_{\mathcal{R}_1}(M) =  \int_{\boldsymbol{\omega}\in \mathcal{R}_1}  \mathbb{P}(A = M | \boldsymbol{\omega})\, \mathrm{d}\mu^{\otimes n}(\boldsymbol{\omega}) = \sum_{1\leq k \neq l\leq n}\int_{\boldsymbol{\omega}\in  \mathcal{R}_1(k,l)}  \mathbb{P}(A = M |\boldsymbol{\omega})\, \mathrm{d}\mu^{\otimes n}(\boldsymbol{\omega}).
\end{equation*}

Given a permutation $\sigma$ of $\{1,\ldots,n\}$, denote by $M^{\sigma}$ the matrix $M^{\sigma}_{ij} = M_{\sigma(i),\sigma(j)}$ with $i,j \in \{1,\ldots, n\}.$ Write $\sigma_{kl}$ for a permutation fulfilling $\sigma(n-1) = k$ and $\sigma(n) = l$. Then, the probability $p_{\mathcal{R}_1}(M)$ is equal to
\begin{equation*}
\sum_{1\leq k \neq l\leq n}\int_{\boldsymbol{\omega}\in  \mathcal{R}_1(k,l)}  \mathbb{P}(A^{\sigma_{kl}} = M^{\sigma_{kl}} |\boldsymbol{\omega}) \mathrm{d}\mu^{\otimes n}(\boldsymbol{\omega}) = \sum_{1\leq k \neq l \leq n}\int_{\boldsymbol{\omega}\in  \mathcal{R}_1(n-1,n)}  \mathbb{P}(A = M^{\sigma_{kl}} |\boldsymbol{\omega}) \mathrm{d}\mu^{\otimes n}(\boldsymbol{\omega}).
\end{equation*}Conditionally to $\boldsymbol{\om},$ the entries of $A$ for $i<j$ are independent Bernoulli variables, so that \begin{equation*}
p_{\mathcal{R}_1}(M) = \sum_{1\leq k \neq l\leq n} \int_{\boldsymbol{\omega}\in \mathcal{R}_1(n-1,n)} \prod_{1 \leq i < j \leq n} \mathbb{P}(A_{ij} = M_{ij}^{\sigma_{kl}} | \om_i,\om_j)  \mathrm{d}\mu^{\otimes n}(\boldsymbol{\omega}).
\end{equation*}

On the event $\mathcal{R}_1(n-1,n)$, the $\om_1,\ldots,\om_{n-2}$ are in $\cC_1 \cup \cC_1$, and $(\om_{n-1}, \om_n)$ are in $\cC_4 \times \cC_5$. As the function $W_n$ of the SBM is equal to $1/2$ on the diagonal blocks $\{\cC_1, \cC_2\}^2$ and $\{\cC_3, \cC_4, \cC_5\}^2$, one has $\mathbb{P}(A_{ij} = M_{ij}^{\sigma_{kl}} |\om_i,\om_j)= \frac{1}{2}$ for all $(i,j)$ in the set $\{(i,j): i<j \leq n-2\} \cup \{(n-1,n)\}$ of cardinality $g_n = n(n-1)/2 - 2(n-2)$. Hence, the probability $p_{\mathcal{R}_1}(M)$ is equal to 
\begin{equation*}\sum_{1\leq k \neq l\leq n} \left(\frac{1}{2}\right)^{g_n} \int_{\boldsymbol{\omega}\in  \mathcal{R}_1(n-1,n)} \prod_{1 \leq i \leq n-2}  \mathbb{P}(A_{i, n-1} = M_{i, n-1}^{\sigma_{kl}} | \om_i,\om_{n-1}) \mathbb{P}(A_{i, n} = M_{i,n}^{\sigma_{kl}} | \om_i,\om_n) \mathrm{d}\mu^{\otimes n}(\boldsymbol{\omega})\end{equation*}
or equivalently to
\begin{equation*}
\sum_{1\leq k \neq l\leq n} \left(\frac{1}{2}\right)^{g_n} \underset{(\omega_{n-1}, \omega_n) \in \cC_4 \times \cC_5}{\int} X_{M^{\sigma_{kl}}}(\omega_{n-1},\omega_n) \mathrm{d}\mu^{\otimes 2}(\omega_{n-1},\omega_n)
\end{equation*}
with \begin{equation*}
X_{M^{\sigma_{kl}}}(\omega_{n-1},\omega_n):=  \prod_{1 \leq i \leq n-2}\int_{\omega_i \in \cC_1 \cup\, \cC_2}\mathbb{P}(A_{i, n-1} = M_{i, n-1}^{\sigma_{kl}} | \omega_i, \omega_{n-1}) \mathbb{P}(A_{i, n} = M_{i,n}^{\sigma_{kl}} | \omega_i, \omega_n) \, \mathrm{d}\mu(\omega_i).
\end{equation*}

Likewise, $\mathcal{R}_2$ is the union $\cup_{1\leq k < l \leq n}\mathcal{R}_2(k,l)$ where each $\mathcal{R}_2(k,l)$ is the event $\mathcal{R}_2 \cap \{ \om_k,\om_l \in \cC_3\}$. Following the same proof as for $\mathcal{R}_1,$ one can show that 
\begin{equation*}
     p_{\mathcal{R}_2}(M) = \sum_{1\leq k < l\leq n} \left(\frac{1}{2}\right)^{g_n} \underset{(\omega_{n-1}, \omega_n) \in \cC_3 \times \cC_3}{\int} X_{M^{\sigma_{kl}}}(\omega_{n-1},\omega_n) \mathrm{d}\mu^{\otimes 2}(\omega_{n-1},\omega_n).
\end{equation*}

\begin{lemma}\label{proof:proof:lem:lem}
There exists a constant $X_{M^{\sigma_{kl}}}$ such that $X_{M^{\sigma_{kl}}}(\omega_{n-1},\omega_n)=X_{M^{\sigma_{kl}}}$ whether $\mathcal{R}_1(n-1,n)$ or $\mathcal{R}_2(n-1,n)$ holds.
\end{lemma}
Using Lemma \ref{proof:proof:lem:lem}, one has \begin{equation*}
p_{\mathcal{R}_1}(M) = \left(\frac{1}{2}\right)^{g_n}  \sum_{1\leq k \neq l\leq n}X_{M^{\sigma_{kl}}}  \mu(\cC_4)\, \mu(\cC_5)    
\end{equation*}and \begin{equation*}
   p_{\mathcal{R}_2}(M) =  \left(\frac{1}{2}\right)^{g_n} \sum_{1\leq k < l\leq n}X_{M^{\sigma_{kl}}} \mu(\cC_3)^2 
\end{equation*}so that $p_{\mathcal{R}_1}(M) = p_{\mathcal{R}_2}(M)/2$, since $\mu(\cC_4) = \mu(\cC_5) = \mu(\cC_3)/2$ (by construction of the SBM). Lemma~\ref{lem:BI:minPro} is proved.\end{proof}

\begin{proof}\textbf{of Lemma~\ref{proof:proof:lem:lem}.} For brevity, write $\mathbb{P}$ for $\mathbb{P}_{(\Omega, \mu, W)}$ in the following. By definition, $X_{M^{\sigma_{kl}}}\,(\omega_{n-1}, \omega_n)$ is the product of the $n-2$ following terms \begin{equation*}
    \int_{\omega_i \in \cC_1 \cup\, \cC_2}\mathbb{P}(A_{i, n-1} = M_{i, n-1}^{\sigma_{kl}} | \omega_i, \omega_{n-1}) \, \mathbb{P}(A_{i, n}= M_{i,n}^{\sigma_{kl}} | \omega_i, \omega_n) \, \mathrm{d}\mu(\omega_i)
\end{equation*}$i=1,\ldots,n-2.$ The above display is equal to 
    \begin{align*}
    & \int_{\omega_i \in \cC_1}\mathbb{P}(A_{i, n-1} = M_{i, n-1}^{\sigma_{kl}} | \omega_i, \omega_{n-1})  \mathbb{P}(A_{i, n} = M_{i,n}^{\sigma_{kl}} | \omega_{i}, \omega_n)  \mathrm{d}\mu(\omega_i)\\
& + \int_{\omega_i \in \cC_2}\mathbb{P}(A_{i, n-1} = M_{i, n-1}^{\sigma_{kl}} | \omega_i, \omega_{n-1})  \mathbb{P}(A_{i, n} = M_{i,n}^{\sigma_{kl}} | \omega_i, \omega_n)  \mathrm{d}\mu(\omega_i).
\end{align*}
If $(\omega_{n-1}, \omega_n) \in \cC_4 \times \cC_5$, then
\begin{align*}
    & = \int_{\omega_i \in \cC_1}[1/2+ (2M_{i, n-1}^{\sigma_{kl}}-1)\,\de]\, [1/2+ (2M_{i, n}^{\sigma_{kl}}-1)\,(1/2)] \, \mathrm{d}\mu(\omega_i)\\
& + \int_{\omega_i \in \cC_2} [1/2- (2M_{i, n-1}^{\sigma_{kl}}-1)\,\de]\, [1/2- (2M_{i, n}^{\sigma_{kl}}-1)\,(1/2)] \, \mathrm{d}\mu(\omega_i)
\end{align*}which is equal to $\Big{[}1/2 + (2M_{i, n-1}^{\sigma_{kl}}-1)(2M_{i, n}^{\sigma_{kl}}-1) \de\Big{]} \, \mu(\cC_1)$, since $\mu(\cC_1) =\mu(\cC_2)$.\\
If $(\omega_{n-1}, \omega_n) \in \cC_3\times \cC_3$, then
\begin{align*}
    & = \int_{\omega_i \in \cC_1}[1/2+ (2M_{i, n-1}^{\sigma_{kl}}-1)\,\sqrt{\de/2}] [1/2+ (2M_{i, n}^{\sigma_{kl}}-1)\,\sqrt{\de/2}]\,  \mathrm{d}\mu(\omega_i)\\
& + \int_{\omega_i \in \cC_2} [1/2- (2M_{i, n-1}^{\sigma_{kl}}-1)\,\sqrt{\de/2}][1/2- (2M_{i, n}^{\sigma_{kl}}-1)\,\sqrt{\de/2}]\, \mathrm{d}\mu(\omega_i)
\end{align*}which is equal to $\Big{[}1/2 + (2M_{i, n-1}^{kl}-1)(2M_{i, n}^{\sigma_{kl}}-1) \de\Big{]} \, \mu(\cC_1)$.

Hence $X_{M^{\sigma_{kl}}}\,(\omega_{n-1}, \omega_n)$ is equal to the same constant whether $(\omega_{n-1}, \omega_n)$ belongs to $\cC_3\times \cC_3$ or $\cC_4\times \cC_5.$ Lemma~\ref{proof:proof:lem:lem} is proved.\end{proof}


\section{Estimation of the Minkowski dimension}

\subsection{Proof of Theorem \ref{coro:asympDim}: the upper bound}

Theorem \ref{coro:asympDim} is a corollary of Theorem \ref{thm:dimMink}, which gives non-asymptotic high-probability bounds for the risk of the data-function $-\textup{log} \, \covh(\epsilon)\big{/} \textup{log}\,\epsilon$.    

\begin{theorem}\label{thm:dimMink} 
Assume the graphon $(\Omega, \mu, W)$ satisfies \ref{ass1} and \ref{ass2} and has a Minkowski dimension $d \in (0,\infty)$. If $n$ is large enough to satisfy the below inequality \begin{equation*}
    2 \textup{log}\, n/n\leq \alpha \left(v/14\right)^{2d} \wedge \left(v/14\right)^4,
\end{equation*}then the following holds with probability at least $1- (2 + 4\alpha M)/n$ with respect to the distribution $\mathbb{P}_{(\Omega, \mu, W)}$. The sum of the distance error bound \eqref{dist:sup} and the sampling error \eqref{def:error:samp} is upper bounded as follows \begin{equation}\label{def:err:thm:proof}
\frac{b_{sup}+s_{\om}}{6} \leq err_{n, d} := \left(\frac{\textup{log}\, n}{n}\right)^{1/4}+  \left(\frac{2 \, \textup{log}\, n}{\alpha \, n}\right)^{1/2d}.    
\end{equation}
For all $\ep \in (2(b_{sup}+s_{\om}), v/7]$, the covering number estimator $\covh(\ep)$ satisfies the following upper bound 
\begin{align*}
\Bigg{|} \frac{\textup{log} \, \covh(\ep) }{-\textup{log}\, \ep} - d \Bigg{|}  \, \leq \, \frac{1}{-\textup{log}\,\ep}\Bigg{[}\textup{log} \,\left( M \vee \frac{1}{m}\right) \,  + \, 6 d  \frac{err_{n, d}}{\ep} \left( 1 +  \frac{err_{n, d}}{\ep}\right)\Bigg{]}
\end{align*}
\end{theorem}\medskip

Theorem \ref{coro:asympDim} follows from Theorem \ref{thm:dimMink} by choosing any radius radius that minimizes the above upper bound, that is, any radius $\ep_D$ of the order of $\underset{\{d: \, d \leq D\}}{\textup{sup}}err_{n, d}$ $=$ $err_{n, D}$.\hfill $\blacksquare$ \bigskip

\noindent \textsc{Comments on Theorem \ref{thm:dimMink} :} We first remark that the above theorem based on the covering number can also be adapted to the packing number (without difficulties). We now comment on the two additive error terms in the upper bound. The term $-\textup{log} \,\left( M \vee (1/m)\right)\big{/}\textup{log}\, \ep$ stands for the gap between the Minkowski dimension and the quantity that we actually estimate, i.e. $-\textup{log}\, \cov(\ep)\big{/}\textup{log}\,\ep$. This gap depends on the parameters of the assumption \ref{ass2}. The second error term $-d\, err_{n, d}\big{/} (\ep \,\textup{log}\, \ep)$ represents the gap between the latter estimated quantity and the estimator $- \textup{log}\, \covh(\ep)\big{/}\textup{log}\,\ep$. To control this gap, we need to estimate the covering number correctly, and thus to control the error sum $b_{sup}+s_{\om}$ involved in Proposition \ref{thm:cov:true}. Actually, the theorem ensures that this error sum is smaller than $err_{n,d}$. This comes from the fact that the difference between the sample $\om_1,\ldots,\om_n$ and the latent space $\Omega$ is not too large, thanks to the assumption \ref{ass1}. See the proof below for details.

Finally, the upper bound holds with probability at least $1- 2/n-4\, \alpha \,M/n$. The first quantity $2/n$ corresponds to the event $\mathcal{E}_{dist}^c$ defined in Theorem \ref{thm:dist:upperBound}, i.e. that the the distance estimator does not satisfy the distance error bound $b_{sup}$. The second quantity $4\, \alpha \,M/n$ corresponds to the probability of the event where the sampled points do not cover well the latent space, leading to a large sampling error $s_{\omega}$ and a large distance error bound $b_{sup}.$ This event, denoted by $\mathcal{E}_{bad}$, is rigorously defined in the following proof.\bigskip

\begin{proof}\textbf{of Theorem \ref{thm:dimMink}.} Assume the event $\mathcal{E}_{dist}$ of Theorem \ref{thm:dist:upperBound} holds, that is the errors of distance-estimator are uniformly bounded by $b_{sup}$. On this event, Proposition \ref{thm:cov:true} gives
\begin{equation*}  \cov\left(\epsilon + b_{sup} + s_{\om}\right) \leq \covh\left(\epsilon\right)  \leq \cov\left(\epsilon - b_{sup} - s_{\om}\right)\end{equation*}for all $\ep \in (b_{sup} + s_{\om}, 1)$, so that 
\begin{equation*}
\frac{\textup{log} \, \cov(\ep + s_{\om}+b_{sup})}{-\textup{log} \ep} - d  \leq \frac{\textup{log} \, \covh(\ep) }{-\textup{log}\, \ep} - d  \leq \frac{\textup{log} \,\cov(\ep - s_{\om}-b_{sup})}{-\textup{log} \ep} - d. 
\end{equation*}
As the assumption \ref{ass2} is valid in the neighborhood $(0,v]$, we need to check that $\ep + s_{\om}+b_{sup} \in (0,v]$ to use this assumption. For clarity, we do this verification at the end of the proof. Hence, using \ref{ass2}, one has
\begin{equation}\label{ineq:dim:proof:tech}
  \frac{\textup{log} \, m }{-\textup{log}\,\ep} - d \left[\frac{\textup{log} (\ep + s_{\om}+b_{sup})}{-\textup{log} \ep} +1 \right] \leq \frac{\textup{log} \, \covh(\ep) }{-\textup{log}\,\ep} - d  \leq 
\frac{\textup{log} \, M }{-\textup{log}\,\ep} - d  \left[\frac{\textup{log} (\ep -s_{\om}-b_{sup})}{-\textup{log} \ep} +1 \right]
\end{equation}
In the right hand side of \eqref{ineq:dim:proof:tech}, the right term is upper bounded by \begin{equation*}
- d \,  \left[\frac{\textup{log} (\ep -s_{\om}-b_{sup})}{-\textup{log}\,\ep} +1 \right]\leq  - d \, \frac{\textup{log} \big{(}1 - (s_{\om}+b_{sup})/\ep\big{)}}{-\textup{log} \,\ep}   
\end{equation*}which is again upper bounded by 
\begin{equation*}
      d \, \frac{ (s_{\om}+b_{sup})/\ep + \left((s_{\om}+b_{sup})/\ep\right)^2}{-\textup{log}\,\ep}
\end{equation*} if $(s_{\om}+b_{sup}) \leq \ep/2$. Similarly in the left  hand side of \eqref{ineq:dim:proof:tech}, the right term is lower bounded by
\begin{equation*}
- d\, \left[\frac{\textup{log} (\ep + s_{\om}+b_{sup})}{-\textup{log} \ep} +1 \right] \geq - d \frac{ (s_{\om}+b_{sup})/\ep }{-\textup{log}\,\ep}.   
\end{equation*}
Combining the above displays, one derive \begin{equation}\label{proof:tech!dim:ccl:mo}
\frac{\textup{log} \,  m}{-\textup{log}\,\ep} - d \frac{ (s_{\om}+b_{sup})/\ep }{-\textup{log} \ep} \leq \frac{\textup{log} \, \covh(\ep) }{-\textup{log}\,\ep} - d  \leq \frac{\textup{log} \, M }{-\textup{log}\,\ep}+
d \, \frac{ (s_{\om}+b_{sup})/\ep + \left((s_{\om}+b_{sup})/\ep\right)^2}{-\textup{log}\,\ep}. 
\end{equation}

It remains to upper bound the error sum $s_{\om}+b_{sup}$ in \eqref{proof:tech!dim:ccl:mo}. Given a cover of $\Omega$, composed of $\cov(\eta)$ balls $B_j$ of radius $\eta$, one define the following event 
\begin{equation}\label{event:badEvent}
\mathcal{E}_{bad}(\eta):=\Big{\{}\exists j :\, B_j\text{ contains exactly 0 or 1 sampled point among $\om_1,\ldots,\om_n$ }\Big{\}}. \end{equation} \label{appen:proof:thm:dimMink}Assume the complementary event $\mathcal{E}_{bad}^c(\eta)$ holds. This means that each ball of the cover of $\Omega$ contains at least two sampled points. Hence, one has
\begin{align*}
    & s_{\om} \leq 2 \eta,\\
    & \textup{sup}_{i \in \{1,..,n\}} \, r_W(\om_i, \om_{m(i)})  \leq 2 \eta.
\end{align*}which directly implies the following upper bound \begin{equation*}b_{sup}+s_{\om} \leq 6 \left(\frac{\textup{log}\, n}{n}\right)^{1/4}+ 4 \sqrt{\eta} + 2 \eta \end{equation*}by definition of $b_{sup}$ in \eqref{dist:sup}. Thus, for the particular radius $\eta_n := \left[2\,\textup{log}(n)/(\alpha \, n)\right]^{1/d}$, \begin{equation*}b_{sup}+s_{\om} \leq  6 \left(\frac{\textup{log}\, n}{n}\right)^{1/4}+ \, 6 \left(\frac{2 \, \textup{log}\, n}{\alpha \, n}\right)^{1/2d}.\end{equation*}It follows from the definition \eqref{def:err:thm:proof} of $err_{n, d}$ that 
\begin{equation*}
    s_{\om}+b_{sup} \leq 6 err_{n, d}.
\end{equation*} Combining the above upper bound with \eqref{proof:tech!dim:ccl:mo}, one deduce the inequalities of the theorem. \smallskip

The above displays hold conditionally to the event $\mathcal{E}_{bad}^c(\eta_n) \cap \mathcal{E}_{dist}$, which happens with probability at least $1-\left(2+4\alpha M\right)/n$ (Lemma \ref{proof:lemma:THMdim:proba}).
\begin{lemma}\label{proof:lemma:THMdim:proba} The probability 
$\mathbb{P}_{(\Omega, \mu, W)}(\mathcal{E}_{bad}(\eta_n) \cup \mathcal{E}_{dist}^c )$ is smaller than $(2+4 \alpha M)/n.$
\end{lemma}

The condition $\ep + s_{\om}+b_{sup} \in (0,v]$ (used at the beginning of the proof) is satisfied (Lemma \ref{lem:proof:condition:bor}).

\begin{lemma}\label{lem:proof:condition:bor}
On the event $\mathcal{E}_{bad}^c(\eta_n) \cap \mathcal{E}_{dist}$, one has $\ep + s_{\om}+b_{sup} \in (0,v]$.
\end{lemma}

 Theorem \ref{thm:dimMink} is proved.\end{proof}\medskip

\begin{proof}\textbf{of Lemma \ref{lem:proof:condition:bor}.} We want to prove that  $\ep + s_{\om}+b_{sup} \in (0,v]$ on the event $\mathcal{E}_{bad}^c(\eta_n) \cap \mathcal{E}_{dist}$. We have already seen that $s_{\om}+b_{sup} \leq 6 err_{n, d}$ on this event, so it is enough to prove that $\ep + 6 err_{n,d} \leq v$. By assumption in Theorem \ref{thm:dimMink}, one has \begin{equation*}2\frac{\textup{log}\, n}{n}\leq \alpha\left(\frac{v}{14}\right)^{2d} \wedge \left(\frac{v}{14}\right)^4,\end{equation*}which implies \begin{equation*}\left(\frac{2 \, \textup{log}\, n}{\alpha \, n}\right)^{1/2d} \leq v/14 \ \, \textup{  and  }  \left(\frac{\textup{log}\, n}{n}\right)^{1/4} \leq v/14,\end{equation*} and thus
\begin{equation*}
    err_{n, d} \leq v/7.
\end{equation*}Finally, one has $\ep + 6 err_{n, d} \leq v$, since $\ep \leq v/7$ by assumption. Hence, $\ep + s_{\om}+b_{sup} \in (0,v]$ on the event $\mathcal{E}_{bad}^c(\eta_n) \cap \mathcal{E}_{dist}$. The lemma is proved.
\end{proof}

\begin{proof}\textbf{of Lemma \ref{proof:lemma:THMdim:proba}.} Let us upper bound the probability $\mathbb{P}_{(\Omega, \mu, W)}(\mathcal{E}_{bad}(\eta_n)\cup \mathcal{E}_{dist}^c )$. The union bound gives\begin{equation*}
\mathbb{P}_{(\Omega, \mu, W)}(\mathcal{E}_{bad}(\eta_n) \cup \mathcal{E}_{dist}^c ) \leq \mathbb{P}_{(\Omega, \mu, W)}(\mathcal{E}_{bad}(\eta_n)) + \mathbb{P}_{(\Omega, \mu, W)}(\mathcal{E}_{dist}^c) \leq \mathbb{P}_{(\Omega, \mu, W)}(\mathcal{E}_{bad}(\eta_n)) + \frac{2}{n}
\end{equation*}where the last inequality comes from Theorem \ref{thm:dist:upperBound}. If the cover defined by $\mathcal{E}_{bad}(\eta_n)$ satisfies the condition \eqref{assump:lem:tec}, then Lemma \ref{lem:bad} ensures that 
\begin{equation}\label{lem:lem:proof:proba}
    \mathbb{P}_{(\Omega, \mu, W)}(\mathcal{E}_{bad}(\eta_n))\leq 2\cov(\eta_n) n \beta \textup{exp}[-\beta (n-1)]. 
\end{equation}

\begin{lemma}\label{lem:bad} Let $B_1, \ldots, B_N$ be $N$ balls in $(\Omega, r_W)$ of measure (strictly) larger than $1/n$, that is,\begin{equation}\label{assump:lem:tec}
\underset{j\leq N}{\textup{min}}\,\mu(B_j)\geq \beta > 1/n     
\end{equation}for some real $\beta$. Then the probability that (at least) one ball contains exactly zero or one sampled point is smaller than \begin{equation*}2N n \beta \textup{exp}[-\beta (n-1)].\end{equation*}
\end{lemma}

Assume that $\eta_n \in (0, v]$ to use the assumption \ref{ass1}. Then, one obtain the following lower bound for the cover defined by $\mathcal{E}_{bad}(\eta_n)$,
\begin{equation*}
    \mu(B_j)\geq \alpha~\eta_n^d = 2\,\textup{log}(n)/n
\end{equation*}so that assumption \eqref{assump:lem:tec} is satisfied. Applying Lemma \ref{lem:bad} for $\beta = \alpha~\eta_n^d $, one has

\begin{equation*}
\mathbb{P}_{(\Omega, \mu, W)}(\mathcal{E}_{bad}(\eta_n)) \leq 2 \cov(\eta) n \alpha~\eta^d  \textup{exp}\left[-\alpha~\eta^d (n-1)\right]. 
\end{equation*}
Combining with the inequality $\cov(\eta) \leq M \eta^{-d}$ from assumption \ref{ass2}, one derive \begin{equation*}\mathbb{P}_{(\Omega, \mu, W)}(\mathcal{E}_{bad}(\eta_n)) \leq 2 M n \alpha~\textup{exp}\left[-\alpha~\eta^d (n-1)\right]
\end{equation*}
and since $\alpha \,\eta_n^d = 2\,\textup{log}(n)/n$, one obtain the upper bound  \begin{equation*} 2 M n \alpha~\textup{exp}\left[-\frac{2\,\textup{log}\,n}{n}(n-1)\right].
\end{equation*} The above display is finally smaller than
\begin{equation*}
    4 M n \alpha \textup{exp}\left[-2\,\textup{log}\,n\right] \leq (4 M \alpha)/n.
\end{equation*}

To conclude the proof, it remains to check the condition $\eta_n \in (0, v]$ that we assume earlier. The following assumption of Theorem \ref{thm:dimMink}
\begin{equation*}2\frac{\textup{log}\, n}{n}\leq \alpha\left(\frac{v}{14}\right)^{2d} \wedge \left(\frac{v}{14}\right)^4\end{equation*}ensures that the radius $\eta_n = \left[2\,\textup{log}(n)/(\alpha \, n)\right]^{1/d}$ satisfies the condition $\eta_n \in (0, v]$. Lemma \ref{proof:lemma:THMdim:proba} is proved.\end{proof}


\begin{proof}\textbf{of Lemma \ref{lem:bad}.} Given $N$ balls $B_1, \ldots, B_N$, let us upper bound the probability that (at least) one of the balls contains exactly zero or one sampled point $\om_i$. With the union bound, this probability is lower than
\begin{equation*}\sum_{j=1}^N\mathbb{P}_{(\Omega, \mu, W)}\left\{B_j\text{ contains exactly 0 or 1 sampled point among } \om_1,\ldots,\om_n\right\}\end{equation*} 
which is again upper bounded with the union bound by \begin{equation*} \sum_{j=1}^N\mathbb{P}_{(\Omega, \mu, W)}\left\{B_j\text{ contains exactly 0 point}\right\}  + \sum_{j=1}^N\mathbb{P}_{(\Omega, \mu, W)}\left\{B_j\text{ contains exactly 1 point}\right\}.  \end{equation*}
Since the probability of the event $\left\{B_j\text{ contains exactly 0 point}\right\}$ is equal to $(1-\mu(B_j))^n$, and since the probability of $\left\{B_j\text{ contains exactly 1 point}\right\}$ is $n \mu(B_j) (1-\mu(B_j))^{n-1}$, the above sum is upper bounded by
\begin{equation*}
     \sum_{j=1}^N(1-\mu(B_j))^n + \sum_{j=1}^N n \mu(B_j) (1-\mu(B_j))^{n-1}.
\end{equation*}
Combining the assumption $\mu(B_j)\geq \beta > 1/n$ with the monotonicity of the functions $x \mapsto (1-x)^n$ and $x \mapsto n x (1-x)^{n-1}$ on $(1/n, 1)$, one has the following upper bound
\begin{equation*}
     N \left[(1-\beta)^n + n \beta (1-\beta)^{n-1}\right]
\end{equation*}which is lower than $2N n \beta (1-\beta)^{n-1}\leq  2N n \beta~\textup{exp}[-\beta (n-1)]$. Lemma \ref{lem:bad} is proved.\end{proof}


\bigskip

\subsection{Lower bound and minimal conditions}

\noindent \textbf{Proof of Theorem \ref{thm:dim:lowerbound}.}
From \citep[][chap.2]{falconer1997techniques}, we deduce directly the following lemma.

\begin{lemma}\label{existence:lem}
Given $L>1$ and $n \geq 2$, there exists a set $\Omega_0 \subset (0,1/(Ln))\times(0,1/(L n))$ with Minkowski dimension $d_2 = 1+\log^{-1}(n)$ w.r.t the Euclidean distance of $[0,1]^2$, and a probability measure $\mu_0$ on $\Omega_0$.
\end{lemma}

Based on $(\Omega_0,\mu_0)$ described in Lemma \ref{existence:lem}, we construct two graphons that are difficult to distinguish for any estimator.
\begin{itemize}
    \item $\Omega_1 = (0,1)\times \{0\} \subset [0,1]^2$ endowed with the uniform measure $\lambda$ on $(0,1)$. In particular, $\lambda ((0,1)\times \{0\}) = 1$.
    \item  $\Omega_2 = \Omega_1 \cup \Omega_0 \subset [0,1]^2$ endowed with the probability measure:
\begin{equation*}
    \mu_2 = (1-n^{-1}) \lambda + n^{-1} \mu_0.
\end{equation*}
\end{itemize}Consider a symmetric function $W : [0,1]^2 \times [0,1]^2 \rightarrow [0,1]$ satisfying a double H\"older condition \eqref{doubleHoldercond} with H\"older exponent $\alpha = 1$. Then, Appendix \ref{subsec:detailsIllustratExamp} shows that the neighborhood distance (associated with such a $W$) behaves like the euclidean distance on $[0,1]^2,$
i.e.:\begin{equation*}
    r_W(\om,\om') \asymp ||\om - \om||_2
\end{equation*}for all $\om, \om' \in [0,1]^2$. Hence, $(\Omega_1, \lambda, W)$ and $(\Omega_2, \mu, W)$ satisfy $dim\, \Omega_2 = 1 + \log^{-1}(n)$ and $dim\, \Omega_1 = 1$, respectively. For brevity, we denote these dimensions by $d_2$ and $d_1$ in the following.

Let us check that all conditions of $\mathcal{W}_{n}(D,\alpha,m,M,v)$ are satisfied by both graphons $(\Omega_1, \lambda, W)$ and $(\Omega_2, \mu_2, W)$. It is clear that $(\Omega_1, \lambda, W)$ belongs to the set $\mathcal{W}_{n}(D,\alpha,m,M,v)$ for large enough $M$ and small enough $\alpha, m$. For the graphon $(\Omega_2, \mu_2, W)$, one has:
\begin{itemize}
    \item \underline{Assumption \ref{ass1}:} for any point $\om \in \Omega_0,$ note $\om_{proj} \in \Omega_1$ its closest point in $\Omega_1$. As $\Omega_0 \subset (0,1/(Ln))^2$, we have $r_W(\om, \om_{proj}) \leq 1/(2n)$ for large enough $L$.  Then, for all $\ep > 1/n$ and all $\om \in \Omega_0,$ one has
\begin{align*}
    \mu_2\left[B(\om,\ep)\right]&\geq (1-n^{-1}) \lambda\left[B(\om,\ep)\right] \\
    &\geq (1-n^{-1}) \lambda\left[B(\om_{proj},\ep-1/(2n))\right]\\
    & \geq \frac{1}{2}\lambda\left[B(\om_{proj},\ep/2)\right].
\end{align*}which is larger than  $\ep$ (up to a numerical constant) since $(\Omega_1, \lambda, W)$ satisfies the condition \ref{ass1} for all $\ep >0$.
\item \underline{Assumption \ref{ass2} lower bound:} $N_{\Omega_2}^{(c)}(\ep) \gtrsim N_{\Omega_1}^{(c)}(\ep) \gtrsim \ep^{-d_1}$ which is larger than $\ep^{-d_2 + \log^{-1}(n)} \gtrsim \ep^{-d_2}$ because $\ep^{\log^{-1}(n)} \asymp 1$ for all $\ep \in (1/n, 1).$ 
    \item \underline{Assumption \ref{ass2} upper bound:} $N_{\Omega_2}^{(c)}(\ep) \lesssim N_{\Omega_1}^{(c)}(\ep) + N_{\Omega_0}^{(c)}(\ep) \lesssim N_{\Omega_1}^{(c)}(\ep)$ since $N_{\Omega_0}^{(c)}(\ep) \lesssim N_{\Omega_0}^{(c)}(1/n) = 1$ for $\ep > 1/n$ and large enough $L$. Combining with the fact that $(\Omega_1,\lambda, W)$ satisfies \ref{ass2}, one obtain $N_{\Omega_2}^{(c)}(\ep) \lesssim \ep^{-d_1}$.
\end{itemize}
Thus, both graphons $(\Omega_1, \lambda, W)$ and $(\Omega_2, \mu_2, W)$ fulfill all conditions of $\mathcal{W}_{n}(D,\alpha,m,M,v)$ for large enough constants $L, M$ and small enough constants $\alpha,m$.

We define the event $\mathcal{E}_{\Omega_1}$ where the i.i.d. sample $\om_1,\ldots,\om_n$ is such that all points $\om_1,\ldots,\om_n$ belong to $\Omega_1$. In particular, for the graphon $(\Omega_2, \mu_2, W)$, the probability of this event is larger than
\begin{equation*}
    \mu_2[\mathcal{E}_{\Omega_1}] \geq (1-n^{-1})^n \geq \frac{1}{3}.
\end{equation*}Then, for any estimator $\hat{d}$ based on the adjacency matrix $A$, one has

\begin{align*}
\Pb_{(\Omega_2, \mu_2, W)} \Big{[}| \hat{d}& - dim\, \Omega_2 | \geq  \frac{1}{2}\log^{-1}(n)\Big{]}  \\ &\geq \Pb_{(\Omega_2, \mu_2, W)} \Big{[}| \hat{d} - dim\, \Omega_2 | \geq  \frac{1}{2}\log^{-1}(n) \big{|} \mathcal{E}_{\Omega_1}\Big{]} \mu_2(\mathcal{E}_{\Omega_1})
\end{align*}which is larger than 
\begin{equation*}
    \frac{1}{3} \Pb_{(\Omega_1, \lambda, W)} \Big{[}| \hat{d} - dim\, \Omega_1 | \leq  \frac{1}{2}\log^{-1}(n)\Big{]} 
\end{equation*}
since $| \hat{d} - dim\, \Omega_1 | \leq  \frac{1}{2}\log^{-1}(n)$ implies $| \hat{d} - dim\, \Omega_2 | \geq  \frac{1}{2}\log^{-1}(n) .$ Thus, by writting \begin{equation*}
    p := \Pb_{(\Omega_1, \lambda, W)} \Big{[}| \hat{d} - dim\, \Omega_1 | >  \frac{1}{2}\log^{-1}(n)\Big{]},
\end{equation*}the above displays entail
\begin{equation*}
    \Pb_{(\Omega_2, \mu_2, W)} \Big{[}| \hat{d} - dim\, \Omega_2 | \geq  \frac{1}{2}\log^{-1}(n)\Big{]} \geq \frac{1}{3}(1-p)
\end{equation*}
which imply that
\begin{align*}
    \underset{ \mathcal{W}_{n}(D,\alpha,m,M,v)}{\textup{sup}} \,&  \Pb_{(\Omega, \mu, W)} \Big{[}| \hat{d} - dim\, \Omega | \geq  \frac{1}{2}\log^{-1}(n)\Big{]}\\  &\geq \underset{(\Omega_1, \lambda, W), (\Omega_2, \mu_2, W)}{\textup{max}} \, \Pb_{(\Omega, \mu, W)} \Big{[}| \hat{d} - dim\, \Omega | \geq  \frac{1}{2}\log^{-1}(n)\Big{]}  \\
    & \geq p \vee
    \frac{1-p}{3}
\end{align*}which is larger than $1/4$. Theorem \ref{thm:dim:lowerbound} is proved.\hfill $\blacksquare$

\bigskip

\noindent \textbf{Proof of Theorem \ref{large:loss:dim}.} There are two cases.\smallskip

\label{append:dim:lowerbound}

\underline{For the class $\mathcal{W}_{n}^{min(1)}(D,\alpha,m,M,v)$}, the condition \ref{ass1} is not imposed. Consider the two following graphons. 
\begin{itemize}
    \item $(\Omega_1, \lambda, W)$ where $\Omega_1 = [0,1]\times \{0\}^{D-1}$ is endowed with the uniform measure $\lambda$ on $[0,1]$, with $\lambda(\Omega_1) =1,$ and where $W : [0,1]^D \times [0,1]^D \rightarrow [0,1]$ is a symmetric function that satisfies a double H\"older condition \eqref{doubleHoldercond} with H\"older exponent $\alpha = 1$.
    \item $(\Omega_2, \mu_2, W)$ where $\Omega_2 = [0,1]^D$ and $\mu_2 = (1-n^{-1}) \lambda + n^{-1} \nu,$ with $\nu$ the uniform measure on $[0,1]^D.$
\end{itemize}
Following the proof of Theorem \ref{thm:dim:lowerbound}, we can show that these two graphons belong to $\mathcal{W}_{n}^{min(1)}(D,\alpha,m,M,v)$, and that  

\begin{equation}\label{repeatSameEq}
 \underset{ \mathcal{W}_{n}^{min(1)}(D,\alpha,m,M,v)}{\textup{sup}} \,  \Pb_{(\Omega, \mu, W)} \Big{[}| \hat{d} - dim\, \Omega | \geq  \frac{D}{2}\Big{]} \geq \frac{1}{4}
\end{equation}which gives the error bound of Theorem \ref{large:loss:dim}. 

\bigskip

\underline{For the class $\mathcal{W}_{n}^{min(2)}(D,\alpha,m,M,v)$}, the assumption \ref{ass2} is not assumed. As in the proof of Theorem \ref{thm:dim:lowerbound}, we can see that the two following graphons belong to $\mathcal{W}_{n}^{min(2)}(D,\alpha,m,M,v)$.
\begin{itemize}
    \item $(\Omega_1, \lambda, W)$ as defined in the above case.
    \item $(\Omega_2, \mu_2, W)$ where $\Omega_2 = [0,1/(Ln)]^D$ for some large enough (numerical) constant $L$, and $\mu_2 = (1-n^{-1}) \lambda + n^{-1} \nu,$ with $\nu$ the uniform measure on $[0,1/(Ln)]^D$.
\end{itemize}
As before, one can prove the error bound \eqref{repeatSameEq} over the class $\mathcal{W}_{n}^{min(2)}(D,\alpha,m,M,v)$. 

Thus, \eqref{repeatSameEq} is proved for $\mathcal{W}_{n}^{min(j)}(D,\alpha,m,M,v)$, with $j\in\{1,2\}$, and Theorem \ref{large:loss:dim} follows.

\hfill $\blacksquare$
\bigskip


\section{Estimation with sparse observations}

\subsection{Proof of Corollary \ref{thm:dist:upperBound:sparse} : estimation of the distances}

Corollary \ref{thm:dist:upperBound:sparse} is a reformulation of Theorem~\ref{thm:dist:upperBound} in the sparse setting and their proofs are almost identical. In this appendix, denote by $W_n$ the function $\rho_n W$. Accordingly, $r_{W_n}$ denotes the neighborhood distance \eqref{def:rW} where $W$ has been replaced with $W_n.$ Hence, $r_{W_n} = \rho_n r_W.$ 

Corollary \ref{thm:dist:upperBound:sparse} is a direct consequence of the two following Lemmas.

\begin{lemma}\label{proof:sparse:inerProdict}
For $\rho_n \geq 2\sqrt{\frac{\textup{log}\, n}{n-2}}$ and $n \geq 5,$ the following event
\begin{equation*}\mathcal{E}_{in}^{sp} := \left\{\forall i,j \in[n]: \ \, \left|  \langle A_{i}, A_{j} \rangle_n - \langle W_n(\omega_{i},.),W_n(\omega_{j},.)\rangle \,\right|  \leq  5\,\rho_n\sqrt{\frac{ \textup{log}\hspace{0.1cm}n}{n}}\right\}\end{equation*}
holds with probability at least $1 - \frac{2}{n}$ with respect to the distribution $\mathbb{P}_{(\Omega, \mu, W),\rho_n}$.
\end{lemma}\label{proof:thm:dist:uppBound:sparse}Following the proof of Proposition \ref{prop:dem:crossTerm}, we show Lemma \ref{proof:sparse:inerProdict} below, by replacing Hoeffding inequality with Bernstein inequality, in order to benefit from the small variance of $A_{ij}$ (which is now of the order of $\rho_n$). 

\begin{lemma} \label{prop:quad:dist:sparse}Conditionally to the event $\mathcal{E}_{in}^{sp}$, the following inequalities
$$ \forall i \in [n]: \ \,\left| \langle A_{i} , A_{\widehat{m}(i)} \rangle_n - \langle W_n(\omega_{i},.),W_n(\omega_{i},.)\rangle \right|
             \leq 3 \rho_n\,r_{W_n}(\om_i,\om_{m(i)}\,) \, + \,25\, \rho_n   \sqrt{\textup{log}(n)/n }$$
hold simultaneously.  
\end{lemma}The proof of lemma \ref{prop:quad:dist:sparse} is almost the same as for Proposition \ref{prop:quad:dist}. It is omitted.\bigskip

\begin{proof}\textbf{of Lemma \ref{proof:sparse:inerProdict}}
Conditionally to $\om_i, \om_j$, $i\neq j$, the $n-2$ random variables $\{A_{ik}A_{kj}: k \in[n], k \neq i,j\}$ are independent with expectation $\mathbb{E}\left[A_{ik}A_{kj}\right] = \int_			{\Omega}W_n(\om_i,z)W_n(\om_{j},z)\mu(dz) $ for all $ k \neq i,j$ (where $\mathbb{E}$ is taken w.r.t. the distribution $\mathbb{P}_{(\Omega, \mu, W),\rho_n})$. Using Bernstein inequality \citep[see][for instance]{sridharam}, one has 

\begin{equation*}\mathbb{P}_{(\Omega, \mu, W),\rho_n}\Bigg{(}\frac{1}{n-2}\Big{|}\sum_{k\neq i,j}A_{ik}A_{kj} - 				(n-2) \int_			{\Omega}W_n(\om_i,z)W_n(\om_{j},z)\mu(dz)			\Big{|} \geq  \epsilon \ \, \Bigg{|} \om_i, \om_j\Bigg{)} \end{equation*}smaller than \begin{equation*}
    \leq 2 \textup{exp}\left(\frac{-(n-2)\ep^2}{2\rho_n^2+2\ep/3}\right) 
\end{equation*}

\noindent for $\epsilon > 0$. Since the above inequality is satisfied for almost every $\om_i, \om_j \in \Omega$, we have the same upper bound for the non-conditional probability. Then, setting $\epsilon = 3\rho_n\sqrt{\frac{ \,\textup{log}\hspace{0.1cm}n}{n-2}}$ gives 

\begin{equation*}2 \textup{exp}\left(\frac{-(n-2)\ep^2}{2\rho_n^2+2\ep/3}\right)\leq 2 \textup{exp}\left(\frac{-9 \,\textup{log}\hspace{0.1cm}n}{2+\frac{2}{\rho_n}\sqrt{\frac{ \,\textup{log}\hspace{0.1cm}n}{n-2}}}\right) \leq \frac{2}{n^3}\end{equation*}since $\rho_n \geq 2\sqrt{\frac{\textup{log}\, n}{n-2}}$ by assumption. Thus, by using the union bound over all $i \neq j$, one obtain 

\begin{equation*} \mathbb{P}_{(\Omega, \mu, W),\rho_n}\Bigg{(}\bigcup_{i,j: i\neq j} \Bigg{\{}\frac{1}{n-2}\left|\sum_{k\neq i,j}A_{ik}A_{kj} - 				(n-2) \int_			{\Omega}W_n(\om_i,z)W_n(\om_{j},z)\mu(dz)			\right| \geq \epsilon \Bigg{\}}\Bigg{)} \leq \frac{2}{n}. \end{equation*}

And finally, following the proof of Proposition \ref{prop:dem:crossTerm} leads to

\begin{equation*}\underset{i,j: i\neq j} {\mbox{max}}  \left|\sum_{k}\frac{A_{ik} A_{kj}}{n} - \int_{\Omega}W_n(\om_i,z)W_n(\om_{j},z)\mu(dz)\right|  \leq 3\rho_n\sqrt{\frac{ \,\textup{log}\, n}{n-2}} + \frac{4}{n} \end{equation*}with probability at least $1 - \frac{2}{n}$. To conclude the proof, observe that above display is upper bounded by 
\begin{equation*}
    \leq 5\rho_n\sqrt{\frac{ \textup{log}\hspace{0.1cm}n}{n}}
\end{equation*}as soon as $n\geq5$.
\end{proof}
\subsection{Proof of of Corollary \ref{coro:asympDim:sparse} : estimation of the dimension}

In the proof of Theorem \ref{thm:dimMink}, one has seen

\begin{equation*}\frac{\textup{log} \,  m}{-\textup{log}\,\ep} - d \frac{ (s_{\om}+b_{sup})/\ep }{-\textup{log} \ep} \leq \frac{\textup{log} \, \covh(\ep) }{-\textup{log}\,\ep} - d  \leq \frac{\textup{log} \, M }{-\textup{log}\,\ep}+
d \, \frac{ (s_{\om}+b_{sup})/\ep + \left((s_{\om}+b_{sup})/\ep\right)^2}{-\textup{log}\,\ep}. 
\end{equation*} 

\noindent The sampling error $s_{\om}$ is not affected by the sparsification of the data through $\rho_n$, and thus takes the same value as in Theorem \ref{thm:dimMink}. On the other hand, the distance error bound $b_{sup}$ changes, and is now defined as
\begin{equation*}
    b_{sup}^2 :=  6 \underset{1 \leq i \leq n}{\textup{max}} r_{W}(\om_i,\om_{m(i)}) + \frac{60}{\rho_n} \sqrt{\textup{log}(n)/n }
\end{equation*}according to Corollary \ref{thm:dist:upperBound:sparse}. Following the proof of Theorem \ref{thm:dimMink}, one has
\begin{equation*}b_{sup}+s_{\om} \leq  6 \left(\frac{2 \, \textup{log}\, n}{\alpha \, n}\right)^{1/2d} + \frac{8}{\sqrt{\rho_n}} \left(\frac{\textup{log}\, n}{n}\right)^{1/4}. \end{equation*}Define\begin{equation}\label{err:sparse:proof:rad}
    err_{n,d,\rho_n} := \left(\frac{2 \, \textup{log}\, n}{\alpha \, n}\right)^{1/2d} + \frac{1}{\sqrt{\rho_n}} \left(\frac{\textup{log}\, n}{n}\right)^{1/4}
\end{equation}so that 
\begin{equation*}
b_{sup}+s_{\om} \leq 8 \, err_{n,d,\rho_n}.    
\end{equation*} \label{append:E:dim:sparse}

Following the proof of Theorem \ref{thm:dimMink}, one obtain the same error bound for the dimension estimation, after replacing $6\,err_{n,d}$ with $8\,err_{n,d,\rho_n}$. Indeed, one has

\begin{equation}\label{ineq:dim:sparse:proof}
\Bigg{|} \frac{\textup{log} \, \covh(\ep) }{-\textup{log}\, \ep} - d \Bigg{|}  \, \leq \, \frac{1}{-\textup{log}\,\ep}\Bigg{[}\textup{log} \,\left( M \vee \frac{1}{m}\right) \,  + \, 8 d  \frac{err_{n, d}}{\ep} \left( 1 +  \frac{err_{n, d}}{\ep}\right)\Bigg{]}
\end{equation} 

\noindent for all $\ep \in ]0, v/9]$ and all $n$ such that

\begin{equation*}
    2 \textup{log}\, n/n\leq \alpha \left(v/18\right)^{2d} \wedge  \rho_n^2\left(v/18\right)^4.
\end{equation*}\smallskip

As in the proof of Theorem \ref{coro:asympDim}, one minimizes the error bound \eqref{ineq:dim:sparse:proof} by choosing a particular radius of the order of $\underset{\{d: \, d \leq D\}}{\textup{sup}}err_{n, d,\rho_n}$ $=$ $err_{n, D,\rho_n}$ . This gives a radius that satisfies the following relation \begin{equation*}
\ep_{D, \rho_n} \asymp  \left(\frac{\textup{log}\, n}{n}\right)^{1/( 2 D)} \vee \frac{1}{\sqrt{\rho_n}}\left(\frac{\textup{log}\, n}{n}\right)^{1/4}.
\end{equation*} \smallskip Corollary \ref{coro:asympDim:sparse} follows from the plug-in of $\ep_{D, \rho_n}$ in \eqref{ineq:dim:sparse:proof}. \hfill $\blacksquare$\bigskip


\section{Testing the complexity via under-estimation of the packing number}

The current appendix is organized as follows. We first analyse the performance of the new distance estimator \eqref{def:new:dist:estim} and then deduce a control on the type I and II errors of the test.

\subsection{Performance of the new distance estimator}
Lemma \ref{lem:undere} shows that the new distance-estimator $\widehat{r}_{new}$ does not over-estimate $r_W$ in the sense of (\ref{ineq:under}), without underestimating too much (\ref{ineq:notunder}). Let $U$ be the function defined by \begin{equation}\label{def:U:new}U(i) = \textup{argmax}_{\,t\in\{i,\widehat{m}(i)\}} \langle W(\omega_{t},.),W(\omega_{t},.)\rangle\end{equation}for all $i \in [n]$. This means that $U(i)$ indicates which of the two functions $W(\omega_{i},.)$ or $W(\omega_{\widehat{m}(i)},.)$ has the largest $l_2$-norm $||.||_{2,\mu}$ (see Section \ref{subsec:def:disEstim} for the definitions of the inner product and the norm). 
\begin{lemma}
\label{lem:undere}Consider $t_n = 12 \,\sqrt{\frac{\textup{log}\, n}{n}}$ a fluctuation term and the function $U$ introduced in \eqref{def:U:new}. One has the following bounds on the new distance estimator \eqref{def:new:dist:estim}  \begin{eqnarray}
\label{ineq:under}
  \widehat{r}_{new}^2(i,j)&\leq&  r_W^2(\om_{U(i)},\om_{U(j)})  + t_n  \\
 \label{ineq:notunder}
  \widehat{r}_{new}^2 (i,j)&\geq&  r_W^2(\om_i,\om_j) -  5 \, r_W(\om_i,\om_{m(i)}) - 5 \, r_W(\om_j,\om_{m(j)}) -  5 t_n  \end{eqnarray}
 holding simultaneously for all $i,j\in [n]$ with probability at least $1-\frac{2}{n}$ with respect to the distribution $\mathbb{P}_{(\Omega,\mu,W)}$.
\end{lemma}

Recall the useful Proposition \ref{prop:dem:crossTerm} on the convergence of the inner products: the event $\mathcal{E}_{in}$ where the following inequalities hold simultaneously for all $i \neq j$\begin{equation}\label{ineq:rappel:pratik}
    \left|  \langle A_{i}, A_{j} \rangle_n - \langle W(\omega_{i},.),W(\omega_{j},.)\rangle \,\right|\leq 3\sqrt{\textup{log}\hspace{0.1cm}n / n}
\end{equation}happens with probability at least $1-2/n$.\bigskip

\begin{proof}\textbf{of \eqref{ineq:under}.} Assume the above event $\mathcal{E}_{in}$ holds. For all $i,j\in [n]$ such that $\{i,\widehat{m}(i)\}\cap \{j,\widehat{m}(j)\}=\emptyset$, the line \eqref{ineq:rappel:pratik} gives  
\begin{equation*}
\widehat{r}^2_{new}(i,j) \leq \langle W(\omega_{i},.),W(\omega_{\widehat{m}(i)},.)\rangle + \langle W(\omega_{j},.),W(\omega_{\widehat{m}(j)},.)\rangle  - 2\,\underset{v \in \{i,\widehat{m}(i)\}, w \in \{j,\widehat{m}(j)\}}{\mbox{max}} \langle W(\omega_{v},.),W(\omega_{w},.)\rangle + t_n
\end{equation*}
with $t_n=12 \,\sqrt{\frac{\textup{log}\, n}{n}}$. Then, using the function $U$ defined by \eqref{def:U:new}, one has
\begin{equation*}
\widehat{r}^2_{new}(i,j)  \leq \langle W(\omega_{U(i)},.),W(\omega_{U(i)},.)\rangle +  \langle W(\omega_{U(j)},.),W(\omega_{U(j)},.)\rangle  - 2 \, \langle W(\omega_{U(i)},.),W(\omega_{U(j)},.)\rangle + t_n
\end{equation*}which is upper bounded by 
\begin{equation*}r_W^2(\om_{U(i)},\om_{U(j)})  + t_n
\end{equation*}with Cauchy-Schwarz inequality.
The line (\ref{ineq:under}) is proved in the case $\{i,\widehat{m}(i)\}\cap \{j,\widehat{m}(j)\}=\emptyset$.\medskip

If $\{i,\widehat{m}(i)\}\cap \{j,\widehat{m}(j)\} \neq \emptyset$, we can see that $\widehat{r}^2_{new}(i,j) \leq 0$. Thus \eqref{ineq:under} trivially holds in this case too. The inequalities \eqref{ineq:under} are proved.\end{proof}

\begin{proof}\textbf{of \eqref{ineq:notunder}.} Assume the event $\mathcal{E}_{in}$ of Proposition \ref{prop:dem:crossTerm} holds.\medskip

\underline{If $i,j\in [n]$ such that $\{i,\widehat{m}(i)\}\cap \{j,\widehat{m}(j)\}=\emptyset$},
\begin{equation*}
    \left| r_W^2(\om_i,\om_j) - \widehat{r}^2_{new}(i,j) \right| \leq 
\left| r_W^2(\om_i,\om_j) - \widehat{r}^2( i, j) \right| + \left| \widehat{r}^2( i, j) - \widehat{r}^2_{new}(i,j) \right|
\end{equation*}by triangle inequality. The left term is upper bounded by \begin{equation*}
3\, r_W(\om_j,\om_{m(j)}\,) + 3\, r_W(\om_i,\om_{m(i)}\,) + 36\, \sqrt{\textup{log}(n)/n }
\end{equation*}thanks to Theorem \ref{thm:dist:upperBound}. The right term is equal to \begin{equation*}
    2\left| \langle A_i,A_j \rangle  - \max_{k\in \{i,\widehat{m}(i)\}, l\in \{j,\widehat{m}(j)\}} \langle A_k,A_l \rangle  \right|
\end{equation*}which is upper bounded by\begin{equation*}
 r_W(\om_j,\om_{m(j)}\,) +  r_W(\om_i,\om_{m(i)}\,) + 12\, \sqrt{\textup{log}(n)/n }
\end{equation*}using the same technique as in the proof of Theorem \ref{thm:dist:upperBound}. Combining the above displays, one has
\begin{equation*}
\left| r_W^2(\om_i,\om_j) - \widehat{r}^2_{new}(i,j) \right| \leq  5\, r_W(\om_j,\om_{m(j)}\,) + 5\, r_W(\om_i,\om_{m(i)}\,) + 60\, \sqrt{\textup{log}(n)/n },
\end{equation*}
which implies  \begin{equation}\label{ineq:desired}
\widehat{r}^2_{new}(i,j) \geq r_W^2(\om_i,\om_j) -  5\, r_W(\om_j,\om_{m(j)}\,) - 5\, r_W(\om_i,\om_{m(i)}\,) - 60\, \sqrt{\textup{log}(n)/n }.
\end{equation}The line (\ref{ineq:notunder}) is therefore proved in the case $\{i,\widehat{m}(i)\}\cap \{j,\widehat{m}(j)\}=\emptyset$.\medskip

\underline{If $i,j\in [n]$ such that $\{i,\widehat{m}(i)\}\cap \{j,\widehat{m}(j)\}\neq \emptyset$}, \label{proof:lem:undere}
\begin{equation*}
    \widehat{r}_{new}(i,j)=0.
\end{equation*}
Hence, it is enough to show that the right hand side of (\ref{ineq:desired}) is non-positive. Consider the particular case where $\widehat{m}(i)= j$ and $i \neq \widehat{m}(j)$ for example. Then, one has
\begin{equation*}
    |\widehat{r}^2(i,j)|=\big| \langle A_{\widehat{m}(j)},A_{j} \rangle -  \langle A_i,A_{j} \rangle \big| \leq | \langle A_i, A_j-A_{\widehat{m}(j)}\rangle| + |\langle A_i- A_j, A_{\widehat{m}(j)}\rangle|
\end{equation*}which is upper bounded by 
\begin{equation*}\widehat{f}(j,m(j)) + \widehat{f}(i,m(i))
\end{equation*}where $\widehat{f}$ has been introduced in \eqref{minim}. As in the proof of Theorem \ref{thm:dist:upperBound}, one can show that the above display is upper bounded by
\begin{equation*}
r_W(\om_j,\om_{m(j)}\,) + r_W(\om_i,\om_{m(i)}\,) + 12\, \sqrt{\textup{log}(n)/n }
\end{equation*}on the event $\mathcal{E}_{in}$. Combining this upper bound of $\widehat{r}$ with the following lower bound from Theorem \ref{thm:dist:upperBound}
 \begin{align}\label{ineq:negative}\widehat{r}^2(i,j)\geq r^2_{W}(\omega_i,\omega_j) - 3\,r_{W}(\omega_i,\omega_{{m(i)}})-  3\,r_{W}(\omega_j,\omega_{{m(j)}})- 36\, \sqrt{\textup{log}(n)/n },\end{align} 
 one derive \begin{equation*}r^2_{W}(\omega_i,\omega_j)\leq 4\,r_W(\omega_i,\omega_{{m(i)}})+4\,r_W(\omega_j,\omega_{{m(j)}})+ 48\, \sqrt{\textup{log}(n)/n }.\end{equation*}
 
This implies that the right hand side of (\ref{ineq:desired}) is non positive. Hence (\ref{ineq:notunder}) is proved in the particular case $\widehat{m}(i)= j$ and $i \neq \widehat{m}(j)$. By symmetry, it remains only the case $\widehat{m}(i)= \widehat{m}(j)$ to do. Following the above proof, we can show taht \eqref{ineq:notunder} holds for this case too. The inequality (\ref{ineq:notunder}) is therefore proved in the case $\{i,\widehat{m}(i)\}\cap \{j,\widehat{m}(j)\}\neq \emptyset$.\medskip
 
The line \eqref{ineq:notunder} is proved.\end{proof}

\subsection{Control on the type I and II errors}

In Theorem \ref{pack:under:last} on the new packing number estimator, the left hand side of \eqref{lem:articl:underestim} is similar to Section \ref{subsection::theoreticGuaranteeCovNUmb} on the covering number estimator, and thus straightforward. The right hand side of \eqref{lem:articl:underestim} and Corollary \ref{coro:typeIerror} are proved together below.\bigskip

\begin{proof}\textbf{for the type I error.} Assume the null-hypothesis $\pac(\ep) \leq K$ holds. We want to show that the same inequality is satisfied by the statistic $\pachne(\widehat{\ep})$. Proof by contradiction: assume the inequality $\pachne(\widehat{\ep}) \geq K + 1$ holds. This means that there are $K+1$ indices $i_1,\ldots,i_{K+1} \in [n]$ such that the following inequalities hold 
\begin{equation*}
\forall s, t \in \{1,\ldots,K+1\}: \ \, \ \, \widehat{\ep}^2 \, < \,\widehat{r}_{new}^2(i_s, i_{t}).
\end{equation*} 
Combining the above inequalities with the under-estimation property \eqref{ineq:under}, one has  \begin{equation*}\forall s, t \in \{1,\ldots,K+1\}: \ \, \ \, \widehat{\ep}^2\, <  \, r_W^2(\om_{U(i_s)},\om_{U(i_t)}) + t_n \end{equation*}
with probability at least $1-2/n$. Replacing the radius $\widehat{\ep}^2$ by its value $\ep^2 + t_n$, it comes \begin{equation*}\forall s, t \in \{1,\ldots,K+1\}: \ \, \ \, \ep^2\, <  \, r_W^2(\om_{U(i_s)},\om_{U(i_t)}). \end{equation*}Thus, $K+1$ sampled points are separated by at least a distance $\ep$, which implies $\pac(\ep) \geq K+1$. This contradicts the null-hypothesis.\bigskip \end{proof}
Corollary \ref{coro:typeIerror} and Theorem \ref{pack:under:last} are therefore proved.\bigskip

\begin{proof}\textbf{for the type II error (Theorem \ref{thm:type2}).} Consider a graphon $(\Omega, \mu, W)$ in the set $\mathcal{W}(\eta, \beta)$. By definition of $\mathcal{W}(\eta, \beta)$, there are $K+1$ balls in $(\Omega, r_W)$ whose centers are separated by at least a distance $\sqrt{\ep^2 + 10\eta + 6 t_n} +  \eta$. Label these balls by $s \in \{1,\ldots,K+1\}$. As in the proof for the dimension estimation, assume the complementary of the event $\mathcal{E}_{bad}$, i.e. assume that each of the $K+1$ balls contains at least two sampled points. Accordingly, denote by $i_1,j_1,\ldots,i_{K+1},j_{K+1}$ the indices of the corresponding sampled points such that $\om_{i_s}, \om_{j_s}$ belong to the $s^{\textup{th}}$ ball with $s\in \{1,\ldots,K+1\}$. Since the radius of these ball is smaller than $\eta/2$, one has  
\begin{equation}\label{dernierProof}r_W(\om_{i_s}, \om_{m(i_s)})\leq r_W(\om_{i_s}, \om_{j_s})\leq \eta\end{equation}for all $s\in \{1,\ldots,K+1\}$. 

On the event $\mathcal{E}_{in}$ of Proposition \ref{prop:dem:crossTerm}, Lemma \ref{lem:undere} gives \begin{equation*}
\widehat{r}_{new}^2 (i_s,i_t)\geq  r_W^2(\om_{i_s},\om_{i_t}) -  5 \, r_W(\om_{i_s},\om_{m(i_s)}) - 5 \, r_W(\om_{j_s},\om_{m(j_s)}) -  5 t_n.
\end{equation*}for all $s\neq t \in \{1,\ldots,K+1\}$.\label{appendiTypetwoerror} Using \eqref{dernierProof}, one derive 
\begin{equation}\label{eq:fin:proof:test}
   \widehat{r}_{new}^2 (i_s,i_t) \geq r_W^2(\om_{i_s},\om_{i_t}) -  10 \eta -  5 t_n.    
\end{equation}The ball centers are separated by at least a distance $\sqrt{\ep^2 + 10\eta + 6 t_n} +  \eta$ by assumption, which implies that the points in these balls are separated by
\begin{equation*}
r_W(\om_{i_s},\om_{i_t}) > \sqrt{\ep^2 + 10\eta + 6 t_n}
\end{equation*}for all $s\neq t \in \{1,\ldots,K+1\}$, since the ball radii are all smaller than $\eta/2$. Combining this inequality with the line \eqref{eq:fin:proof:test}, one obtain
\begin{equation*}
\widehat{r}_{new}^2 (i_s,i_t)  > \ep^2 + t_n 
\end{equation*}for all $s \neq t \in \{1,\ldots,K+1\}$. Since $\widehat{\ep}= \sqrt{\ep^2 + t_n},$ this gives $\pachne(\widehat{\ep}) \geq K+1$. Thus, the alternative hypothesis is confirmed correctly. \medskip

The above displays hold on the event $\mathcal{E}_{in} \cap \mathcal{E}_{bad}^c.$ Let us upper bound the probability of the complementary event. The union bound gives
\begin{equation*}
    \mathbb{P}(\mathcal{E}_{in}^c\cup \mathcal{E}_{bad}) \leq \frac{2}{n}+(K+1) 2 n \beta~\textup{exp}[-\beta (n-1)]
\end{equation*}thanks to Proposition \ref{prop:dem:crossTerm} and Lemma \ref{lem:bad}. Theorem \ref{thm:type2} is then proved. \end{proof}\bigskip

\begin{proof}\textbf{for the improvement of the type II error (Theorem \ref{coro:type2}).} We have seen that the type II error is upper bounded by the probability of the event $\mathcal{E}^c_{in} \cap \mathcal{E}_{bad}.$ Here the only difference is that $\mathcal{E}_{bad}$ refers to the new event where, for each of the $M$ collections of $K+1+K'$ balls, at least $K'+1$ balls contain strictly less than two sampled points. For clarity, label these collections by $\{1,\ldots,M\}$, and denote by $\mathcal{C}_j$ the event where at least $K'+1$ balls of the $j^{\textup{th}}$ collection contain strictly less than two sampled points. Then, we have

\begin{equation*}
    \mathbb{P} [\mathcal{E}_{bad}] = \mathbb{P} [  \mathcal{C}_{1} \cap \ldots \cap \mathcal{C}_{M} ]
\end{equation*}where $\mathbb{P}$ denote the probability distribution $\mathbb{P}_{(\Omega,\mu,W)}$ of the W-random graph. The above display is equal to 
\begin{equation*}
     \mathbb{P} [ \, \mathcal{C}_{1}] \times \mathbb{P}[\mathcal{C}_{2} \big{|} \mathcal{C}_{1}] \times \ldots \times \mathbb{P}[\mathcal{C}_{M} \big{|} \mathcal{C}_1,\ldots,\mathcal{C}_{M-1}]
\end{equation*}which is upper bounded by 
\begin{equation*}
    \mathbb{P}[ \, \mathcal{C}_{1}] \times \mathbb{P}[\mathcal{C}_{2}] \times \ldots \times \mathbb{P}[\mathcal{C}_{M}]
\end{equation*}since the events $\mathcal{C}_1,\ldots\mathcal{C}_M$ are negatively associated (it is shown at the end of the proof). Finally, we have
\begin{equation}\label{improv:proof:res1}
     \mathbb{P} [\mathcal{E}_{bad}] \leq \mathbb{P} [  \mathcal{C}_{1}]^M. 
\end{equation}\smallskip

Given the first collection of $K+1+K'$ balls, denote by $\mathcal{E}_{j}$ the event where the $j^{\textup{th}}$ ball of the collection contains strictly less than two sampled points. By definition of the event $\mathcal{C}_{1}$, we have
\begin{equation*}
    \mathbb{P} [ \, \mathcal{C}_{1}]  = \mathbb{P} [ \, \exists \, i_1,\ldots,i_{K'+1} \in \{1,\ldots,K+1+K'\} : \mathcal{E}_{i_1} \cap \ldots \cap \mathcal{E}_{i_{K'+1}} ].
\end{equation*}The union bound gives  
\begin{equation*}
   \mathbb{P} [ \, \mathcal{C}_{1}] \leq \underset{i_1,\ldots,i_{K'+1}}{\sum} \mathbb{P} [ \, \mathcal{E}_{i_1} \cap \ldots \cap \mathcal{E}_{i_{K'+1}} ]
\end{equation*}where the sum is taken over all possible $K'+1$ different indices. The above upper bound is equal to 
\begin{equation*}
    \underset{i_1,\ldots,i_{K'+1}}{\sum} \mathbb{P} [ \, \mathcal{E}_{i_1}] \times \mathbb{P}[\mathcal{E}_{i_2} \big{|} \mathcal{E}_{i_1}] \times \ldots \times \mathbb{P}[\mathcal{E}_{i_{K'+1}} \big{|} \mathcal{E}_{i_1},\ldots,\mathcal{E}_{i_{K'}}]. 
\end{equation*}which is smaller than
\begin{equation}\label{test:improv:proof}
    \underset{i_1,\ldots,i_{K'+1}}{\sum} \mathbb{P} [ \, \mathcal{E}_{i_1}] \times \ldots \times P[\mathcal{E}_{i_{K'+1}}] 
\end{equation}by negative association of the events $\mathcal{E}_k$ (this fact is proved at the end). Finally, Lemma \ref{lem:bad} ensures that \begin{equation*}
     \mathbb{P} [ \, \mathcal{E}_{k}]\leq  2\beta n~\textup{exp}[-\beta (n-1)]
\end{equation*}for all $k$, which allows to upper bound \eqref{test:improv:proof} and have
\begin{equation}\label{improv:proof:res2}
\mathbb{P} [ \, \mathcal{C}_{1}] \leq  \binom{K+K'+1}{K'+1}\, \Big(2\beta n~\textup{exp}[-\beta (n-1)]\Big)^{(K'+1)}.     
\end{equation}Thus, setting $\widetilde{p}_n = \mathbb{P} [  \mathcal{C}_{1}]$, we deduce from \eqref{improv:proof:res1} that
\begin{equation*}
    \mathbb{P}(\mathcal{E}_{in}^c\cup \mathcal{E}_{bad})\leq \mathbb{P}(\mathcal{E}_{in}^c) + \mathbb{P}(\mathcal{E}_{bad}) \leq \frac{2}{n}+\widetilde{p}_n^M,
\end{equation*}where $\widetilde{p}_n$ is upper bounded by \eqref{improv:proof:res2}.

\bigskip

It remains to show the negative association that we use in the above proof.
Given the first collection of $K+1+K'$ balls, let us show that the corresponding events $\mathcal{E}_1,\ldots,\mathcal{E}_{K+1+K'}$ are negatively associated. For the $n$ sampled points $\om_1,\ldots,\om_n$, define $n_j$ the number of points in the $j^{\textup{th}}$ ball of the collection. Theorem 13 of \citet{dubha} ensures that the variables $n_1,\ldots,n_{K+1+K'}$ are negatively associated. Define the non-increasing function $h (n_j) = \mathbb{I}_{\mathcal{E}_j}$ where $\mathbb{I}_{\mathcal{E}_j}$ is the indicator function of $\mathcal{E}_j$. The second point of Proposition 7 of \citet{dubha} shows that $h (n_1),\ldots,h(n_{K+1+K'})$  are negatively associated. This means that the events $\mathcal{E}_1,\ldots,\mathcal{E}_{K+1+K'}$ are negatively associated. \smallskip

Similarly, we show the negative association of the events $\mathcal{C}_1,\ldots,\mathcal{C}_M$. Consider $n_j^t$ the number of sampled points in the $j^{\textup{th}}$ ball of the $t^{\textup{th}}$ collection. These variables are negatively associated according to Theorem 13 of \citet{dubha}. Define the non-increasing functions $h_t(n_1^t,\ldots,n_{K+1+K'}^t) = \mathbb{I}_{\,\mathcal{C}_j}$ for all $t \leq M$. Then, Proposition 7 of \citet{dubha} shows that $\mathbb{I}_{\,\mathcal{C}_1},\ldots,\mathbb{I}_{\,\mathcal{C}_M}$ are negatively associated. 

Theorem \ref{coro:type2} is proved.\end{proof}

\end{document}